\documentclass{article} %
\usepackage{iclr2027_arxiv,times}

\newcommand{\rebuttal}[1]{#1}%
\newcommand{\chgd}[1]{#1}%

\usepackage{amsmath,amsfonts,bm}

\def\eqref#1{equation~\ref{#1}}
\def\1{\bm{1}}

\def\rd{{\textnormal{d}}}

\DeclareMathAlphabet{\mathsfit}{\encodingdefault}{\sfdefault}{m}{sl}
\SetMathAlphabet{\mathsfit}{bold}{\encodingdefault}{\sfdefault}{bx}{n}

\newcommand{\E}{\mathbb{E}}

\newcommand{\R}{\mathbb{R}}

\newcommand{\Var}{\mathrm{Var}}

\usepackage[utf8]{inputenc} %
\usepackage[T1]{fontenc}    %
\usepackage[
    colorlinks=true,
    linkcolor=BrickRed,
    citecolor=ForestGreen,
    filecolor=MidnightBlue,
    urlcolor=MidnightBlue,
]{hyperref}
\usepackage{url}            %
\usepackage{booktabs}       %
\usepackage{amsfonts}       %
\usepackage{nicefrac}       %
\usepackage{microtype}      %
\usepackage[dvipsnames]{xcolor}         %

\usepackage{dcolumn}
\makeatletter
\newcolumntype{d}{>{\DC@{.}{.}{1}}c<{\DC@end}}
\newcolumntype{B}{>{\boldmath\DC@{.}{.}{1}}c<{\DC@end}}
\makeatother
\newcommand{\hd}[1]{\multicolumn{1}{c}{#1}}
\newcommand{\bd}[1]{\multicolumn{1}{B}{#1}}
\newcommand{\mc}[2]{\makecell[l]{#1\\[-2pt] {\scriptsize #2}}}

\usepackage{xurl}    
\usepackage{cancel}
\usepackage{amsmath}
\usepackage{amssymb}
\usepackage{mathtools}
\usepackage{amsthm}
\usepackage{bm}
\usepackage{wrapfig}
\usepackage[capitalize,noabbrev,sort]{cleveref}
\usepackage{todonotes}
\usepackage{changes}
\definechangesauthor[name={Eliot}, color=blue]{EW}
\definechangesauthor[name={Miltos}, color=Bittersweet]{MK}
\definechangesauthor[name={Dim}, color=red]{DK}
\definechangesauthor[name={Nikolaj}, color=cyan]{NM}
\definechangesauthor[name={Stratis}, color=green]{EG}
\usepackage{adjustbox}
\usepackage{makecell}
\usepackage{enumitem}

\usepackage{subcaption}
\usepackage{capt-of} %
\usepackage{algorithm}
\usepackage{algorithmic}
\usepackage{tabularx}
\usepackage{multirow}
\usepackage{siunitx} %
\usepackage{etoc} %
\usepackage[dvipsnames]{xcolor}         % colors

\newcommand{\norm}[1]{\left\lVert #1 \right\rVert}
\newcommand{\Id}{\boldsymbol{\mathrm{I}}} %\mathrm{Id}}          % identity. NOT \mI: math_commands.tex
\newcommand{\dt}{\Delta{t}}
\newcommand{\dtau}{\Delta{\tau}}
\newcommand{\condprob}[2]{p(#1 \mid #2)}   % \mid, not |, for correct spacing

\newcommand{\sistate}{\bm{x}}
\newcommand{\bh}{\bm{h}}                   % conditioning history x^{t-\dt:t}
\newcommand{\gen}[1]{\hat{#1}}             % model-generated realisation
\newcommand{\bz}{\bm{z}}

\newcommand{\bI}{\bm{I}}                   % interpolant
\newcommand{\SIstate}{\bm{X}}              % SDE state
\newcommand{\bW}{\bm{W}}                   % Wiener process

\newcommand{\bR}{\bm{R}}
\newcommand{\drift}{\bm{B}}                % drift term in the SDE
\newcommand{\bD}{\bm{D}}                   % denoiser
\newcommand{\bWnet}{\bm{W}_\theta}         % W-prediction network
\newcommand{\prior}{\bm{G}_\phi}           % pretrained deterministic prior
\newcommand{\lossw}[1]{\lambda_{#1\mid\bD}}

\title{Xaurora: Generative Weather Forecasting with Denoising Stochastic Interpolants from a Foundation Model Prior}

\author{
Eliot Walt$^{1,2}$, Miltiadis Kofinas$^1$, Nikolaj T. M{\"u}cke$^{3,4}$, Efstratios Gavves$^2$, Dim Coumou$^1$ \\
$^1$Vrije Universiteit Amsterdam \quad $^2$University of Amsterdam \quad $^3$Delft University of Technology \\ $^4$Centrum Wiskunde \& Informatica \\
\texttt{\{e.r.walt, m.kofinas, d.coumou\}@vu.nl}, \texttt{e.gavves@uva.nl}, \\\texttt{n.t.mucke-1@tudelft.nl}
}

\iclrfinalcopy %
\begin{document}
\etocdepthtag.toc{main}%
\maketitle

\begin{abstract}
Deep learning has revolutionised weather forecasting in recent years, especially through atmospheric foundation models, which offer competitive skill for a fraction of the computational costs of classic physics-based models. However, most existing foundation models are deterministic, limiting the generation of large ensembles for accurate uncertainty quantification, extreme weather risk assessment, and long-range weather forecasting. Furthermore, these models incur a large, often prohibitive, computational overhead to train from scratch. To address these shortcomings, we turn a pretrained deterministic prior model, namely the Aurora foundation model, into a generative ensemble-prediction model. To that end, we introduce a novel generative method, \textit{Denoising Stochastic Interpolants}, combined with a \emph{replay buffer} for Stochastic Differential Equation (SDE) rollout, enabling probabilistic training of SDE trajectories. Our stochastic foundation model, \textit{Xaurora}, is finetuned from the small Aurora version, yet it approaches the state-of-the-art on global ensemble metrics and is competitive with the large version of Aurora. Our method is parameter and sample efficient, and generates skilful 15-day forecasts in \chgd{7 minutes}. Our results demonstrate that deterministic foundation models can be efficiently extended into even stronger stochastic models.
\end{abstract}

\begin{center}
  \vspace{-\intextsep}
  \includegraphics[width=1.07\textwidth]{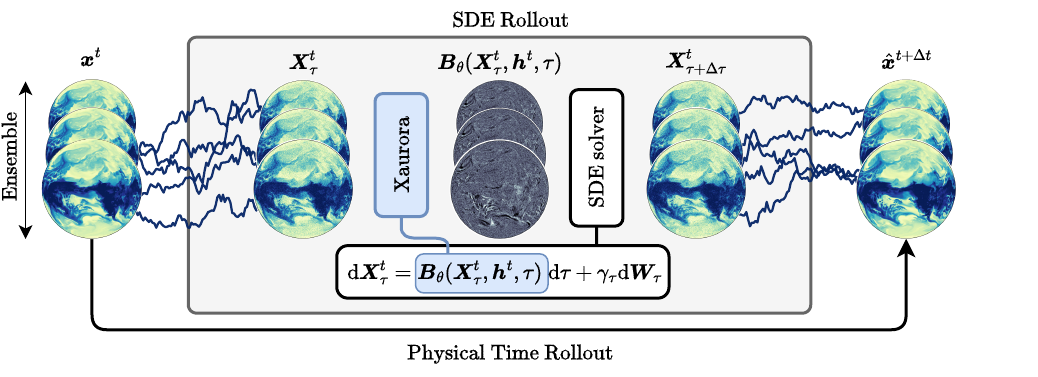}
  \vspace{-1.5\intextsep}
  \captionof{figure}{\textbf{Xaurora produces ensemble forecasts with a nested rollout}. Each physical step $\sistate^{t}\to\gen{\sistate}^{t+\dt}$ is generated by integrating a stochastic differential equation (SDE) over pseudo-time $\tau\in[0,1]$ (SDE time, grey box). Starting from the current state $\SIstate_{0}^t=\sistate^t$, the state $\SIstate^{t}_{\tau}$ follows an SDE formed by the Xaurora drift term ($\drift_\theta$) and a Wiener process ($\bW_\tau$). At $\tau=1$, the terminal SDE state is the forecast, $\gen{\sistate}^{t+\dt}\coloneq\SIstate^{t}_{1}$, and the procedure is repeated autoregressively for the physical time rollout. Ensemble members are generated by sampling different Wiener processes $\bW_{\tau}$.}
  \label{fig:xaurora-overview}
\end{center}

\section{Introduction}\label{sec:introduction}
Accurate weather forecasting is paramount for human activities, from daily life to high-stakes decision making in public safety, agriculture, or transportation. However, the atmosphere is a chaotic system whose predictability is fundamentally limited, with forecast errors growing non-linearly as lead times increase \citep{lorenz_predictability_1969,lorenz_atmospheric_1982}. Therefore, a forecast by itself has limited value without reliable uncertainty quantification, a crucial element for decision making or risk assessment of extreme events.
Hence, weather agencies and decision makers rely on ensemble forecasts, which, however, come with substantial computational cost, limiting their size.

Recently, Deep Learning Weather Prediction (DLWP) models have matched the performance of traditional Numerical Weather Prediction (NWP) models for only a fraction of the inference cost, offering the community a new avenue for forecasting. However, training ensemble DLWP models is computationally very expensive \citep{lang_aifscrps_2024}. On the other hand, deterministic foundation models, such as Aurora \citep{bodnar_foundation_2025}, achieve remarkable forecasting performance with a modest cost, and can be finetuned for different applications. Instead of training stochastic models from scratch, we propose to build generative models using these pretrained deterministic models as priors. Our strategy is motivated by prior work showing that deep neural networks learn generalisable features that can be transferred to related tasks \citep{yosinski_how_2014,devlin_bert_2019,raffel_exploring_2020} and accommodate new parameters \citep{houlsby_parameter-efficient_2019,hu_lora_2021,zhang_adding_2023}.
Capitalising on these concepts, we finetune Aurora into a generative model with only minimal architectural changes.
We call our method \textit{Xaurora}, short for \textit{eXtended Aurora}.

We postulate that state-of-the-art generative methods like Denoising Diffusion Probabilistic Models (DDPM) \citep{ho_denoising_2020} and Flow Matching (FM) \citep{lipman_flow_2023} are ill-suited for our task, since they generate states from pure noise. Instead, we employ Stochastic Interpolants (SI) \citep{albergo_building_2023,albergo_stochastic_2023,chen_probabilistic_2024} that build a stochastic bridge between two consecutive states, which is more suited for finetuning autoregressive state prediction models.
We further introduce a new parameterisation, Denoising Stochastic Interpolants, enabling a warm initialisation from a pretrained deterministic prior. We also propose an efficient rollout training strategy for SDE-based ensemble generative models. Our contributions are as follows:
\begin{enumerate}[leftmargin=*]
    \item \textbf{Denoising Stochastic Interpolant.} A new SI parameterisation: it preserves Aurora's accuracy at initialisation while remaining stable in very high-dimensional pixel space \citep{li_back_2026};
    \item \textbf{Finetuning strategy.} We generalise Aurora's replay buffer-based rollout training approach to the optimisation of generative models, combining SDE rollout with a single-backpropagation Continuous Ranked Probability Score (CRPS) \citep{gneiting_strictly_2007} objective;
    \item \textbf{Performance.} Applied to the Aurora-Small model, the method reaches ensemble mean unbiased RMSE and CRPS close to state-of-the-art (SOTA) operational NWP/DLWP systems, and outperforms all non-operational Aurora models on most variables and lead times despite using 10$\times$ fewer parameters. \chgd{We introduce a sampling-free post-processing method that balances ensemble dispersion and improves probabilistic skill.} \rebuttal{Finally, we show that our model is skilful at predicting extreme temperatures;}
    \item \textbf{Computational cost.} Our finetuning method turns Aurora into a generative model faster than existing probabilistic finetuning approaches \citep{weyn_aurora_2026}. The method only increases the parameter count by 0.64\% and scales to the large version of Aurora. Xaurora generates a 15-day forecast in \chgd{7 minutes} on a single A100 GPU and ensemble members can be parallelised.
\end{enumerate}

\section{Background}\label{sec:background}

\subsection{Problem statement}\label{subsec:background_problem_statement}
\textbf{Notation.} Superscripts denote physical time, subscripts pseudo-time $\tau\in[0,1]$, lowercase bold ($\sistate$) is data or a sample, uppercase bold ($\SIstate_\tau^t$) a stochastic process, and hats ($\gen{\sistate}$) mark generated states.

Let $\sistate^t\in\R^{N_\text{var} \times N_\text{lat} \times N_\text{lon}}$ be the atmospheric state at physical time $t$, where $N_\text{var}, N_\text{lat},$ and $N_\text{lon}$ represent the number of variables, latitude points, and longitude grid points, respectively. Probabilistic weather forecasting aims to sample from $\condprob{\sistate^{t+\dt}}{\bh^t}$, where $\bh^t$ is the history of the most recent states, and $\dt$ denotes the temporal interval.
This is usually applied autoregressively so that each member $(m)$ conditions on its own generated states, $\gen{\sistate}^{t+k\dt,(m)}\sim\condprob{\cdot}{\gen{\bh}^{t+(k-1)\dt,(m)}}$. 

In practice, we follow Aurora \citep{bodnar_foundation_2025} and use $N_\text{var}=69$ variables (see \cref{app:method/data} for a detailed list), as well as a $0.25^{\circ}$ resolution grid (\emph{i.e.} $N_\text{lat}=721$ and $N_\text{lon} = 1440$). The timestep is set to $\dt=\qty{6}{\hour}$, and we define history as the two most recent states, $\bh^t\coloneq(\sistate^{t-\dt},\sistate^{t})$. Additionally, we strictly focus on modelling atmospheric uncertainty and deliberately ignore other sources of uncertainty, such as that of the data assimilation process generating the initial conditions.

\subsection{Stochastic interpolants for autoregressive forecasting}\label{subsec:background_SI}
The Stochastic Interpolant (SI) framework learns a continuous-time generative map between two distributions \citep{albergo_building_2023,albergo_stochastic_2023}. Unlike FM \citep{lipman_flow_2023} or DDPMs \citep{ho_denoising_2020}, SIs do not place a Gaussian assumption on the initial distribution. Setting the endpoints to consecutive atmospheric states therefore turns the interpolant into a stochastic bridge that enables autoregressive forecasting \citep{chen_probabilistic_2024,mucke_physics-aware_2025,andrae_daisi_2025,kossaifi_demystifying_2026} (see \cref{fig:xaurora-overview}). We describe the framework directly in this setting. The interpolant connects $\sistate^t$ and $\sistate^{t+\dt}$ along a path indexed by a pseudo-time $\tau \in [0,1]$,
\begin{align} \label{eq:stochastic_interpolant}
\bI^t_{\tau} = \alpha_\tau \sistate^t + \beta_\tau \sistate^{t+\dt} + \gamma_\tau \bW_\tau,
\end{align}
where $\bW_\tau \sim \mathcal{N}(\bm 0, \tau\Id)$ is a standard Wiener process. During training, we sample $\bW_\tau = \sqrt\tau\bz$ with $\bz\sim\mathcal{N}(\bm 0,\Id)$. The scalar functions satisfy $\alpha_0=\beta_1=1$ and $\alpha_1=\beta_0=\gamma_1=0$, so that $\bI^t_0 = \sistate^t$ and $\bI^t_1 = \sistate^{t+\dt}$. By It\^o's lemma, $\bI^t_\tau$ solves
\begin{align}\label{eq:true_SI_SDE}
\rd \bI^t_{\tau} = \underbrace{\left(\dot{\alpha}_\tau \sistate^t + \dot{\beta}_\tau \sistate^{t+\dt} + \dot{\gamma}_\tau \bW_\tau \right)}_{\eqcolon\bR^t_{\tau}} \rd \tau + \gamma_\tau \rd \bW_\tau.
\end{align}
The drift $\bR^t_\tau$ is not available at inference, as it depends on the target $\sistate^{t+\dt}$. It is therefore approximated by a neural network $\drift_\theta(\cdot, \bh^t, \tau)$, usually trained for drift matching:
\begin{align} \label{eq:SI_drift_loss}
\mathcal{L}_{\drift}(\theta) =  \int_0^1 \E_{(\bh^t,\sistate^{t+\dt})\sim p_\text{data},\bW_\tau\sim\mathcal{N}(\bm{0},\tau\Id)}\left[ \norm{ \drift_{\theta}(\bI^t_\tau, \bh^t, \tau) - \bR^t_\tau}^2_2 \right] \rd \tau.
\end{align}
After training, samples are drawn from $\condprob{\sistate^{t+\dt}}{\bh^t}$ by integrating the SDE below from $\SIstate^t_0 = \sistate^t$ with $\drift_\theta(\cdot, \bh^t, \tau)$ as drift term,
\begin{align}\label{eq:SI_SDE}
\rd \SIstate^t_{\tau} = \drift_\theta(\SIstate^t_\tau, \bh^t, \tau)\, \rd \tau
  + \gamma_\tau \rd \bW_\tau, \quad
\tau \in [0,1], \quad \SIstate^t_0 = \sistate^t,
\end{align}
and the forecast is the terminal SDE state, $\gen{\sistate}^{t+\dt} \coloneq \SIstate^t_1$. Repeating this autoregressively gives the physical-time rollout (as in \cref{fig:xaurora-overview}). The integration scheme and the step count are chosen based on the desired sample fidelity and available compute resources.

Note the distinction between the interpolant ($\bI^t_\tau$) and the SDE state ($\SIstate^t_\tau$). The former is built from the \emph{true} endpoints, which can only be done during training. The latter is generated by the model and fed to the neural network at inference. In practice, in our rollout training algorithm (\cref{subsec:dsi/rb_ft}), we never use the interpolant and train exclusively on generated SDE states.

\section{Related work}\label{sec:related_work}
\textbf{Deterministic DLWP.} The first generation of DLWP models were deterministic \citep{bi_accurate_2023,pathak_fourcastnet_2022,lam_graphcast_2023}. These models are competitive with NWPs and often outperformed them. However, they suffered from smooth predictions (blurring) as lead times increased, due to their deterministic objective \citep{subich_fixing_2025}. Aurora \citep{bodnar_foundation_2025} showed that the vision transformer (ViT) \citep{dosovitskiy_image_2021} could be scaled to a billion-parameter atmospheric foundation model. \cite{yu_scaling_2026} confirmed that scaling laws \citep{kaplan_scaling_2020} also apply to global DLWP models. \citet{bodnar_foundation_2025} and \citet{chen_operational_2025} used a replay buffer to cheaply emulated rollout training of large DLWP models. In contrast to these approaches, we propose a novel generative model, building on Aurora and extending the replay buffer mechanism to SDE-based ensemble models. 

\textbf{Generative DLWP.} Many works have explored generative DLWP. SEEDS \citep{li_seeds_2023} corrected NWP outputs with diffusion at low resolution ($2^\circ$). GenCast \citep{price_gencast_2023} proposed an autoregressive conditional diffusion model for global weather forecasting at $0.25^\circ$ resolution. ArchesWeatherGen \citep{couairon_archesweather_2024} used flow matching to model the residuals of a deterministic model (at $1.5^\circ$ resolution), which was challenging since the residuals have a different distribution on the test set. \rebuttal{\citet{chen_operational_2025} enhanced the prediction of a high-resolution deterministic backbone with a separate non-autoregressive diffusion model.} \citet{mardani_residual_2025} used a similar setup for downscaling (super-resolution). \citet{kossaifi_demystifying_2026} trained a Diffusion Transformer \citep{peebles_scalable_2023} at $0.25^\circ$ resolution using DDPM and SI frameworks with similar performance. \citet{stock_swift_2026} drastically reduced the inference cost of a generative DLWP with distillation. \citet{aich_wind_2026} trained a low-resolution ($1.5^\circ$) diffusion model and implemented posterior sampling for a variety of weather and climate-relevant tasks. Similarly, \citet{andrae_daisi_2025,manshausen_generative_2025,chen_flowdas_2025} explored generative data assimilation. Unlike these methods, we extend and finetune a deterministic foundation model into a generative DLWP model, allowing high-resolution ($0.25^\circ$) for limited training costs.

\textbf{CRPS training.} Another approach to build a probabilistic DLWP consists of optimising ensemble forecasts with a probabilistic loss. \rebuttal{Early works used a latent perturbation model combined with variational autoencoder and a KL-divergence loss \citep{chen_fuxi_2023,chen_machine_2024}.} Recently, supervising ensembles with the CRPS score \citep{gneiting_strictly_2007} has gained popularity. \citet{lang_aifscrps_2024} showed that an ensemble size of 2 was sufficient, roughly doubling the compute and memory requirements compared with deterministic or generative training. CRPS training requires stochastic networks, which is often achieved by injecting noise in FiLM \citep{perez_film_2018} layers \citep{lang_aifscrps_2024,alet_skillful_2025,kossaifi_demystifying_2026}. \citet{cachay_u-cast_2026} showed that dropout \citep{hinton_improving_2012} could also be used as a source of randomness. %

\paragraph{Probabilistic extensions of Aurora.} Two probabilistic extensions of Aurora have recently been proposed \citep{ozdemir_earth_2026,weyn_aurora_2026}. Both models use CRPS training and focus on extending the foundational aspect of Aurora by increasing the number of variables and modalities. Our work provides a complementary perspective exploring generative finetuning.

\section{Denoising Stochastic Interpolants and Xaurora}\label{sec:method}
In this section we introduce a novel generative model, Denoising Stochastic Interpolants (DSI), that is robust in high-dimensional pixel spaces and well-suited for finetuning deterministic priors (\cref{sec:dsi}). Furthermore, we design a probabilistic SDE-rollout training to capture the uncertainty emerging from the forecasts (\cref{subsec:dsi/rb_ft}). Finally, we build \emph{Xaurora} by applying this methodology to an atmospheric foundation model (\cref{subsec:xaurora/denoiser}).

\subsection{Denoising Stochastic Interpolants}\label{sec:dsi}
Diffusion models can be expressed in three equivalent formulations, $\bm{\epsilon}$-, $\sistate$-, and $\bm{v}$-\textit{prediction}, respectively predicting the noise added to the clean signal, the clean signal itself, and the velocity field from the noised signal to the clean signal \citep{ho_denoising_2020, salimans_progressive_2022, li_back_2026}. We show that the same reparameterisation trick can be applied to SIs. Namely, we define $\bW$-, $\sistate_1$- and $\drift$-prediction as analogues of the diffusion parameterisations (more details in \cref{app:method/drift_param}). $\sistate_1$-prediction is the SI equivalent of $\sistate$-prediction, which in our temporal setting corresponds to predicting $\sistate^{t+\dt}$. In diffusion, this parameterisation is often referred to as \textit{denoising}, hence we call this method \textit{Denoising Stochastic Interpolants} (DSI).

Let $\bD_\theta$ be a \textit{denoiser} model predicting the terminal state $\sistate^{t+\dt}$ from an arbitrary SDE state $\SIstate^t_\tau$,
\begin{equation}\label{eq:dsi/denoiser}
    \bD_\theta(\SIstate^t_\tau,\bh^t,\tau) \approx \E[\sistate^{t+\dt}\mid \SIstate^t_\tau,\bh^t].
\end{equation}
The drift in \cref{eq:SI_SDE} can be rewritten as (see \cref{app:method/drift_param} for derivation details)
\begin{align}\label{eq:dsi/drift_from_denoiser}
\drift_\theta(\SIstate^t_\tau,\bh^t,\tau) = \dot \alpha_\tau \sistate^t + \dot \beta_\tau \bD_\theta(\SIstate^t_\tau,\bh^t,\tau) + \frac{\dot \gamma_\tau}{\gamma_\tau}\big[\SIstate^t_\tau - \alpha_\tau \sistate^t - \beta_\tau \bD_\theta(\SIstate^t_\tau,\bh^t,\tau)\big],
\end{align}
with $\sistate^t$ becoming a generated state when performing autoregressive rollouts ($\gen{\sistate}^{t+k\dt}$, $k>0$). During training, \cref{eq:SI_drift_loss} is replaced by the corresponding denoising loss,
\begin{equation}\label{eq:dsi/xloss}
\mathcal{L}_{\bD}(\theta) = \int_0^1 \E_{(\bh^t,\sistate^{t+\dt})\sim p_\text{data},\bW_\tau\sim\mathcal{N}(\bm{0},\tau\Id)}\big[\norm{\bD_\theta(\bI^t_\tau,\bh^t,\tau)-\sistate^{t+\dt}}_2^2\big]\rd\tau.
\end{equation} 

\chgd{Since we use $\gamma_\tau=1-\tau$, we clip the denominator of the last term to $0.05$ to avoid instabilities at $\tau=1$ in \cref{eq:dsi/drift_from_denoiser}. This results in some leftover additive noise during training, which we found useful for the model. At inference, we use $S=10$ uniform Euler--Maruyama steps, so \cref{eq:dsi/drift_from_denoiser} remains stable throughout the integration. The final SDE state carries residual noise from the last Wiener increment $\gamma_{1-1/S}\Delta\bm{w}_{1-1/S}$. We record the forecasts with this increment subtracted, and pass the noisy state to the next physical step so that ensemble members keep diverging, as during training. Removing
residual noise after sampling is not new: in image generation, a final denoising step substantially improves FID without visibly changing the
samples \citep{jolicoeur-martineau2021adversarial,song2020improved}. We discuss the details further in \cref{app:sde_rollout_rb_algo,app:method/eval/forecast_settings,app:results/importance_subtract_noise}.} 

Inspired by \cite{li_back_2026}, we \rebuttal{establish that DSI can be implicitly trained in $\drift$- and $\bW$-space without output transformation by weighting the denoising loss} (more details in \cref{app:method/drift_param}).

\subsection{Denoising Stochastic Interpolants from a Pretrained Prior}\label{subsec:dsi/dsi_from_pretrained}
\begin{figure}[t]
    \centering
    \begin{minipage}{0.48\textwidth}
        \vspace{-1.2\intextsep}
        
        \begin{adjustbox}{width=1.0\textwidth}
            \includegraphics[width=1\linewidth]{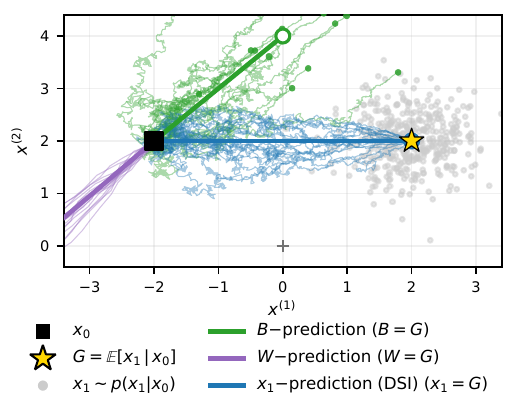}
        \end{adjustbox}
    \end{minipage}%
    \hfill
    \begin{minipage}{0.50\textwidth}
    \caption{\textbf{2D toy example: DSI preserves an ideal prior.} In each parameterisation (\cref{tab:drift_param}), we replace the network with an ideal prior $\bm{G}(\sistate_0)=\E[\sistate_1\mid\sistate_0]$ and integrate \cref{eq:SI_SDE} from a
    common initial state (black square), no training. \rebuttal{Without output transformation, }DSI recovers the true mean (gold star) as its network models $\E[\sistate_1\mid\sistate_0]$, exactly what $\bm{G}$ computes. The deterministic prior has no spread, hence the SDE trajectories (thin lines, thick line is the average) land on the expectation. Hence, finetuning (\cref{subsec:dsi/dsi_from_pretrained}) aims to capture the spread and maintain the skill. We design our algorithm (\cref{subsec:dsi/rb_ft}) for that purpose and apply it to an atmospheric foundation model (\cref{subsec:xaurora/denoiser}).}
    \label{fig:dsi/toy_example}
    \end{minipage}
    \vspace{-2.\intextsep}
\end{figure}

Let $\prior$ be a pretrained deterministic prior whose parameters $\phi$ are a subset of $\theta$, such that
\begin{align}\label{eq:dsi/G(x0)}
    \prior(\bh^t) \approx \E[\sistate^{t+\dt}\mid\bh^t].
\end{align}
Our objective is to finetune a stochastic interpolant model from this prior, providing a warm-start to the optimisation. In this context, \rebuttal{DSI is the only parameterisation in which the prior can be used without output transformation}. We assess this intuition with a toy example shown in \cref{fig:dsi/toy_example}. The mathematical motivation for the general case in arbitrary dimensions is presented in \cref{app:toy_maths}.

\textbf{Warm start.} In \cref{fig:dsi/toy_example}, DSI reaches the correct endpoint \rebuttal{without any finetuning or output transformation}. That is, initialising the denoiser such that $\bD_{\theta^\text{init}}\equiv\prior$ (where $\equiv$ denotes functional equivalence) fully maintains the capabilities of the pretrained prior, while they are partially lost with the \rebuttal{direct} $\drift$- and $\bW$-prediction parameterisations. Based on \cref{app:drift_param/beta=tau}, we use $\beta_\tau=\tau$. 

\textbf{Consistent network target.} In DSI, the denoiser and the pretrained prior are trained against the same kind of target, a conditional expectation of the terminal state $\sistate^{t+\dt}$. This is not the case for the other parameterisations whose targets typically involve a combination of states and random fields, which introduces new modalities to the underlying model, potentially degrading convergence.

\textbf{Curse of dimensionality.} \cite{li_back_2026} showed that $\sistate$-prediction was preferable for diffusion in high-dimensional spaces. That is because predicting noise ($\bm{\epsilon}$) or noisy quantities ($\bm{v}$) is subject to the \textit{curse of dimensionality}. In contrast, $\sistate$-prediction remains on the data manifold, achieving higher performance in those cases. Given the very high dimensionality of atmospheric states, we hypothesise that these conclusions hold.

As long as $\bD_{\theta^\text{init}}\equiv\prior$ (functionally equivalent), these observations carry over to the realistic scenario from \cref{eq:dsi/G(x0)}. In that case, DSI finetuning serves two purposes: (1) modelling the spread around the prior and (2) surpassing the skill of the prior through ensemble predictions.

For completeness, we note that in the case of residual prediction, \emph{i.e.} $\prior(\bh^t) \approx \E[\sistate^{t+\dt}-\sistate^{t}\mid\bh^t]$, another common formulation in autoregressive state prediction, $\drift$-prediction, is likely preferable (see \cref{app:toy_maths}). Nevertheless, the network still targets a noisy quantity and could be less robust to the curse of dimensionality, following the arguments of \citet{li_back_2026}.

\subsection{SDE-Rollout Replay Buffer Finetuning with Probabilistic Loss}\label{subsec:dsi/rb_ft}
\begin{figure}[t]
    \centering
    \begin{minipage}{0.47\textwidth}
        \vspace{-0.5\intextsep}
        \begin{adjustbox}{trim=12 0 0 0, clip, width=1.\textwidth}
            \includegraphics[width=1\linewidth]{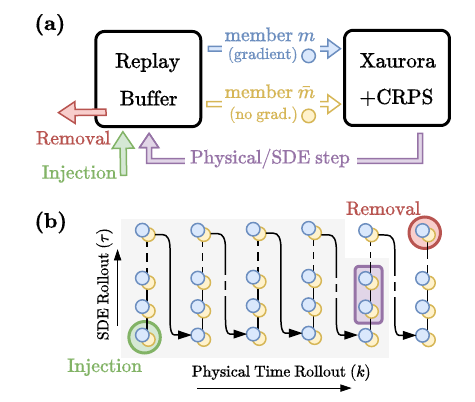}
        \end{adjustbox}
    \end{minipage}%
    \hfill
    \begin{minipage}{0.52\textwidth}
    \caption{\textbf{SDE-rollout training with single-backpropagation CRPS loss.} \textbf{(a)} Two members are sampled from the buffer and fed to the network, only member $m$ is backpropagated. Both advance one step (physical step if $\tau$ reaches 1, SDE step otherwise), and are pushed back. Fresh samples are randomly injected, old (FIFO) or finished rollouts are removed. \textbf{(b)} The buffer tracks pairs of members (yellow and blue circles) throughout nested rollouts ($\tau,k$). Both members start from the same initial condition, but receive different noise samples. A training iteration costs the same at every depth (2 forward, 1 backward, purple box), unlike regular rollout training which requires full nested rollouts (shown in grey).}
    \label{fig:replay_buffer_viz}
    \end{minipage}
    \vspace{-2.1\intextsep}
\end{figure}

Rollout training, or training a model on its own autoregressive forecasts, is crucial for long-term performance in DLWP \citep{lam_graphcast_2023,lang_aifs_2024,lang_aifscrps_2024}. However, traditional approaches \citep{lam_graphcast_2023} are not tractable with generative models, as each physical step involves solving an entire SDE (nested rollout, see \cref{fig:xaurora-overview}). We solve this by blending the physical and SDE rollouts together. We take inspiration from \citet{bodnar_foundation_2025} and \citet{chen_operational_2025}, who emulate rollout training using a replay buffer. We extend this procedure to nested rollouts, enabling rollout training with negligible computational overhead compared to single-step optimisation.

In the original SI \citep{albergo_building_2023}, the drift approximates $\E[\bR_\tau^t\mid\SIstate_\tau^t,\bh^t]$ and solving the SDE (\cref{eq:SI_SDE}) recovers samples from $\condprob{\sistate^{t+\dt}}{\bh^t}$. We instead propose to train a sampler end-to-end by directly optimising SDE rollouts with a proper score, the fair CRPS loss \citep{lang_aifscrps_2024}. This results in a different minimiser, corresponding to a probability flow with increased variance. Since our ensemble is underdispersive (\cref{subsec:experiments/mwf}), this is generally beneficial to performance (see ablation experiments in \cref{subsec:experiments/ablations}). Given the ground truth $\sistate^{t+\dt}$, a pair of members $(\gen{\sistate}^{(m)}, \gen{\sistate}^{(\bar{m})})$, and $\norm{\cdot}_\omega$ as the variable-weighted $\ell_1$-norm (with weights $\omega$), our fair CRPS loss is given by:
\begin{equation}\label{eq:fair_crps_loss}
    \ell_\text{fCRPS}(\sistate^{t+\dt}, \gen{\sistate}^{(m)}, \gen{\sistate}^{(\bar{m})} ; \theta) \coloneq%
    \underbrace{\tfrac12\left(\sum_{i\in\{m,\bar{m}\}}\norm{\gen{\sistate}^{(i)}-\sistate^{t+\dt}}_\omega\right)}_{\text{skill term } (\downarrow)} -\underbrace{\tfrac12\norm{\gen{\sistate}^{(m)}-\gen{\sistate}^{(\bar m)}}_\omega}_{\text{spread term }(\uparrow)}.
\end{equation}
The first term is the \emph{skill term} that measures the prediction error and is equivalent to the sum of the MAE of two member. The second term is the \emph{spread term} and measures the dispersion across samples.
Unlike other methods \citep{lang_aifscrps_2024,alet_skillful_2025,kossaifi_demystifying_2026,cachay_u-cast_2026}, we propose to supervise only one randomly selected member when computing the loss, resulting in the same stationary points while minimising the computational cost. The process is visualised in \cref{fig:replay_buffer_viz}, and more details are given in \cref{app:method/dsi}. We finetuned Xaurora for about 13\% of the samples used to pretrain Aurora, which took 42 hours on 16 A100 GPUs, $12--16\times$ fewer GPU hours than the probabilistic Aurora model by \citet{weyn_aurora_2026}.

\subsection{Denoiser model}\label{subsec:xaurora/denoiser}
The denoiser of Xaurora is based on Aurora \citep{bodnar_foundation_2025}, a SOTA deterministic atmospheric foundation model. By initialising $\prior$ to Aurora-S (the small version of Aurora), the approach discussed in \cref{subsec:dsi/dsi_from_pretrained} can be used for weather forecasting. As Aurora predicts $\sistate^{t+\dt}$ from $\bh^t$, we make minimal architectural changes (see \cref{fig:xaurora/denoiser_arch} in \cref{app:method/xaurora}) to accommodate the pseudo-time $\tau$ and the SDE state $\SIstate^t_\tau$ according to \cref{eq:dsi/denoiser}. These changes increase the parameter count by 0.64\% and ensure $\bD_{\theta^\text{init}}\equiv\prior$ with \textit{zero-initialisation} \citep{zhang_adding_2023}.

\textbf{Conditioning on $\SIstate^t_\tau$.} We create new surface and atmospheric patch embedding layers in the encoder dedicated to the SDE state. We sum these embeddings with those of $\bh^t$.

\textbf{Conditioning on $\tau$.} The pseudo-time is embedded with a Fourier expansion (\cref{eq:fourier_expansion_tau}) and a zero-initialised MLP. The resulting vector is added to the existing lead time embedding and fed to AdaLN layers \citep{perez_film_2018,peebles_scalable_2023} in the Swin Transformer \citep{liu_swin_2021}.

\textbf{LoRA.} The Swin Transformer's weights are frozen and adapted using LoRA \citep{hu_lora_2021}.

\textbf{Network preconditioning.} We adapt the preconditioning from \cite{karras_elucidating_2022} to DSI. This ensures that the model inputs, particularly $\SIstate^t_\tau$, are approximately zero-mean unit-variance on average. This is entirely parameter-free (more details in \cref{app:method/xaurora/norm_precond}).

\section{Experiments}\label{sec:experiments}

\subsection{Medium-range weather forecasting}\label{subsec:experiments/mwf}
\begin{figure}
    \centering
    \vspace{-0.5\intextsep}
    \includegraphics[width=1\linewidth]{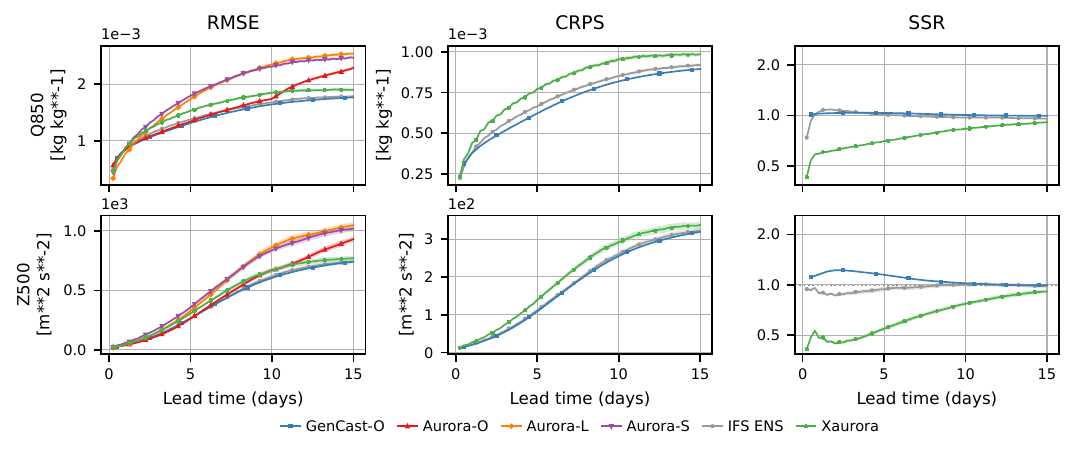}
    \vspace{-1\intextsep}
    \caption{\chgd{\textbf{Global medium-range forecast skill in 2022.} From left to right, columns represent RMSE (uRMSE for ensemble models), fair CRPS (ensemble only), and SSR (ensemble only). Rows represent different variables; Q850 is a proxy for mesoscale variability (small scale), Z500 for large-scale circulation and synoptic-scale variability. Xaurora's ensemble mean unbiased RMSE outperforms Aurora-S/L on most lead times and is competitive with operational models. \rebuttal{The shading indicates the 95\% bootstrap intervals over initialisations.} All models are well calibrated at 15 days (SSR$\approx$1). Xaurora's ensemble is generally underdispersive (SSR$<$1).}}
    \vspace{-1\intextsep}
    \label{fig:headline_metrics_psd}
\end{figure}
We use WeatherBench \citep{rasp_weatherbench_2024} to benchmark Xaurora against SOTA ensemble systems, using the year 2022 as the test set. \chgd{The Aurora prior enables fast sampling with a few SDE steps, namely 10 Euler--Maruyama integration steps. We remove the residual noise from the last SDE iteration by subtracting the corresponding Wiener increment (more details in \cref{app:method/eval/forecast_settings,app:results/importance_subtract_noise}).} Metrics are defined in \cref{app:method/eval/metrics}. We compare Xaurora against the three Aurora versions: Small (Aurora-S), Large (Aurora-L), and Operational (Aurora-O). Furthermore, we compare against operational GenCast\footnote{Among the ensemble DLWP models available on WeatherBench, we choose GenCast because it is widely considered the reference in probabilistic and generative DLWP.} (GenCast-O; \citealp{price_gencast_2023}), which serves to represent DLWP, and the SOTA NWP system, IFS ENS. \rebuttal{All baselines in this section are evaluated on the same initialisations and ensemble size as Xaurora, except GenCast, whose forecasts are not publicly available.}

\textbf{Qualitative results.} Xaurora produces high-quality forecasts that are realistic compared to ERA5 samples (see \cref{fig:mosaic_3d,fig:mosaic_10d,fig:mosaic_15d}). The model generates multi-scale features, from planetary waves in geopotential fields (Z) to grid-scale structures in the moisture fields (Q). After 3 days, individual Xaurora members remain close to Aurora-S, whereas Xaurora diverges from Aurora-S after 10 days, indicating that Xaurora benefits from the prior without being overly constrained.

\textbf{Ensemble mean performance.} 
The RMSE panels in \cref{fig:headline_metrics_psd} show that the ensemble mean of Xaurora is close to that of the operational ensemble forecasting systems (IFS ENS, Aurora-O, GenCast-O). The performance gap is minimal at initialisation and the error curves follow similar trends. Xaurora also outperforms its prior, Aurora-S, beyond 1 day and Aurora-L beyond 3 days (see also \cref{fig:xaurora_vs_aurora_s_scorecard,fig:xaurora_vs_aurora_l_scorecard}). Beyond 7 to 12 days, Xaurora achieves a lower error than Aurora-O, which has only been finetuned for 10-day rollouts. The error reduction over Aurora models cannot be entirely attributed to our method since averaging ensembles provides a statistical advantage (which is also a motivation for using ensembles). An ensemble size of 2 is already sufficient to outperform the prior (see \cref{app:results/ablations/ens_size}). More RMSE curves can be found in \cref{fig:all_rmse}.

\chgd{\textbf{Probabilistic performance.} 
The CRPS scores of Xaurora closely follow those of the operational models (see \cref{fig:headline_metrics_psd,fig:all_crps}). Note that we use the fair CRPS since the ensemble sizes differ between systems. Overall, our ensemble is underdispersive (SSR$<$1), especially on short lead times (see \cref{fig:headline_metrics_psd,fig:all_ssr}). Unlike GenCast-O, Xaurora does not take into account initial condition uncertainty, which could improve short-term calibration. Finally, \cref{fig:mc_drop_crps_scorecard} shows that Xaurora consistently outperforms a naive probabilistic version of Aurora.}

\chgd{\subsection{Spectral post-processing}\label{subsec:experiments/postprocessing}}
\begin{figure}%
    \centering
    \vspace{-0.5\intextsep}
    \includegraphics[width=1\linewidth]{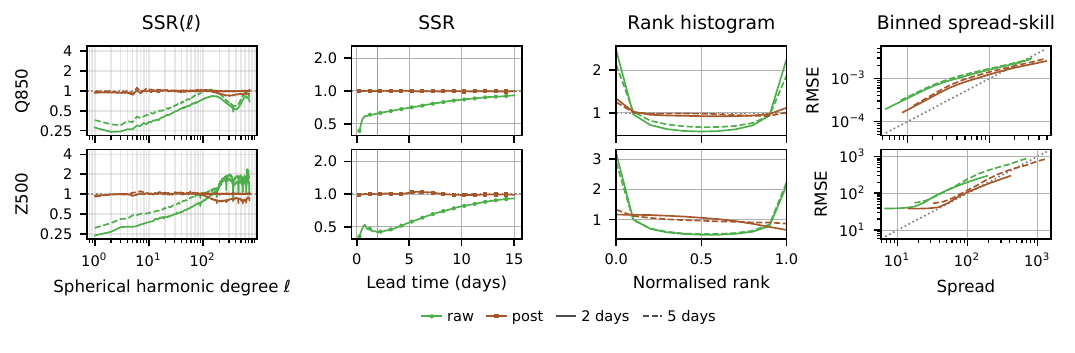}
    \vspace{-1.\intextsep}
    \caption{\chgd{\textbf{Spectral post-processing of Xaurora forecasts.} From left to right, columns represent SSR decomposed over spherical harmonics degrees, global SSR up to 15 days, rank histogram (flatter is better, underdispersion is $\cup$-shaped, overdispersion $\cap$-shaped), and binned spread-skill (dotted identity line is ideal, over the line means overconfident at a given spread, below the line underconfident at a given spread). Raw forecasts are shown in green, post-processed (post) in brown. Apart from the global SSR panel, solid line shows 2-day forecasts, dashed 5-day. Post-processing succeeds in calibrating the ensemble but does not structurally change the forecasts.}}
    \label{fig:spectral_postprocessing_headline}
\end{figure}

\chgd{A spherical harmonics decomposition of the SSR highlights a consistent per-variable spectral shape with a level that varies with lead time (see $\text{SSR}({\ell})$ panels in \cref{fig:spectral_postprocessing_headline} and \cref{fig:ssr_l_more_vars} for more variables). In general, the large scales are more underdispersive than small scales. This motivates a simple, yet effective post-processing approach where a spectral shape function ($\mathcal{G}(\ell)$) with a lead time correction ($\mathcal{H}(s)$) is applied to existing forecasts such that the SSR is equal to 1 on average. With $\gen{\sistate}(s)$ and $\bar{\sistate}(s)$ denoting an ensemble member and the ensemble mean at rollout step $s$, respectively, we calibrate the deviation from the ensemble mean as follows:}
\begin{equation}\label{eq:post_processing_core}
    \gen{\sistate}^\text{cal}(s) = \bar{\sistate}(s) + \text{SHT}^{-1}[\underbrace{\mathcal{G}(\ell)}_{\text{Spectral shape}}\cdot \underbrace{\mathcal{H}(s)}_{\text{Per-lead level}}\cdot \text{SHT}(\gen{\sistate}(s)-\bar{\sistate}(s))],
\end{equation}
\chgd{where $\text{SHT}$ is the spherical harmonics transform \citep{kurth2026library}. We fit these functions on the validation set (2021) and evaluate on the test set (2022). Details can be found in \cref{app:results/post_processing/procedure}.}

\chgd{\cref{fig:spectral_postprocessing_headline} shows that post-processing significantly improves calibration. The SSR panels confirm the agreement between spread and error across scales and lead times. Significant CRPS improvements (up to 15\%) are also observed across variables (see \cref{app:results/post_processing/skill_improvement}). The rank histograms of the raw forecasts indicate underdispersion, whereas those of the post-processed predictions correspond to a well-calibrated ensemble, with a slight bias (tilt) for Z500. After post-processing, the skill-spread curves are closer to the identity line, highlighting overall calibration improvements.}

\chgd{Post-processing cannot structurally change the forecasts. The method is unaware of location and flow, and leaves the ensemble mean unchanged. The correction is therefore largest where the members already disagree, and it cannot generate new dynamical structures. Hence, post-processed members closely resemble the raw ones and show the same synoptic features with modified amplitudes (see \cref{app:results/postprocessing/effect_on_members}). This explains why the binned spread-skill curves are shifted to the right but keep their slope, including the overconfidence in low-spread regions. \cref{app:results/spectral_analysis} briefly explores replacing white noise with structured noise (\emph{e.g.} red noise) during sampling. Despite higher computational costs, this approach can effectively reshape the spread in the spectral domain and could be a promising alternative to residual noise subtraction and post-processing.}

\subsection{Forecasting extreme temperatures}\label{subsec:experiments/extreme}
\rebuttal{Temperature extremes are among the most damaging weather hazards. Heat caused about 61,000 deaths in Europe in summer 2022 \citep{ballester_heat-related_2023}. Cold increases the risk of mortality \citep{guo_large_2011} and can disrupt energy infrastructure \citep{doss-gollin_how_2021}. Definitions of hot and cold extremes and the evaluation protocol can be found in \cref{app:results/extremes/def}.}

\begin{figure}
    \centering
    \vspace{-0.5\intextsep}
    \includegraphics[width=\linewidth]{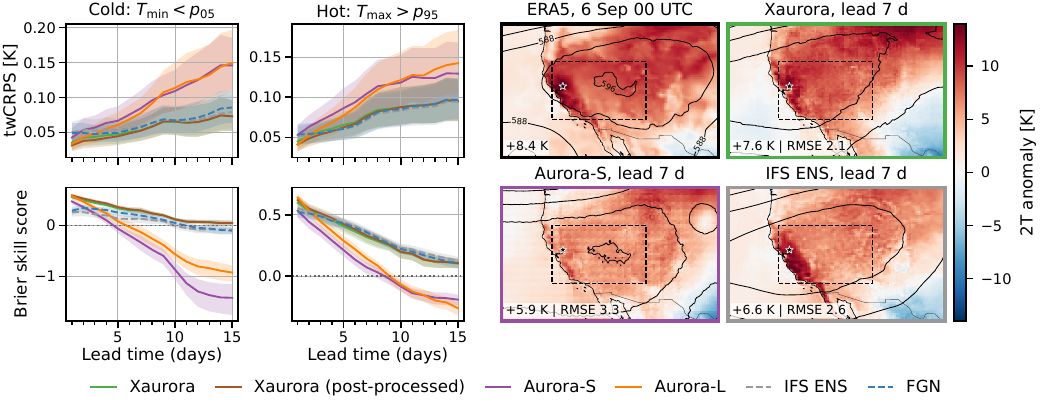}
    \vspace{-1.2\intextsep}
    \caption{\rebuttal{\textbf{Extreme temperature forecasting.} \textbf{Left:} threshold-weighted CRPS and Brier skill score (higher is better, the dotted line is the climatology) of daily cold and hot extremes for Xaurora, Aurora, and operational models. The shading indicates the 95\% intervals over initialisations. Dashed lines highlight operational models evaluated on ERA5; thus, model comparison should be handled with caution. \textbf{Right:} Forecast of the September 2022 western US heatwave. Colours show 2T anomaly, contours Z500. The star is Sacramento, where the maximum temperature was recorded ($\qty{46.7}{\degreeCelsius}$). Maps show ERA5 and forecasts initialised a week earlier for Xaurora and IFS ENS ensemble means, as well as Aurora-S. Panel text metrics are computed within the dashed box.}}
    \label{fig:extremes_compact}
\end{figure}

\rebuttal{In \cref{fig:extremes_compact}, Aurora models become worse than climatology for cold extremes at about 5 days and for hot extremes at about 9 days, whereas Xaurora stays skilful up to 15 days. Furthermore, \cref{fig:extremes_lead} shows that Xaurora is more reliable than Aurora-S on both extremes. The underdispersion of the ensemble affects reliability on short lead times, which is substantially improved by post-processing, most significantly for cold extremes. Comparison with ERA5 suggests that Xaurora could forecast cold extremes better than operational models and is competitive on hot extremes.}

\rebuttal{\textbf{Case studies.} The maps in \cref{fig:extremes_compact} confirm the results of the threshold-based analysis. A week ahead, Xaurora captures the event substantially better than Aurora-S and is comparable with IFS ENS. \cref{fig:uswest_heat_mosaic} shows additional lead times and includes FGN fields \citep{alet_skillful_2025}. Similar observations can be made on the September 2022 heatwave in southern France (\cref{fig:swfrance_heat_mosaic}).}

\subsection{Ablations}\label{subsec:experiments/ablations}
\begin{table}
\centering
\caption{\chgd{\textbf{Ablation results.} Skill scores against Xaurora, $100\cdot(1 - \mathcal{M}_\text{model} / \mathcal{M}_{\text{Xaurora}})$: positive values are better than the reference. The best values are underlined. Methods are detailed in \cref{app:results/ablations/main_paper_details}. $\dagger$ marks differences that are not significant at the 95\% level (see \cref{app:method/eval/significance}). Absolute values for Q850 and Z500 expressed in \unit{\gram\per\kilo\gram} and \unit{\meter\squared\per\second\squared} respectively.}}
\label{tab:ablations}
\footnotesize
\begin{tabularx}{\linewidth}{ll>{\centering\arraybackslash}X>{\centering\arraybackslash}X>{\centering\arraybackslash}X>{\centering\arraybackslash}X}
\toprule
 &  & \multicolumn{2}{c}{\textbf{5-day uRMSE}} & \multicolumn{2}{c}{\textbf{5-day CRPS}} \\
\cmidrule(lr){3-4} \cmidrule(lr){5-6}
 &  & Z500 & Q850 & Z500 & Q850\\
\midrule
\multicolumn{2}{l}{\textbf{Xaurora} (Replay buf. + CRPS + Aurora-S)} & $\phantom{{+}}305.8\phantom{\%^{\dagger}}$ & $\phantom{{+}0}1.53\phantom{\%}$ & $\phantom{{+}0}\underline{139.4}\phantom{\%}$ & $\phantom{{+}00}\underline{0.773}\phantom{\%}$ \\
\multicolumn{2}{l}{Aurora-S} & $\phantom{0}{-}16.9\%\phantom{^{\dagger}}$ & ${-}17.1\%\phantom{0}$ & \multicolumn{2}{c}{\textit{Deterministic}} \\
\midrule
\multirow{3}{*}{Optim.} & Single-step + MAE & $\phantom{00}{-}2.9\%^{\dagger}$ & $\phantom{0}{-}1.9\%\phantom{0}$ & $\phantom{00}{-}11.3\%$ & $\phantom{0}{-}10.4\%\phantom{00}$ \\
 & Curriculum + MAE & $\phantom{00}\underline{{+}2.2\%}\phantom{^{\dagger}}$ & $\phantom{0}\underline{{+}2.2\%}\phantom{0}$ & $\phantom{00}{-}10.2\%$ & $\phantom{0}{-}11.1\%\phantom{00}$ \\
 & Replay buf. + MAE & $\phantom{00}{-}3.1\%^{\dagger}$ & $\phantom{0}{-}3.8\%\phantom{0}$ & $\phantom{00}{-}24.7\%$ & $\phantom{0}{-}24.4\%\phantom{00}$ \\
\midrule
\multirow{3}{*}{Param.} & $\sistate_1$-pred. + DDPM & $\phantom{0}{-}94.7\%\phantom{^{\dagger}}$ & ${-}32.0\%\phantom{0}$ & $\phantom{0}{-}159.9\%$ & $\phantom{0}{-}48.8\%\phantom{00}$ \\
 & $\drift$-pred. & $\phantom{0}{-}87.0\%\phantom{^{\dagger}}$ & ${-}27.3\%\phantom{0}$ & ${-}1421.0\%$ & ${-}210.8\%\phantom{00}$ \\
 & $\bW$-pred. & \multicolumn{4}{c}{\textit{Training diverges}} \\
\midrule\midrule
\multicolumn{2}{l}{\textbf{Xaurora} (Single-step + MAE + {Aurora-L})} & $\phantom{{+}}\underline{311.2}\phantom{\%^{\dagger}}$ & $\phantom{{+}0}\underline{1.60}\phantom{\%}$ & $\phantom{{+}0}164.9\phantom{\%}$ & $\phantom{{+}00}0.845\phantom{\%}$ \\
\multicolumn{2}{l}{Aurora-L} & $\phantom{00}{-}5.3\%\phantom{^{\dagger}}$ & $\phantom{0}{-}8.6\%\phantom{0}$ & \multicolumn{2}{c}{\textit{Deterministic}} \\
\bottomrule
\end{tabularx}
\end{table}

We conduct a suite of ablation studies to measure the benefits of our optimisation method, the DSI parameterisation, and whether Xaurora scales to using Aurora-L as the prior. The results confirm the effectiveness of our method for probabilistic weather forecasting and are summarised in \cref{tab:ablations}.

\textbf{Optimisation.} \chgd{Training with MAE results in poorer probabilistic metrics (up to 24.7\%). Interestingly, the 5-day uRMSE is generally worse with a deterministic loss, except for the curriculum approach which outperforms Xaurora by 2.2\%. Since the main aim of Xaurora is ensemble forecasting, we prioritise CRPS, which favours our design.}

\textbf{Parameterisation.} \rebuttal{Without output transformation, }$\drift$- and $\bW$-pred cannot be warm-started (see the toy example in \cref{fig:dsi/toy_example}) and perform poorly ($\bW$-pred diverges). The real competitor is $\sistate_1$-pred + DDPM, which also massively underperforms. \chgd{This could be explained by the 10 integration steps used in the rollout training algorithm, which might not be sufficient with DDPM since it starts from pure noise. In contrast, the stochastic bridge of the DSI supports the efficiency of our setup.}

\textbf{Prior.} Initialising the denoiser to Aurora-L reproduces the core result, \emph{i.e.} the ensemble outperforms the prior, even when training with Single-step + MAE and without having fully converged. Based on the optimisation results, we expect significantly better results should the training be scaled and made probabilistic. This was not pursued further due to the high computational cost. 

\rebuttal{\textbf{Non-SDE baseline.} \cref{app:results/ablations/non-sde} compares Xaurora with a non-SDE baseline. Consistent with the literature \citep{kossaifi_demystifying_2026}, the results are comparable, with a small but significant advantage for Xaurora on longer lead times. In general, the added value of generative models over single-step approaches is flexibility at inference: the sampler can be adapted, the noise can be tuned, the model can be guided. In contrast, once a single-step model is trained, little can be changed.}

\section{Conclusion}
In this work, we introduced Denoising Stochastic Interpolants (DSI), a new generative model, and a probabilistic rollout training algorithm for SDE-based autoregressive state prediction. This setup enables warm starts from pretrained deterministic priors in high-dimensional settings. With these methods, we build Xaurora by efficiently finetuning Aurora, a deterministic atmospheric foundation model. Our 15-day ensemble forecasts are generated in \chgd{7 minutes}, approach the performance of SOTA probabilistic weather forecasting systems, and predict extreme temperatures skilfully.

\textbf{Limitations and future work.} \chgd{Following GenCast, the underdispersion could partially be addressed with ERA5 initial condition uncertainty. Furthermore, our preliminary results illustrate that calibration could be improved with spectral interventions on the injected noise.} A more ambitious direction would be to learn the diffusion along with the denoiser. Other relevant directions include distillation for sampling efficiency \citep{song_consistency_2023,boffi_flow_2024,stock_swift_2026} and physics-constrained modelling \citep{jacobsen_cocogen_2024,cheng_gradient-free_2025,utkarsh_physics-constrained_2025}.

\textbf{Broader impact.} Finetuning Xaurora and generating forecasts has a relatively low computational cost, making skilful probabilistic weather forecasting more accessible to institutes with limited compute (\emph{e.g.}  academia or national meteorological offices). Our method is also scalable to larger models or output spaces and is expected to transfer to other problems beyond weather forecasting. 

\subsection*{AI use statement}
This work used generative AI to help develop theoretical models or conceptual frameworks, formulate mathematical claims, propose or refine hypotheses, design or provide feedback on research methodology or experiments, implement methods, and interpret results. Furthermore, generative AI was used to create or modify scientific figures or images, suggest experimental parameters, create or edit software code, as well as general brainstorming. The LLM-generated code, interpretations, and suggestions were always subjected to human verification.

AI was not used to generate synthetic data sets, provide critical ingredients for proving mathematical claims, assist in the writing of proofs, assist with translation, clean and reformat dataset, or support qualitative and thematic data analysis.

\subsection*{Ethics statement}
This work uses only publicly available reanalysis and operational forecast data
(ERA5, HRES-T0 and IFS ENS) and involves no human subjects or personally
identifiable information. By turning an existing deterministic foundation model
into a probabilistic ensemble model at a small fraction of the cost of training a
generative forecaster from scratch, our method lowers the barrier to ensemble
forecasting for institutions without large compute budgets, which we consider its
principal societal benefit. Xaurora is nonetheless a research prototype: it
inherits the biases and uneven observational coverage of its ERA5 training data,
so its skill is likely lower over data-sparse regions, and it remains
underdispersive at short lead times for some variables, and it should therefore
not be used for operational warnings or other high-stakes decisions without
independent validation by forecasters.

\subsection*{Reproducibility statement}
This work is reproducible. We use an appropriate software stack for version control, experiment logging, and configuration tracking. The source code will be made available upon acceptance. All data used in this study were downloaded from WeatherBench. This represents tens of terabytes of storage.

\subsubsection*{Acknowledgments}
This work was funded by the European Union's Horizon 2020 research and innovation programme under grant agreement No 101081276. This work used the Dutch national e-infrastructure with the support of the SURF Cooperative using grant no. NWO-2025.042.

\bibliography{references}
\bibliographystyle{iclr2027_conference}

\newpage
\appendix
\etocdepthtag.toc{appendix}%

\begingroup
  \etocsettagdepth{main}{none}%
  \etocsettagdepth{appendix}{subsubsection}%
  \etocsettocstyle{\section*{Appendix contents}}{}
  \tableofcontents
\endgroup
\newpage

\crefname{appendix}{Appendix}{Appendices}
\Crefname{appendix}{Appendix}{Appendices}
\crefalias{section}{appendix}
\crefalias{subsection}{appendix}
\crefalias{subsubsection}{appendix}

\section{Methodological details}\label{app:method}

\subsection{Data}\label{app:method/data}
\subsubsection{Datasets}\label{app:method/data/datasets}
\textbf{ERA5.} The global climate reanalysis dataset generated by ECMWF \citep{hersbach_era5_2020} by combining historical observations with physical models. It spans 1940 to the present at hourly resolution on a 0.25$^\circ$ grid, contains 37 pressure levels from the surface to the top of the atmosphere and includes a broad range of atmospheric, surface, and oceanic variables. In this study, we use a subset of ERA5 from WeatherBench2 \citep{rasp_weatherbench_2024} which is downsampled to 6-hourly resolution and has a reduced set of variables and pressure levels\footnote{Stored on Google cloud at \url{gs://weatherbench2/datasets/era5/1959-2023\_01\_10-wb13-6h-1440x721.zarr}} that are listed in \cref{tab:list_of_variables}. 

\textbf{HRES-T0.} The initial conditions used for operational NWP \citep{rasp_weatherbench_2024}. Operational DLWP models are usually finetuned on this dataset, rendering them theoretically ready for real-time forecasting. We use the WeatherBench2 version of the dataset\footnote{Stored on Google cloud at \url{gs://weatherbench2/datasets/hres\_t0/2016-2022-6h-1440x721.zarr}}.

\subsubsection{List of variables}\label{app:method/data/variables}
Variables modelled by Xaurora are the same as those of Aurora. They are listed in \cref{tab:list_of_variables}.
\begin{table}
\centering
\caption{\textbf{List of variables}. Atmospheric variables are measured on discrete pressure levels (50, 100, 150, 200, 250, 300, 400, 500, 600, 700, 850, 925, 1000 \unit{\hecto\pascal}).}
\label{tab:list_of_variables}
\footnotesize
\begin{tabularx}{\linewidth}{XXlX}
\toprule
\textbf{Symbol} & \textbf{Units} & \textbf{Name} & \textbf{Kind} \\
\midrule
2T & \unit{\kelvin} & 2m-temperature & Surface \\
MSL & \unit{\pascal} & Mean sea level pressure & Surface \\
10U & \unit{\meter\per\second} & 10m-zonal wind component (east-west) & Surface \\
10V & \unit{\meter\per\second} & 10m-meridional wind component (north-south) & Surface \\
\midrule
T & \unit{\kelvin} & Temperature & Atmosphere \\
Z & \unit{\meter^2\per\second^2} & Geopotential & Atmosphere \\
U & \unit{\meter\per\second} & Zonal wind component (east-west) & Atmosphere \\
V & \unit{\meter\per\second} & Meridional wind component (north-south) & Atmosphere \\
Q & \unit{\kilo\gram\per\kilo\gram} & Specific humidity & Atmosphere \\
\bottomrule
\end{tabularx}
\end{table}

\subsection{Aurora: the foundation model prior}\label{subsec:xaurora/aurora}
Aurora \citep{bodnar_foundation_2025} is one of the first atmospheric foundation models (along with ClimaX \citep{nguyen_climax_2023}). It is competitive with SOTA operational NWP and DLWP models, and was shown to generalise to other tasks with limited finetuning. Aurora is a Transformer-based \citep{vaswani_attention_2017} deterministic autoregressive model that predicts $\sistate^{t+\dt}$ from $\bh^t$. The model has three main components: (1) a Perceiver-based \citep{jaegle_perceiver_2021} encoder that projects 3D atmospheric states into a shared latent grid, (2) a Swin Transformer \citep{liu_swin_2021} backbone operating on the latents, and (3) a Perceiver-based decoder to project the latents back to physical space. Spatiotemporal information (scale, position, absolute time) is vectorised with Fourier expansions and added to the output of the encoder. A Fourier expansion of the lead time is embedded in the Swin Transformer with AdaLN \citep{perez_film_2018, peebles_scalable_2023}. Three versions of the model are publicly available and are listed in \cref{tab:aurora_models}.
\begin{table}
\centering
\caption{Aurora model variants in \cite{bodnar_foundation_2025}.}
\label{tab:aurora_models}
\footnotesize
\begin{tabularx}{0.82\linewidth}{@{}llcX@{}}
\toprule
\textbf{Model} & \textbf{Size} & \textbf{Accuracy} & \textbf{Key characteristics} \\
\midrule
Aurora-S & 113M & Lowest  & Fastest and lightest. \\
Aurora-L & 1.3B & High    & Pretrained foundation model. \\
Aurora-O & 1.3B & Highest & Aurora-L finetuned for operational weather forecasts. \\
\bottomrule
\end{tabularx}
\end{table}

\subsection{Xaurora}\label{app:method/xaurora}
\subsubsection{Architecture details}\label{app:method/xaurora/arch}
\begin{figure}
    \centering
    \includegraphics[width=1\linewidth]{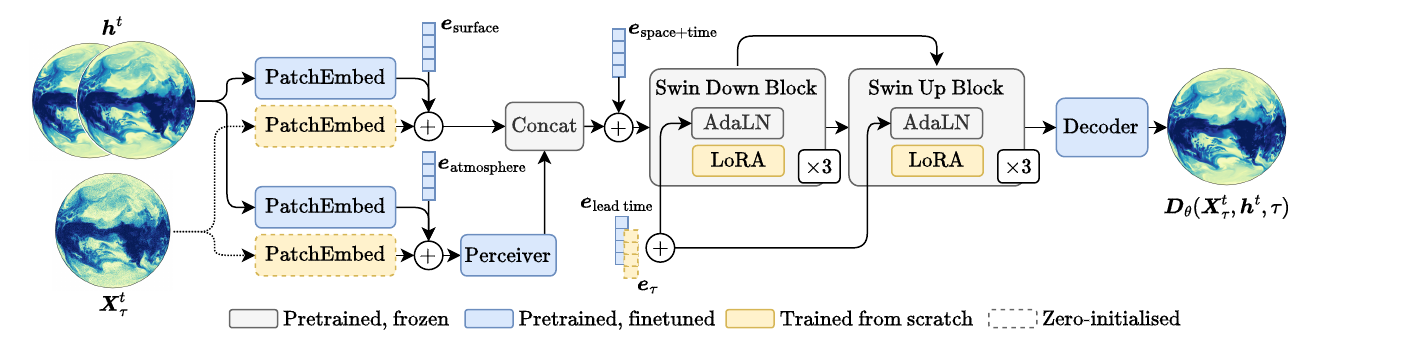}
    \caption{\textbf{Xaurora denoiser architecture}. Minimal modifications to Aurora that turn it into the denoiser $\bD_\theta(\SIstate^t_\tau, \bh^t, \tau)$. New parameters are initialised such that the model reduces to Aurora before training. Embeddings: $\bm{e}_\text{surface}$, $\bm{e}_\text{atmosphere}$ are Aurora's surface/atmospheric variable embeddings, $\bm{e}_\text{space+time}$ the spatiotemporal position embedding, $\bm{e}_\text{lead time}$ the lead-time embedding, and $\bm{e}_\tau$ the new (zero-initialised) pseudo-time embedding.}
    \label{fig:xaurora/denoiser_arch}
\end{figure}

\begin{table}
\centering
\caption{Parameter increase in the denoiser compared to Aurora-S and Aurora-L.}
\label{tab:denoiser_params}
\begin{tabularx}{0.435\linewidth}{lrr}
\toprule
\textbf{Component} & \textbf{Aurora-S} & \textbf{Aurora-L} \\
\midrule
LoRA                        & 540,672 & 2,850,816 \\
SDE state $\SIstate^t_\tau$     &  49,664 &    99,328 \\
Pseudo-time $\tau$          & 131,584 &   525,312 \\
\midrule
Total    & 721,920 & 3,475,456 \\
Increase & +0.64\% & +0.28\% \\
\bottomrule
\end{tabularx}
\end{table}
\cref{fig:xaurora/denoiser_arch} shows the architecture of the Xaurora denoiser with the changes described in \cref{subsec:xaurora/denoiser}. \cref{tab:denoiser_params} shows the parameter count for each addition. Most of the parameters come from the LoRA layers in the Swin transformer. The total parameter increase is less than 1\% for Aurora-S and Aurora-L. 

The Fourier expansion used to embed the pseudo-time $\tau$ into a vector of size $D$ is taken from Aurora's encoding of position, scale, level, and time. It can be expressed as follows:
\begin{equation}\label{eq:fourier_expansion_tau}
    \mathrm{Vec}(\tau) = \left[\cos\frac{2\pi\tau} {\lambda_i},\sin\frac{2\pi\tau}{\lambda_i}\right], \quad 0 \leq i < D/2,
\end{equation}
where
\begin{equation}
    \lambda_i = \exp\left(\log\lambda_{\min} +i\cdot\frac{\log\lambda_{\max}-\log\lambda_{\min}}{D/2-1}\right),
\end{equation}
with $\lambda_\text{min}=10^{-6}$ and $\lambda_\text{max}=1$.

\subsubsection{Normalisation and network preconditioning}\label{app:method/xaurora/norm_precond}
We reuse the per-variable normalisation statistics from Aurora (see \cite{bodnar_foundation_2025}, Section B.5). In addition, we ensure that the new inputs to the denoiser have a zero-mean and a unit-variance on average by adapting \cite{karras_elucidating_2022} to the DSI parameterisation:
\begin{align}\label{eq:training_details/precond_generic_karras}
    \bD_\theta(\SIstate_\tau^t, \bh^t, \tau) \coloneq c_\text{out}(\tau)\tilde{\bD}_\theta\big(c_{\SIstate}(\tau)\SIstate_\tau^t, c_{\sistate}(\tau)\bh^t, c_\tau(\tau)\,\tau\big).
\end{align}
As the physical states $\sistate^{t-\dt:t+\dt}$ are standardised, $c_\text{out}(\tau)=c_{\sistate}(\tau)=1$. The pseudo-time is vectorised with a Fourier expansion, which is standardised by construction. The SDE state, $\SIstate_\tau^t$, has zero mean but non-unit variance, therefore we set the conditioning function to the approximate inverse standard deviation. For simplicity, we ignore the correlation between $\sistate^t$ and $\sistate^{t+\dt}$.
\begin{align}\label{eq:training_details/precond}
    \bD_\theta(\SIstate^t_\tau, \bh^t, \tau) \coloneq \tilde{\bD}_\theta\big(c(\tau)\,\SIstate^t_\tau,\; \bh^t,\; \tau\big), \qquad c(\tau) \coloneq \frac{1}{\sqrt{\Var[\bI^t_\tau]}} \approx \frac{1}{\sqrt{\alpha_\tau^2+\beta_\tau^2+\tau\gamma_\tau^2}}.
\end{align}

\subsection{Evaluation}\label{app:method/eval}

\subsubsection{Metrics}\label{app:method/eval/metrics}
We use the latest version of the WeatherBench evaluation\footnote{Adapted from \url{https://github.com/google-research/weatherbenchX/blob/main/public\_benchmark/run\_benchmark\_evaluation.py}}. All the metrics below are adapted from \citet{rasp_weatherbench_2024}. They are computed per variable and per lead time, so we drop both indices: $\sistate$ is the ground truth, and $\gen{\sistate}^{(m)}$ the $m$-th of the $M$ ensemble members, all on the $N_\text{lat}\times N_\text{lon}$ grid. The metrics are latitude-weighted to compensate for the non-uniformity of the grid. With $\mathrm{latitude}(i)$ mapping a cell index to its latitude in radians, the weights are
\begin{equation}\label{eq:latitdue_weights}
    w(i) \coloneq \frac{\cos(\mathrm{latitude}(i))}{\frac{1}{N_\text{lat}}\sum_{j=1}^{N_\text{lat}}\cos(\mathrm{latitude}(j))}.
\end{equation}
We denote the number of forecast initialisations with $N_\text{init}$, the ensemble mean with $\bar{\sistate}=\frac{1}{M}\sum_m\gen{\sistate}^{(m)}$ and the unbiased sample variance with $\bm{\sigma}^2=\frac{1}{M-1}\sum_m(\gen{\sistate}^{(m)}-\bar{\sistate})^2$.

\textbf{Root Mean Squared Error (RMSE).} The RMSE quantifies the discrepancies between predictions and ground truth.
\begin{equation}\label{eq:rmse_wb2}
    \text{RMSE} \coloneq \sqrt{\frac{1}{N_\text{init}N_\text{lat}N_\text{lon}}\sum_{t=1}^{N_\text{init}}\sum_{i=1}^{N_\text{lat}}\sum_{j=1}^{N_\text{lon}}w(i)\big(\gen{\sistate}_{t,i,j}-\sistate_{t,i,j}\big)^2}.
\end{equation}
The RMSE can be used to assess an ensemble mean, but is biased by the ensemble size $M$. It is the right choice to evaluate the effect of different ensemble sizes, but not to compare ensembles of varying sizes.

\textbf{Unbiased Ensemble Mean RMSE (uRMSE).} The uRMSE removes the finite-ensemble bias $\bm{\sigma}^2_{t,i,j}/M$ from the residuals, so that ensembles of different sizes can be directly compared. 
\begin{equation}\label{eq:urmse_wb2}
    \text{uRMSE} \coloneq \sqrt{\frac{1}{N_\text{init}N_\text{lat}N_\text{lon}}\sum_{t=1}^{N_\text{init}}\sum_{i=1}^{N_\text{lat}}\sum_{j=1}^{N_\text{lon}}w(i)\left[\big(\bar{\sistate}_{t,i,j}-\sistate_{t,i,j}\big)^2 - \frac{\bm{\sigma}^2_{t,i,j}}{M}\right]}.
\end{equation}
Note that uRMSE is not defined for deterministic models since the bias term is singular with $M=1$.

\textbf{Fair Continuous Ranked Probability Score (CRPS).} The CRPS measures the difference between the cumulative distributions of the forecasts and the observations \citep{gneiting_strictly_2007}. It is defined as the combination of a skill and a spread component. With
\begin{align}\label{eq:crps_terms_wb2}
    \text{CRPS}_\text{skill} &\coloneq \frac{1}{N_\text{init}MN_\text{lat}N_\text{lon}}\sum_{t=1}^{N_\text{init}}\sum_{m=1}^{M}\sum_{i=1}^{N_\text{lat}}\sum_{j=1}^{N_\text{lon}} w(i) \big\lvert \gen{\sistate}^{(m)}_{t,i,j}-\sistate_{t,i,j} \big\rvert \\
    \text{CRPS}_\text{spread} &\coloneq \frac{1}{N_\text{init}}\sum_{t=1}^{N_\text{init}}\frac{1}{2M(M-1)} \sum_{m=1}^{M}\sum_{\bar{m}=1}^{M}\frac{1}{N_\text{lat}N_\text{lon}}\sum_{i=1}^{N_\text{lat}}\sum_{j=1}^{N_\text{lon}} w(i) \big\lvert \gen{\sistate}^{(m)}_{t,i,j}-\gen{\sistate}^{(\bar{m})}_{t,i,j} \big\rvert,
\end{align}
the \emph{fair} CRPS, which is unbiased for the $M\to\infty$ limit and therefore comparable across ensemble sizes, is given by
\begin{equation}\label{eq:crps_wb2}
    \text{CRPS} \coloneq \text{CRPS}_\text{skill} - \text{CRPS}_\text{spread}.
\end{equation}

\textbf{Spread-Skill Ratio (SSR).} The SSR measures the calibration as the ratio of the ensemble spread to the unbiased RMSE of the ensemble mean. 
\begin{equation}\label{eq:ssr_wb2}
    \text{SSR} \coloneq \frac{\sqrt{\frac{1}{N_\text{init}N_\text{lat}N_\text{lon}}\sum_{t=1}^{N_\text{init}}\sum_{i=1}^{N_\text{lat}}\sum_{j=1}^{N_\text{lon}}w(i)\,\bm{\sigma}^2_{t,i,j}}}{\text{uRMSE}}.
\end{equation}

\textbf{Power Spectral Density (PSD).} We use the zonal energy spectrum of \citet{rasp_weatherbench_2024}. Along the circle of
constant latitude $\varphi$, which has $L=N_\text{lon}$ samples $f_0(\varphi),\dots,f_{L-1}(\varphi)$
and circumference $C(\varphi)=2\pi R_e\cos\varphi$, the discrete Fourier transform is
$F_k(\varphi)=\frac{1}{L}\sum_{l=0}^{L-1}f_l(\varphi)\,e^{-i2\pi kl/L}$, and the spectrum of that
single circle is
\begin{equation}\label{eq:psd_lat}
    S_0(\varphi) \coloneq C(\varphi)\,\lvert F_0(\varphi)\rvert^2, \qquad
    S_k(\varphi) \coloneq 2\,C(\varphi)\,\lvert F_k(\varphi)\rvert^2,
    \quad k=1,\dots,\lfloor L/2\rfloor ,
\end{equation}
where the factor $2$ for $k>0$ accounts for both positive and negative frequency content. This
normalisation makes Parseval's relation hold at each latitude,
$\int_0^{C(\varphi)}\lvert f(l,\varphi)\rvert^2\,\rd l \approx
\frac{C(\varphi)}{L}\sum_l \lvert f_l(\varphi)\rvert^2 = \sum_k S_k(\varphi)$,
so $S_k(\varphi)$ decomposes that circle's total power by wavenumber $k$. The reported power
spectral density is the latitude average
\begin{equation}\label{eq:psd}
    P(k) \coloneq \frac{\sum_{\varphi\in\Phi} w(\varphi)\,S_k(\varphi)}{\sum_{\varphi\in\Phi} w(\varphi)} ,
\end{equation}
over the latitude set $\Phi$ with weights $w(\varphi)$. Unless stated otherwise, we use the entire globe ($\Phi=[-90^\circ,90^\circ]$) in our evaluation.

\chgd{\textbf{Spectral Spread-Skill Ratio (SSR($\ell$)).} The SSR can be decomposed by spatial scale with the spherical harmonics transform (SHT) implemented in torch-harmonics \citep{kurth2026library}. A field $f$ on the grid is expanded on orthonormal spherical harmonics with coefficients $a_{\ell m}[f]$, where the degree $\ell$ indexes the spatial scale, and its angular power spectrum is $C_\ell[f]\coloneq\frac{1}{2\ell+1}\sum_{m=-\ell}^{\ell}\lvert a_{\ell m}[f]\rvert^2$. The spectra of the ensemble spread, of the ensemble mean error, and of the unbiased error are}
\begin{align}\label{eq:spectral_terms}
    S(\ell) &\coloneq \frac{1}{N_\text{init}}\sum_{t=1}^{N_\text{init}}\frac{1}{M-1}\sum_{m=1}^{M} C_\ell\big[\gen{\sistate}^{(m)}_t-\bar{\sistate}_t\big]\\
    \bar{E}(\ell) &\coloneq \frac{1}{N_\text{init}}\sum_{t=1}^{N_\text{init}} C_\ell\big[\bar{\sistate}_t-\sistate_t\big] \\
    E_u(\ell) &\coloneq \bar{E}(\ell)-\frac{S(\ell)}{M},
\end{align}
\chgd{where $E_u$ removes the finite-ensemble bias as in \cref{eq:urmse_wb2}. The spectral SSR is}
\begin{equation}\label{eq:ssr_l}
    \text{SSR}(\ell) \coloneq \sqrt{S(\ell)/E_u(\ell)}.
\end{equation}

\chgd{\textbf{Rank Histogram (RH).} The rank histogram measures the reliability of the ensemble by counting where the ground truth falls among the sorted members \citep{hamill_interpretation_2001}. With the rank of the ground truth $r_{t,i,j}\coloneq\sum_{m=1}^{M}\mathbf{1}\big[\gen{\sistate}^{(m)}_{t,i,j}<\sistate_{t,i,j}\big]\in\{0,\dots,M\}$, the histogram is}
\begin{equation}\label{eq:rank_hist}
    \text{RH}(r) \coloneq \frac{M+1}{N_\text{init}N_\text{lat}N_\text{lon}}\sum_{t=1}^{N_\text{init}}\sum_{i=1}^{N_\text{lat}}\sum_{j=1}^{N_\text{lon}}w(i)\,\mathbf{1}\big[r_{t,i,j}=r\big], \qquad r=0,\dots,M,
\end{equation}
\chgd{normalised such that a calibrated ensemble is flat ($\text{RH}=1$); we plot it against the normalised rank $r/M$. A $\cup$-shape indicates underdispersion (the truth falls too often outside the ensemble), a $\cap$-shape overdispersion, and a tilt a bias.}

\chgd{\textbf{Binned spread-skill.} A global SSR of 1 does not guarantee that the spread tracks the error locally. Inspired by \citet{leutbecher_ensemble_2007}, we bin the grid points by their spread relative to the area-weighted field mean of the same forecast,}
\begin{equation}\label{eq:rel_spread}
    \rho_{t,i,j} \coloneq \bm{\sigma}_{t,i,j}\Big/\sqrt{\frac{1}{N_\text{lat}N_\text{lon}}\sum_{i'=1}^{N_\text{lat}}\sum_{j'=1}^{N_\text{lon}}w(i')\,\bm{\sigma}^2_{t,i',j'}},
\end{equation}
\chgd{using 14 bins $\mathcal{B}_b$ with log-spaced edges $2^{-3},2^{-2.5},\dots,2^{3}$ (the outer two bins are open). With $\mathbf{1}_b\coloneq\mathbf{1}[\rho_{t,i,j}\in\mathcal{B}_b]$, the spread and unbiased RMSE of each bin are}
\begin{equation}\label{eq:binned_ss}
    \text{spread}_b \coloneq \sqrt{\frac{\sum_{t,i,j}w(i)\,\bm{\sigma}^2_{t,i,j}\mathbf{1}_b}{\sum_{t,i,j}w(i)\,\mathbf{1}_b}}, \qquad
    \text{RMSE}_b \coloneq \sqrt{\frac{\sum_{t,i,j}w(i)\Big[\big(\bar{\sistate}_{t,i,j}-\sistate_{t,i,j}\big)^2-\frac{\bm{\sigma}^2_{t,i,j}}{M}\Big]\mathbf{1}_b}{\sum_{t,i,j}w(i)\,\mathbf{1}_b}}.
\end{equation}
\chgd{A reliable ensemble lies on the identity line in every bin; above it, the ensemble is overconfident at that spread level, below it underconfident. Because the bins are defined on relative spread, a global rescaling of the spread shifts the curve along the spread axis without changing its slope; only flow-dependent spread can bring it onto the identity line.}

\subsubsection{Statistical significance}\label{app:method/eval/significance}
In most experiments, we report statistical significance. We obtain the intervals by re-sampling the metrics over our 56 initialisations before temporal aggregation. When comparing two models (\emph{e.g.}, ablations), we use paired bootstrapping by comparing the models on the same set of resampled initialisations with replacement. This approach removes the weather variability that dominates with 56 initialisations. Formally, for a metric $\mathcal{M}$, we bootstrap the relative difference $\Delta=(\mathcal{M}_{\text{model}}-\mathcal{M}_{\text{ref}})/\mathcal{M}_{\text{ref}}$ with 10,000 replicates and report percentile intervals and two-sided percentile $p$-values at $\alpha=0.05$.

The GenCast-O forecasts are not publicly available. Hence, uncertainty quantification cannot be reported. Instead, we download the aggregated benchmark metrics directly from WeatherBench.

\subsubsection{Forecast settings}\label{app:method/eval/forecast_settings}
All forecasts are initialised on 56 different evenly spread dates across 2022 (either at 12 or 00 UTC). This is in line with \cite{kossaifi_demystifying_2026}, who report statistically significant results using only 28 initialisations, massively decreasing the compute and storage requirements over the WeatherBench approach (2 initialisations per day throughout 2022). To reduce the forecasts size on disk and maximise write speed, we store all surface variables, but only a subset of pressure levels (250, 500, 700, 850 \unit{\hecto\pascal}).

\chgd{\textbf{Xaurora.} We generate ensemble forecasts of 10 members to minimise computational cost without compromising on the metrics. Xaurora forecasts are generated autoregressively using \cref{alg:ar_sampling} with an Euler--Maruyama solver and $S=10$ integration steps. The SDE state evolves as}
\begin{equation}
\SIstate_{\tau+\dtau}^t=\SIstate^t_{\tau} + \drift_\theta(\SIstate^t_{\tau}, \bh^t, \tau) \,\frac{1}{S} + \gamma_{\tau} \Delta \bm{w}_{\tau},
\end{equation}
\chgd{with $\Delta\bm{w}_{\tau}=\sqrt{\Delta\tau}\,\bm{z},\,\bm{z}\sim\mathcal{N}(\bm{0},\Id)$. Hence, the final state, $\hat{\sistate}^{t+\dt}=\SIstate^t_{1}$, carries residual noise with variance $\gamma^2_{1-1/S}\,\cdot1/S$. Before writing the forecasts to disk we filter out the noise by subtracting the last Wiener sample:}
\begin{equation}
    \hat{\sistate}^{t+\dt}_\text{toDisk} := \hat{\sistate}^{t+\dt} - \gamma_{1-1/S} \Delta \bm{w}_{1-1/S}.
\end{equation}
\chgd{This is analogous to recording the output of the denoiser before applying \cref{eq:dsi/drift_from_denoiser}. The noise is kept in the SDE state used to generate the next rollout step. \cref{app:results/importance_subtract_noise} discusses the importance of subtracting the residual noise from the predictions. Note that since we use 10 steps, the clipping introduced in \cref{sec:dsi} never engages at inference.}

\begin{figure}
    \centering
    \includegraphics[width=1\linewidth]{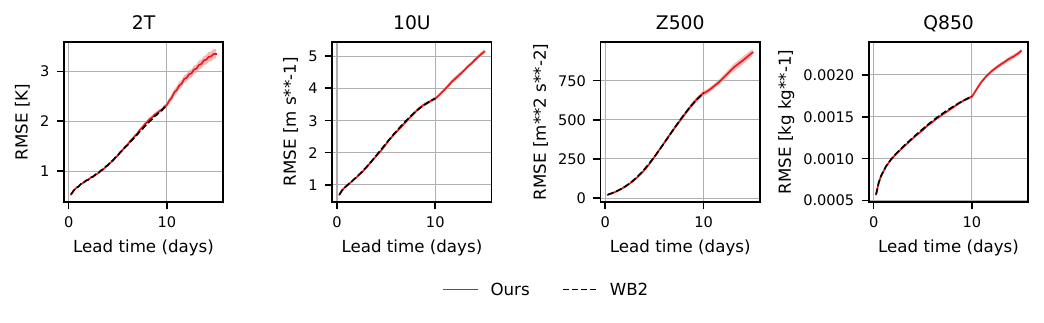}
    \caption{\textbf{Aurora-O validation.} RMSE of Aurora-O from the official WeatherBench2 (WB2, dashed black) and from our own run (Ours, red). The two runs are essentially equivalent, with negligible variations coming from different sets of initialisations. The WB2 run only goes up to 10 days and does not show the discontinuity at 10 days, beyond which Aurora-O was not rollout-finetuned. \rebuttal{The shading is the 95\% confidence interval bootstrapped over the initialisations.}}
    \label{fig:aurora_o_validation}
\end{figure}

\textbf{Aurora.} Aurora-O metrics are only publicly available up to 10 days of lead time\footnote{ \url{https://storage.googleapis.com/weatherbench2/benchmark_results/aurora_vs_hres_t0_1440x721_2022.nc}} and the metrics of Aurora-S and Aurora-L are not available. Therefore, we run all the Aurora forecasts discussed in this paper ourselves with the settings discussed above. \cref{fig:aurora_o_validation} shows that our results align with the official evaluation.

\subsubsection{Baseline models}\label{app:method/eval/baselines}

\begin{figure}
    \centering
    \includegraphics[width=1\linewidth]{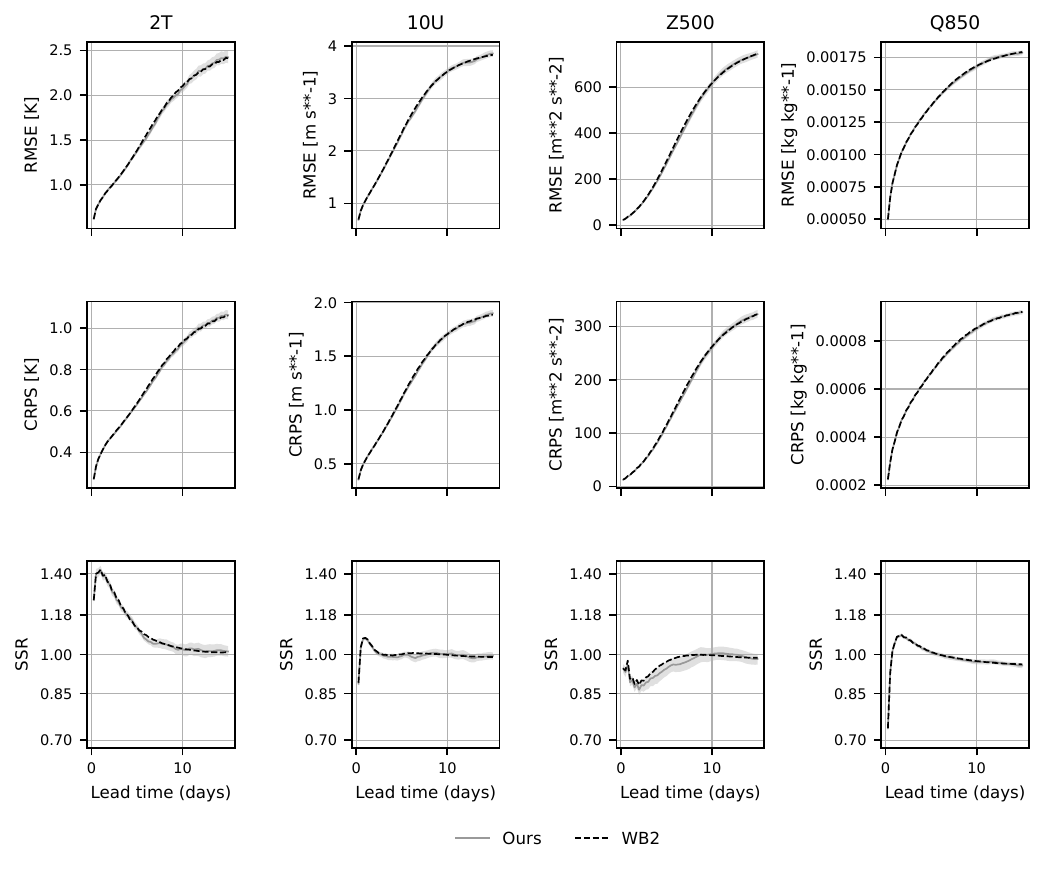}
    \caption{\rebuttal{\textbf{IFS ENS validation.} uRMSE, CRPS, and SSR of IFS ENS from the official WeatherBench2 (WB2, dashed black) and from our own run (Ours, grey). The two runs are essentially equivalent: the official values lie within our 95\% confidence intervals (bootstrapped over initialisations) except for Z500 around day 5, where our subset yields a slightly lower uRMSE ($-3.8\%$), CRPS ($-2.5\%$) and SSR ($-2.8\%$). Since uRMSE and fair CRPS correct for ensemble size, this difference comes mainly from the choice of initialisations, and it slightly favours IFS ENS over Xaurora.}}
    \label{fig:ifs_validation}
\end{figure}

We compare the predictions with the following baselines:
\begin{itemize}[leftmargin=*]
    \item \textbf{Operational ensemble models:} the Integrated Forecasting System Ensemble (IFS ENS)\footnote{ \url{https://storage.googleapis.com/weatherbench2/benchmark_results/ens_vs_hres_t0_1440x721_2022.nc}} and GenCast operational (GenCast-O)\footnote{ \url{https://storage.googleapis.com/weatherbench2/benchmark_results/gencast_operational_100m_uv_vs_hres_t0_1440x721_2022.nc}} \citep{price_gencast_2023}. In \cref{subsec:experiments/mwf}, we evaluate the IFS ENS forecasts to match the Xaurora setup (10 members, 56 initialisations) and provide confidence intervals. \cref{fig:ifs_validation} shows that the numbers are essentially equivalent to the full evaluation, with the largest deviation for Z500 around day 5, where our subset slightly favours IFS ENS. Since we never claim that Xaurora is better than IFS, minor differences do not affect the conclusions.
    \item \textbf{Operational deterministic models:} Aurora operational (Aurora-O) \citep{bodnar_foundation_2025};
    \item \textbf{Non-operational Aurora versions:} Aurora-Large (Aurora-L) and Aurora-Small (Aurora-S) \citep{bodnar_foundation_2025}. 
\end{itemize}    
\cref{tab:aurora_models} lists the characteristics of the different Aurora models. \rebuttal{Aurora 1.5 \citep{weyn_aurora_2026} is not included since it is evaluated on 2024 (which cannot be compared to our test set) and its forecasts and metrics are not publicly available at the time of writing.}

Operational systems are evaluated against HRES-T0, others against ERA5. Therefore, the models are not strictly comparable. Nevertheless, this is standard practice in the field. It is imperfect but the fairest approach. Operational models perform worse on ERA5 than on HRES-T0, especially at shorter lead times. Therefore, the evaluation protocol used in the paper does not favour Xaurora.

\subsection{Drift parameterisations}\label{app:method/drift_param}
\begin{table}
\centering
\caption{Neural network, target and drift term for each parameterisation}
\label{tab:drift_param}
\footnotesize
\begin{tabularx}{\linewidth}{lllX}
\toprule
 & \textbf{Network} & \textbf{Target} & \textbf{Drift} $(\E[\bR^t_\tau \mid \SIstate^t_\tau,\bh^t])$ \\
\midrule
$\drift$-prediction & $\drift_\theta$ & $\bR^t_\tau$ & $\drift_\theta$ \\
$\bW$-prediction & $\bWnet$ & $\bW_\tau$ & $\dot \alpha_\tau \sistate^t + \frac{\dot \beta_\tau}{\beta_\tau}\big[\SIstate^t_\tau - \alpha_\tau \sistate^t - \gamma_\tau \bWnet(\SIstate^t_\tau,\bh^t,\tau)\big] + \dot\gamma_\tau \bWnet(\SIstate^t_\tau,\bh^t,\tau)$ \\
$\sistate_1$-prediction & $\bD_\theta$ & $\sistate^{t+\dt}$ & $\dot \alpha_\tau \sistate^t + \dot \beta_\tau \bD_\theta(\SIstate^t_\tau,\bh^t,\tau) + \frac{\dot \gamma_\tau}{\gamma_\tau}\big[\SIstate^t_\tau - \alpha_\tau \sistate^t - \beta_\tau \bD_\theta(\SIstate^t_\tau,\bh^t,\tau)\big]$ \\
\bottomrule
\end{tabularx}
\end{table}

\cref{tab:drift_param} lists the drift parameterisations of the Stochastic Interpolant discussed in \cref{sec:dsi}. We obtain these results by expanding the drift as an expectation and isolating relevant terms in \cref{eq:stochastic_interpolant},
\begin{equation}
    \drift_\theta(\SIstate^t_\tau,\bh^t,\tau) = \E[\bR^t_\tau \mid \SIstate^t_\tau,\bh^t]
    =\dot \alpha_\tau \sistate^t + \dot \beta_\tau \E[\sistate^{t+\dt} \mid \SIstate^t_\tau,\bh^t] + \dot \gamma_\tau \E[\bW_\tau \mid \SIstate^t_\tau,\bh^t].
\end{equation}

\subsubsection{Loss function equivalences}\label{app:method/drift_param/loss_equivalences} 
Similarly to \cite{li_back_2026}, the $\sistate_1$-space loss in \cref{eq:dsi/xloss} can be expressed in $\drift$- and $\bW$-space by a single $\tau$-dependent reweighting: 
\begin{align}
\mathcal{L}_{\drift\mid\bD}(\theta) &= \int_0^1\E_{(\bh^t,\sistate^{t+\dt})\sim p_\text{data},\bW_\tau\sim\mathcal{N}(\bm{0},\tau\Id)}\Big[\underbrace{\left(\dot\beta_\tau-\frac{\dot\gamma_\tau}{\gamma_\tau}\beta_\tau\right)^2}_{\eqcolon\lossw{\drift}(\tau)}\norm{\bD_\theta(\bI^t_\tau,\bh^t,\tau)-\sistate^{t+\dt}}_2^2\Big]\rd\tau  \label{eq:xloss_in_drift_space} \\
\mathcal{L}_{\bW\mid\bD}(\theta) &= \int_0^1 \E_{(\bh^t,\sistate^{t+\dt})\sim p_\text{data},\bW_\tau\sim\mathcal{N}(\bm{0},\tau\Id)}\Big[\underbrace{\left(\frac{\beta_\tau}{\gamma_\tau}\right)^2}_{\eqcolon\lossw{\bW}(\tau)}\norm{\bD_\theta(\bI^t_\tau,\bh^t,\tau)-\sistate^{t+\dt}}_2^2\Big]\rd\tau. \label{eq:xloss_in_W_space}
\end{align}
The derivation is purely algebraic. For $\lossw{\drift}(\tau)$, we plug \cref{eq:dsi/drift_from_denoiser} and the definition of $\bR^t_\tau$ in \cref{eq:SI_drift_loss}. For $\lossw{\bW}(\tau)$, use $\bWnet(\SIstate^t_\tau,\bh^t,\tau) = \frac{1}{\gamma_\tau}[\SIstate^t_\tau - \alpha_\tau \sistate^t - \beta_\tau\bD_\theta(\SIstate^t_\tau,\bh^t,\tau)]$ and $\mathcal{L}_{\bW}(\theta) \coloneq \int_0^1 \E_{(\bh^t,\sistate^{t+\dt})\sim p_\text{data},\bW_\tau\sim\mathcal{N}(\bm{0},\tau\Id)}\left[\norm{ \bWnet(\bI^t_\tau,\bh^t,\tau) - \bW_\tau}^2_2 \right] \rd \tau$. We train in $\sistate_1$-space and leave the comparison for future work. \rebuttal{Note that similar expressions can be derived for $\mathcal{L}_{\bD\mid\drift}$, $\mathcal{L}_{\bW\mid\drift}$, $\mathcal{L}_{\bD\mid\bW}$, and $\mathcal{L}_{\drift\mid\bW}$.}

\subsubsection{Mathematical generalisation of the 2D toy example}\label{app:toy_maths}
The 2-dimensional toy example shown in \cref{fig:dsi/toy_example} supports that the DSI parameterisation is preferable when finetuning an ideal prior model of the form $\bm{G}=\E[\sistate_1\mid\sistate_0]$ (we ignore the temporal dimension $t$ for conciseness, without loss of generality). This section provides the same motivation from a mathematical perspective valid for problems of arbitrary finite dimension.

The central idea is always the same: we use the prior $\bm{G}$ in lieu of the network in the expression of the drift corresponding to each parameterisation (last column of \cref{tab:drift_param}). The obtained drift, referred to as $\drift(\bm{G})$ below, is then substituted in the SI SDE (\cref{eq:SI_SDE}). We can then approximate the mean SDE trajectory (the bold lines in \cref{fig:dsi/toy_example}) by ignoring the Wiener process and solving the following ODE
\begin{equation}
    \rd \bar\SIstate_\tau = \drift(\bm{G}) \rd\tau, \quad \bar\SIstate_0 = \sistate_0.
\end{equation}
We use $\alpha_\tau=\gamma_\tau=1-\tau$, $\beta_\tau=\tau$, and $\sistate^t=\sistate_0$.

\textbf{$\drift$-prediction.} Let $\drift=\bm{G}$:
\begin{align*}
    \drift(\bm{G}) =  \bm{G} \implies
    \rd \bar\SIstate_\tau = \drift(\bm{G}) \rd\tau = \E[\sistate_1\mid\sistate_0] \rd\tau.
\end{align*}
Hence, $\bar\SIstate_\tau = \E[\sistate_1\mid\sistate_0]\cdot\tau + \sistate_0$. The ODE endpoint is therefore $\bar\SIstate_1 = \E[\sistate_1\mid\sistate_0] + \sistate_0 \not= \E[\sistate_1\mid\sistate_0]$. Hence, $\drift$-prediction does not converge to $\E[\sistate_1\mid\sistate_0]$.

\textbf{$\bW$-prediction.} Let $\bW=\bm{G}$:
\begin{align*}
    \drift(\bm{G}) &= \dot \alpha_\tau \sistate_0 + \frac{\dot \beta_\tau}{\beta_\tau}\big[\SIstate_\tau - \alpha_\tau \sistate_0 - \gamma_\tau \bm{G}\big] + \dot\gamma_\tau \bm{G} \\
    &= - \sistate_0 + \frac{1}{\tau}\big[\SIstate_\tau - (1-\tau) \sistate_0 - (1-\tau) \bm{G}\big] - \bm{G} \\
    &= \frac{\SIstate_\tau - (\sistate_0+\bm{G})}{\tau}.
\end{align*}
As mentioned in the main text, we clip the denominator of the second term to 0.05 to avoid numerical instabilities. Hence,
\begin{align*}
    \rd \bar\SIstate_\tau = \frac{\bar\SIstate_\tau - (\sistate_0+\bm{G})}{\max(\tau, 0.05)} \rd \tau = \frac{\bar\SIstate_\tau - (\sistate_0+\E[\sistate_1\mid\sistate_0])}{\max(\tau, 0.05)} \rd \tau.
\end{align*}
Solving for $0\leq\tau<0.05$ and $0.05\leq\tau\leq 1$ separately and imposing continuity at $\tau=0.05$ gives
\begin{equation*}
    \bar\SIstate_\tau = \begin{cases}
        \sistate_0 + \E[\sistate_1\mid\sistate_0] - \E[\sistate_1\mid\sistate_0] e^{\tau/0.05} &\quad \text{if }0\leq\tau<0.05, \\
        \sistate_0 + \E[\sistate_1\mid\sistate_0] - \frac{e}{0.05}\E[\sistate_1\mid\sistate_0]\cdot\tau &\quad\text{if }0.05\leq\tau\leq 1.
    \end{cases}
\end{equation*}
In particular, $\bar\SIstate_1 = \sistate_0 + \E[\sistate_1\mid\sistate_0] - \frac{e}{0.05}\E[\sistate_1\mid\sistate_0] \not= \E[\sistate_1\mid\sistate_0]$. Hence, $\bW$-prediction does not converge to $\E[\sistate_1\mid\sistate_0]$.

\textbf{$\sistate_1$-prediction (DSI).} Let $\bD=\bm{G}$:
\begin{align*}
    \drift(\bm{G}) &= \dot \alpha_\tau \sistate_0 + \dot \beta_\tau \bm{G} + \frac{\dot \gamma_\tau}{\gamma_\tau}\big[\SIstate_\tau - \alpha_\tau \sistate_0 - \beta_\tau \bm{G}\big] \\
    &= - \sistate_0 + \bm{G} - \frac{1}{1-\tau}\big[\SIstate_\tau - (1-\tau)\sistate_0 - \tau \bm{G}\big] \\
    &= -\frac{1}{1-\tau}(\bar\SIstate_\tau - \bm{G}).
\end{align*}
Hence, 
\begin{equation*}
    \rd\bar\SIstate_\tau = -\frac{1}{1-\tau}(\bar\SIstate_\tau - \bm{G})\rd\tau = -\frac{1}{1-\tau}(\bar\SIstate_\tau - \E[\sistate_1\mid\sistate_0])\rd\tau \implies \bar\SIstate_\tau = \sistate_0 + (\E[\sistate_1\mid\sistate_0]-\sistate_0)\cdot\tau,
\end{equation*}
which converges to the correct endpoint, $\bar\SIstate_1 = \E[\sistate_1\mid\sistate_0]$.

\textbf{The residual prediction case.} Aurora is parameterised for direct prediction. Another common approach in autoregressive forecasting is \textit{residual prediction}, \emph{i.e.} $\bm{G}=\E[\sistate_1-\sistate_0\mid\sistate_0]$. In that case, the preferable parameterisation would be $\drift$-prediction. Proceeding as above with the new prior,
\begin{equation*}
    \drift(\bm{G}) = \bm{G} \implies \rd\bar\SIstate_\tau  = \bm{G}\rd\tau = \E[\sistate_1-\sistate_0\mid\sistate_0]\rd\tau,
\end{equation*}
and $\bar\SIstate_\tau = \E[\sistate_1-\sistate_0\mid\sistate_0]\cdot\tau + \sistate_0$, which has the correct endpoint, $\bar\SIstate_1 = \E[\sistate_1\mid\sistate_0]$. Nevertheless, the network still targets $\bR_\tau^t$ and could therefore be less robust to the curse of dimensionality, following the arguments of \citet{li_back_2026}.

\subsection{Denoising Stochastic Interpolants}\label{app:method/dsi}
\subsubsection{General algorithms}\label{app:method/dsi/algo}
\cref{alg:ar_training} and \cref{alg:ar_sampling} show the DSI training and sampling algorithms. Our probabilistic SDE-rollout training algorithm is defined in \cref{app:sde_rollout_rb_algo}.
\begin{algorithm}[H]
    \caption{Vanilla autoregressive DSI training}
    \label{alg:ar_training}
    \begin{algorithmic}[1]
    \STATE {\bfseries Input:} $(\alpha_\tau, \beta_\tau, \gamma_\tau)$, training data $\mathcal{D}_\text{train}$, pretrained $\bD_\theta$, number of training samples $N_\text{samples}$, optimiser
    \FOR{$n=1$ {\bfseries to} $N_\text{samples}$}
        \STATE $(\sistate^{t-\dt}, \sistate^{t}, \sistate^{t+\dt}) \sim \mathcal{D} $
        \STATE $\tau \sim U[0,1]$
        \STATE $\bz\sim\mathcal{N}(\bm{0},\Id)$
        \STATE $\bW_\tau \gets \sqrt{\tau}\bz$
        \STATE $\bI^t_{\tau} \gets \alpha_\tau \sistate^t + \beta_\tau \sistate^{t+\dt} + \gamma_\tau \bW_\tau$ \quad (this Vanilla algorithm explicitly forms the interpolant.)
        \STATE $\ell(\theta) \gets \norm{\bD_\theta(\bI^t_{\tau},\bh^t,\tau)-\sistate^{t+\dt}}_2^2$ \quad (from \cref{eq:dsi/xloss})
        \STATE $\theta \gets \text{optimiser}\big(\ell(\theta), \theta\big)$
    \ENDFOR
    \STATE {\bfseries Output:} Finetuned denoiser $\bD_\theta$
    \end{algorithmic}
\end{algorithm}

\begin{algorithm}[H]
    \caption{Autoregressive DSI sampling (without residual noise subtraction)}
    \label{alg:ar_sampling}
    \begin{algorithmic}[1]
    \STATE {\bfseries Input:} Initial conditions $\bh^t$, trained denoiser $\bD_\theta$, SDE steps $S$, forecast steps $K$, SDE Solver
    \STATE $\dtau \gets 1/S$
    \STATE $\gen{\bh}^{t}\gets\bh^t$ \quad (running history; after the first step it holds generated states)
    \FOR{$k=0$ {\bfseries to} $K-1$}
        \STATE $s \gets t+k\dt$
        \STATE $\tau \gets 0$
        \STATE $\SIstate^{s}_0 \gets$ last element of $\gen{\bh}^{s}$
        \FOR{$n=1$ {\bfseries to} $S$}
            \STATE $\tau' \gets \tau+\dtau$
            \STATE $\SIstate^{s}_{\tau'} \gets \text{SDE Solver}\big(\SIstate^{s}_{\tau},\; \drift_\theta(\cdot,\gen{\bh}^{s},\cdot),\; \tau,\; \dtau\big)$ \quad (solver needs the drift function)
            \STATE $\tau \gets \tau'$
        \ENDFOR
        \STATE $\gen{\sistate}^{s+\dt} \gets \SIstate^{s}_{1}$
        \STATE $\gen{\bh}^{s+\dt} \gets (\SIstate_0^s, \gen{\sistate}^{s+\dt})$
    \ENDFOR
    \STATE {\bfseries Output:} forecast $\big\{\gen{\sistate}^{t+k\dt}\big\}_{k=1}^{K}$
    \end{algorithmic}
\end{algorithm}

\subsubsection{Single-backpropagation CRPS}\label{app:single_backprop_crps}
The fair CRPS loss used to train Xaurora is shown in the main paper, in \cref{eq:fair_crps_loss}. We use the same variable weights ($\omega$) as Aurora\footnote{Surface variables are weighted by 0.25 and atmospheric ones by 1. Individual variables also have dedicated weights, 1.5 for MSL, 0.77 for 10U, 0.66 for 10V, 3.0 for 2T, 2.8 for Z, 0.8 for Q, 1.7 for T, 0.87 for U, 0.6 for V. No latitude-weighting is used.}, such that at $\tau=0$ where the spread term ($\tfrac12\norm{\gen{\sistate}^{(m)}-\gen{\sistate}^{(\bar m)}}_\omega$) is zero, the loss is equivalent to Aurora's weighted MAE. Both members are functions of $\theta$, so by the chain rule
\begin{equation}
    \nabla_\theta\,\ell_\text{fCRPS}
    = \partial_{\gen{\sistate}^{(m)}}\ell \cdot \partial_\theta \gen{\sistate}^{(m)}
    + \partial_{\gen{\sistate}^{(\bar m)}}\ell \cdot \partial_\theta \gen{\sistate}^{(\bar m)}.
\end{equation}
To avoid backpropagating through both trajectories, we supervise only the $\gen{\sistate}^{(m)}$-trajectory and apply a stop-gradient to $\gen{\sistate}^{(\bar m)}$. Hence $\partial_\theta \gen{\sistate}^{(\bar m)}=0$ and the second term vanishes. 

The roles $\bar{m}$ and $m$ are assigned at random between the two members. Since $\ell_\text{fCRPS}$ is symmetric in its two member arguments and the members are exchangeable draws from the forecast distribution, the two terms have equal expectation. The one-sided estimator is therefore an unbiased estimate of $\tfrac12\nabla_\theta\,\ell_\text{fCRPS}$. The constant $1/2$ scaling is compensated by the learning rate and the stationary points of $\ell_\text{fCRPS}$ remain unchanged. We use a replay buffer to emulate rollouts, which introduces a minor off-policy relative to naive rollout training (see \cref{fig:replay_buffer_offpolicy,fig:replay_buffer_stats}).

\subsubsection{Probabilistic SDE-rollout replay buffer training algorithm}\label{app:sde_rollout_rb_algo}
The algorithm used to train Xaurora on ensemble forecasts is shown in \cref{alg:rb_crps}. Below, we summarise the main characteristics of the algorithm.

\begin{minipage}{0.48\linewidth}
\includegraphics[width=\linewidth]{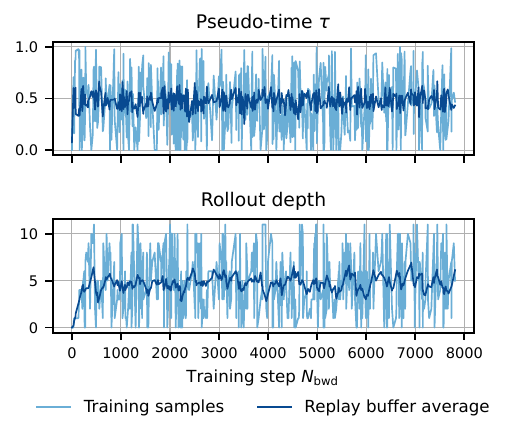}
    \captionof{figure}{\textbf{Replay-buffer rollout training statistics.} \textbf{Top:} Time series of the pseudo-time ($\tau$) of the samples. \textbf{Bottom:} Time series of the rollout depth of the samples. The values seen by the model at each training step are in light blue and the replay buffer averages are depicted in dark blue. The x-axis shows the training step $N_\text{bwd}$ as defined in \cref{tab:hyperparam}. These plots show that throughout the training, the denoiser is exposed to a variety of pseudo-times (mean=0.5) and rollout steps (mean=4.7).}
    \label{fig:replay_buffer_stats}
\end{minipage}
\hfill
\begin{minipage}{0.48\linewidth}
    \includegraphics[width=\linewidth]{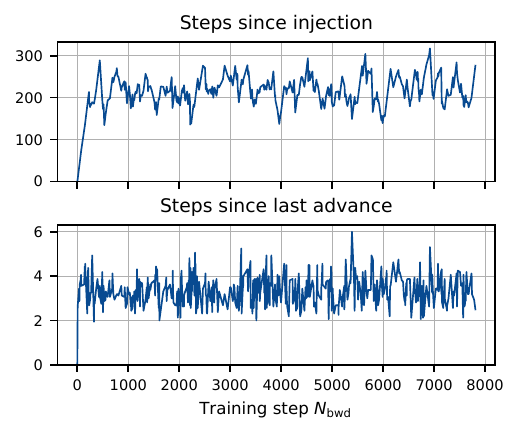}
    \captionof{figure}{\textbf{Replay-buffer training off-policy discrepancy.} \textbf{Top:} Buffer mean sample age, \emph{i.e.}, number of training steps since the trajectory was first pushed to the buffer.  \textbf{Bottom:} Average number of training steps since a sample was updated, \emph{i.e.}, popped, advanced by one SDE step, and pushed back to the buffer. The x-axis shows the training step $N_\text{bwd}$ as defined in \cref{tab:hyperparam}. These plots characterise the minor off-policy discrepancy introduced by the replay buffer compared to a naive rollout training strategy (intractable).}
    \label{fig:replay_buffer_offpolicy}
\end{minipage}

\textbf{Leak-free trajectories.} In the single-step autoregressive DSI training algorithm (\cref{alg:ar_training}), the target $\sistate^{t+\dt}$ can leak into the model input through the interpolant $\bI^t_\tau$. We start every rollout at $\tau=0$, where $\bI^t_0 = \sistate^t$, and evolve SDE trajectories using only the denoiser drift (\emph{i.e.} the network sees $\SIstate^t_\tau$, never $\bI^t_\tau$). This ensures that the target $\sistate^{t+\dt}$ appears solely in the loss calculation and that the model is trained on the rollout distribution (rather than the distribution of the interpolant). 

\textbf{Replay buffer design.} The replay buffer is a FIFO list of SDE states ($\SIstate^t_\tau$) with their associated history ($\gen{\bh}^t$), (pseudo-)temporal coordinates ($t, \tau$) and rollout information (\emph{e.g.} rollout depth). All tensors are kept in memory in FP16.

\textbf{Hyperparameters.} Our replay buffer algorithm has a few hyperparameters that define the rollout training statistics. Their values are listed in \cref{tab:hyperparam}. \cref{fig:replay_buffer_stats} shows the pseudo-time and rollout depth over the course of the training.

\textbf{Probabilistic training.} We unroll pairs of trajectories (\emph{i.e.} 2 members) that we supervise with the single-backpropagation CRPS described in \cref{eq:fair_crps_loss} and \cref{app:single_backprop_crps}.

\chgd{Furthermore, we reset the drift of both members to that of one of them at $\tau=0$ (chosen randomly, see line 21 in \cref{alg:rb_crps}). In our experiments, this configuration kept the training stable. We hypothesise that since the pretrained prior already has high skill, Xaurora might follow the path of least resistance and minimise the CRPS by inflating the spread, which eventually results in instabilities in the gradients and activations. In that context, resetting the drift of both members zeroes the spread and could act as a regularisation mechanism.}

\chgd{Finally, the clipping of \cref{eq:dsi/drift_from_denoiser} mentioned in \cref{sec:dsi} is crucial during training. We use a fairly high clipping value ($\gamma_\text{clip}=0.05$), which has the effect of bounding the minimal spread of the next state to $\gamma^2_{1-\gamma_\text{clip}}\cdot \gamma_\text{clip}$ (by the argument in \cref{app:method/eval/forecast_settings}). We also observed that a lower value ($\gamma_\text{clip}=\num{1e-16}$) degrades probabilistic metrics, indicating that the background stochasticity induced by clipping could be used as a seed for flow-dependent spread.}

\subsubsection{DSI with pretrained prior requires $\dot\beta_0\not=0$}\label{app:drift_param/beta=tau}
We show that the warm start properties of the DSI require $\dot\beta_0\not=0$. More specifically, this means that it is preferable to use a linear interpolant ($\beta_\tau=\tau$) rather than a quadratic one ($\beta_\tau=\tau^2$) when finetuning a pretrained prior using the DSI framework. 

\begin{figure}
    \centering
    \includegraphics[width=1\linewidth]{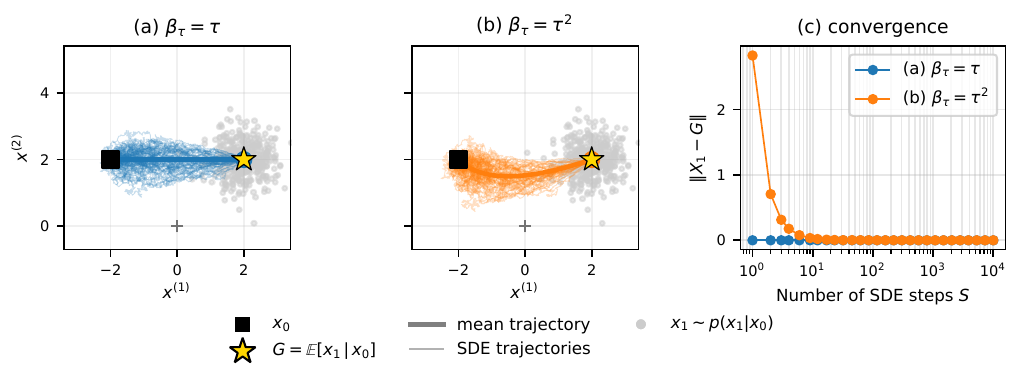}
    \caption{\textbf{Denoiser parameterisation: effect of the interpolation schedule $\beta_\tau$.} \textbf{(a,b)} Warm-start SDE trajectories for the denoiser with a linear schedule $\beta_\tau=\tau$ and a quadratic schedule $\beta_\tau=\tau^2$ (same 2D setup and markers as \cref{fig:dsi/toy_example}). \textbf{(c)} Distance of the mean SDE trajectory endpoint to the unbiased prior, $\lVert\SIstate_1-\prior\rVert$, versus the number of integration steps (Euler--Maruyama). The linear schedule is exact regardless of the integration discretisation, whereas the quadratic schedule requires more than 20 steps to converge.}
    \label{fig:drift_param/toy_dsi_beta}
\end{figure}

Substituting $\alpha_\tau=\gamma_\tau=1-\tau$ and $\beta_\tau=\tau^2$ in \cref{eq:dsi/drift_from_denoiser}, we get:
\begin{equation}
    \drift^\text{quadratic}_\tau(\SIstate^t_\tau,\bh^t,\tau) = -\frac{1}{1-\tau}\left[\SIstate^t_\tau-\tau(2-\tau)\bD_\theta(\SIstate^t_\tau,\bh^t,\tau)\right].
\end{equation}
At $\tau=0$, $\drift_0^\text{quadratic} = -\sistate^t$, which is independent of the model output and generally does not point in the direction of $\sistate^{t+\dt}$. This wrong initial step must be corrected by the subsequent integration steps, which can be directly observed in panel (c) of \cref{fig:drift_param/toy_dsi_beta} (same toy setup as \cref{fig:dsi/toy_example}); the quadratic formulation does not exactly reach $\prior$ for small numbers of SDE steps because it must correct the incorrect initial step. Meanwhile, the linear interpolant is exact regardless of the integration scheme.

Note that any choice of $\beta_\tau$ such that $\dot\beta_0=0$ will suffer from an initial drift independent of the model prediction. Conversely, any $\beta_\tau$ with non-zero initial derivative could theoretically succeed, for instance $\beta_\tau = \sin(\pi\tau/2)$. 

\subsection{Training details}\label{app:method/training_details}
We finetune Aurora-S with the SDE-rollout replay buffer described in \cref{alg:rb_crps}. The denoiser is trained in FP32 with a cosine annealing learning rate, no weight decay, and an AdamW optimiser. The total gradient norm is clipped to 1. \cref{tab:hyperparam} summarises the hyperparameters. Following the WeatherBench standards, we train Xaurora on ERA5 from 1979 to 2020, validate on 2021, and test on 2022. Aurora-S was pretrained on the same ERA5 splits, so there is no risk of data leakage.
\begin{table}[t]
\centering
\caption{\textbf{Xaurora hyperparameters}}
\label{tab:hyperparam}
\begin{tabularx}{\textwidth}{lXX}
\toprule
\textbf{Symbol} & \textbf{Meaning} & \textbf{Value} \\
\midrule
-- & Aurora variant & Aurora-S \\
$p_\text{dropout}$ & Dropout rate \citep{hinton_improving_2012} & 0.0 \\
$p_\text{dropoutAttn}$ & Dropout rate (attention layers-only) & 0.1 \\
$p_\text{droppath}$ & Stochastic depth \citep{huang_deep_2016} & 0.0 \\
$r_\text{LoRA}$ & LoRA rank & 8 \\
\midrule
$\alpha_\tau$ & Stochastic Interpolant coefficient & $1-\tau$ \\
$\beta_\tau$ & Stochastic Interpolant coefficient & $\tau$ (according to \cref{app:drift_param/beta=tau}) \\
$\gamma_\tau$ & Stochastic Interpolant coefficient & $1-\tau$ \\
\midrule
$N_\text{batch}$ & batch size per rank & 1 \\
$N_\text{samples}$ & total training samples & 500k \\
$N_\text{accum}$ & gradient accumulation steps & 4 \\
$N_\text{GPU}$ & number of GPUs (ranks) & 16 \\
$N_\text{bwd}$ & backward steps per rank ($\frac{N_\text{samples}}{N_\text{batch}N_\text{accum}N_\text{GPU}}$) & 7812 \\
\midrule
-- & Optimiser & AdamW \citep{loshchilov_decoupled_2019} \\
$\eta_\text{min}$ & Initial learning rate & 0.0 \\
$\eta^\text{base}_\text{max}$ & peak LR for pre-existing parameters & \num{5e-5} \\
$\eta^\text{new}_\text{max}$ & peak LR for new parameters & \num{1e-3} \\
$\eta^\text{base}_\text{end}$ & final LR for pre-existing parameters & \num{5e-5} \\
$\eta^\text{new}_\text{end}$ & final LR for new parameters & \num{5e-5} \\
$N_\text{warmup}$ & warm-up (backward) steps & 781 (10\%) \\
\midrule
$N_\text{B}$ & replay buffer capacity (per rank) & 16 \\
$p_\text{inject}$ & fresh sample injection rate & \num{5e-3} \\
$K_\text{max}$ & maximal rollout depth & 12 (=\qty{72}{\hour}) \\
$S$ & average SDE steps & 10 \\
$j$ & $\dtau$ sampling jitter & 0.5 \\
\bottomrule
\end{tabularx}
\end{table}

\subsection{Computational requirements}\label{app:compute_requirements}
We argue that (1) the training strategy of Xaurora is cheap compared to existing probabilistic finetuning methods, and (2) inference is also cheap relative to other generative DLWP models. 

\cref{tab:compute_requirements} shows the computational requirements of training and inference for Aurora and Xaurora models. The number of samples and training time required to convert Aurora-S into Xaurora is significantly lower than the pretraining of Aurora models. Xaurora also converts Aurora-S much faster than Aurora 1.5, which requires 4 times more samples and $12--16\times$ more GPU hours to learn the conditional distribution from Aurora-O. Furthermore, the rollout finetuning that takes Aurora-L to Aurora-O requires an additional 276k samples, which is on the same order of magnitude as the Xaurora finetuning. This allows us to call our finetuning method \textit{cheap}, in relative terms. 

Inference is more expensive with generative models than with their single-step counterparts, as the former require multiple function evaluations for a single physical rollout step (nested rollout from \cref{fig:xaurora-overview}). Thanks to the warm-start and the DSI, Xaurora can use a small number of integration steps (10) with \chgd{an Euler--Maruyama} solver, keeping the \qty{24}{\hour} forecast time \chgd{under 30 seconds} (single member). Note that solving the SDE requires some data manipulations (\emph{e.g.} concatenating tensors, moving data in and out of the GPUs), which adds some computational overhead and explains the non-linear scaling from Aurora-S to Xaurora. \chgd{We generate 15-day forecasts at \qty{6}{\hour} frequency in 7 minutes. This is faster than GenCast, which generates 15-day forecasts with \qty{12}{\hour} timesteps in 8 minutes on a single Cloud TPUv5 device. The ATLAS paper \citep{kossaifi_demystifying_2026} reports that generating a single step forecast on one A100 GPU  takes \qty{140}{\second} for GenCast, \qty{94}{\second} for ATLAS-SI, and \qty{88}{\second} for ATLAS-EDM (diffusion). This is significantly slower than Xaurora (\qty{6.7}{\second}), even with an Aurora-L prior (\qty{27.5}{\second}). These numbers cannot be directly compared to \cref{tab:compute_requirements} as the evaluation code should be matched exactly. Nevertheless, they showcase the efficiency allowed by the warm-start and the DSI in the Xaurora model.}

\begin{table}
\centering
\caption{\chgd{\textbf{Computational requirements of training and running Aurora and Xaurora models.} Training times only include pretraining for Aurora-S and Aurora-L. GPUs are 40GB A100. Aurora and Aurora 1.5 numbers are taken from \citet{bodnar_foundation_2025,weyn_aurora_2026}. Aurora 1.5 numbers indicate additional finetuning over Aurora-O. The last columns shows an untrained Xaurora model initialised with Aurora-L, hence the training numbers are not provided. We also show the number of function evaluations (NFE). Xaurora uses 10 Euler--Maruyama steps.}}
\label{tab:compute_requirements}
\footnotesize
\begin{tabularx}{\linewidth}{lXXXXX}
\toprule
 & \textbf{Aurora-L} & \textbf{Aurora-S} & \textbf{Aurora 1.5} & \textbf{Xaurora} & \textbf{Xaurora init. with Aurora-L} \\
 \midrule
 Training samples & 3.8M & 3.8M & 2.1M & 500k & N/A \\ 
 Training time  & 2 weeks & unknown & 2 weeks & 42 hours & N/A \\
 Number of training GPUs & 32 & unknown & 24 to 32 & 16 & N/A \\
 \midrule
 \qty{6}{\hour} forecast time (in seconds) & 2.0 & 0.6 & unknown & \chgd{6.7} & \chgd{27.5} \\
 \qty{24}{\hour} forecast time (in seconds) & 8.2 & 2.4 & unknown & \chgd{26.8} & \chgd{110.0} \\
 NFE per \qty{6}{\hour}-step & 1 & 1 & 1 & \chgd{10} & \chgd{10} \\
 \bottomrule
\end{tabularx}
\end{table}

\section{Supplementary results}\label{app:results}

\subsection{Additional metrics}\label{app:results/additional_metrics}
\begin{figure}
    \centering
    \includegraphics[width=1\linewidth]{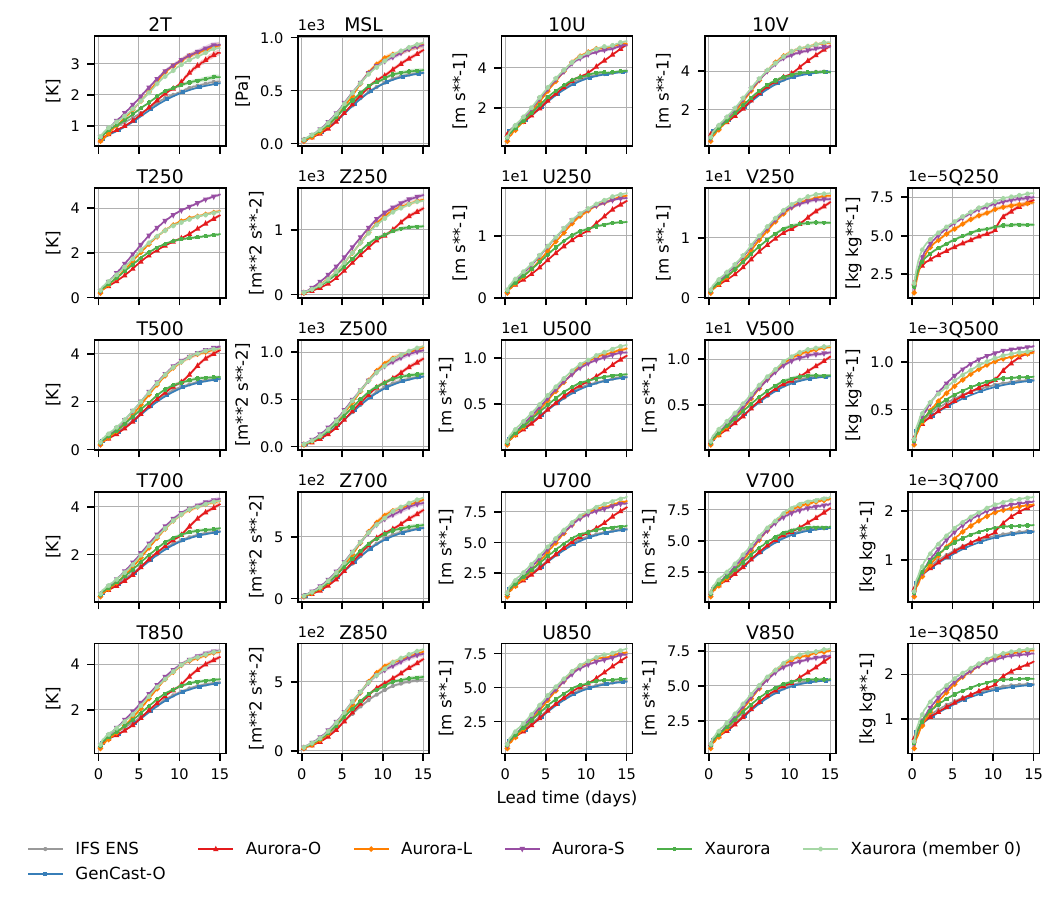}
    \caption{\rebuttal{\textbf{Global medium-range weather forecasting RMSE (uRMSE for ensemble models) for all recorded variables in 2022 at $0.25^\circ$.}}}
    \label{fig:all_rmse}
\end{figure}

\begin{figure}
    \centering
    \includegraphics[width=1\linewidth]{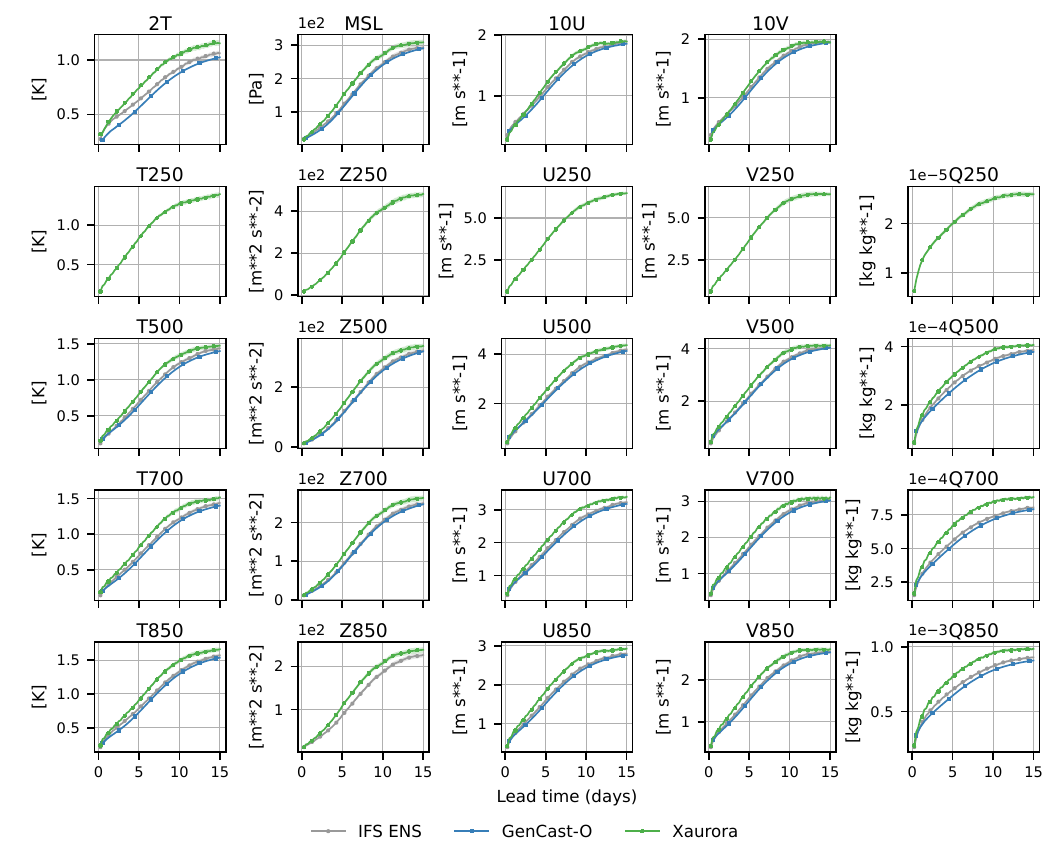}
    \caption{\rebuttal{\textbf{Global medium-range weather forecasting CRPS for all recorded variables in 2022 at $0.25^\circ$.}}}
    \label{fig:all_crps}
\end{figure}

\begin{figure}
    \centering
    \includegraphics[width=1\linewidth]{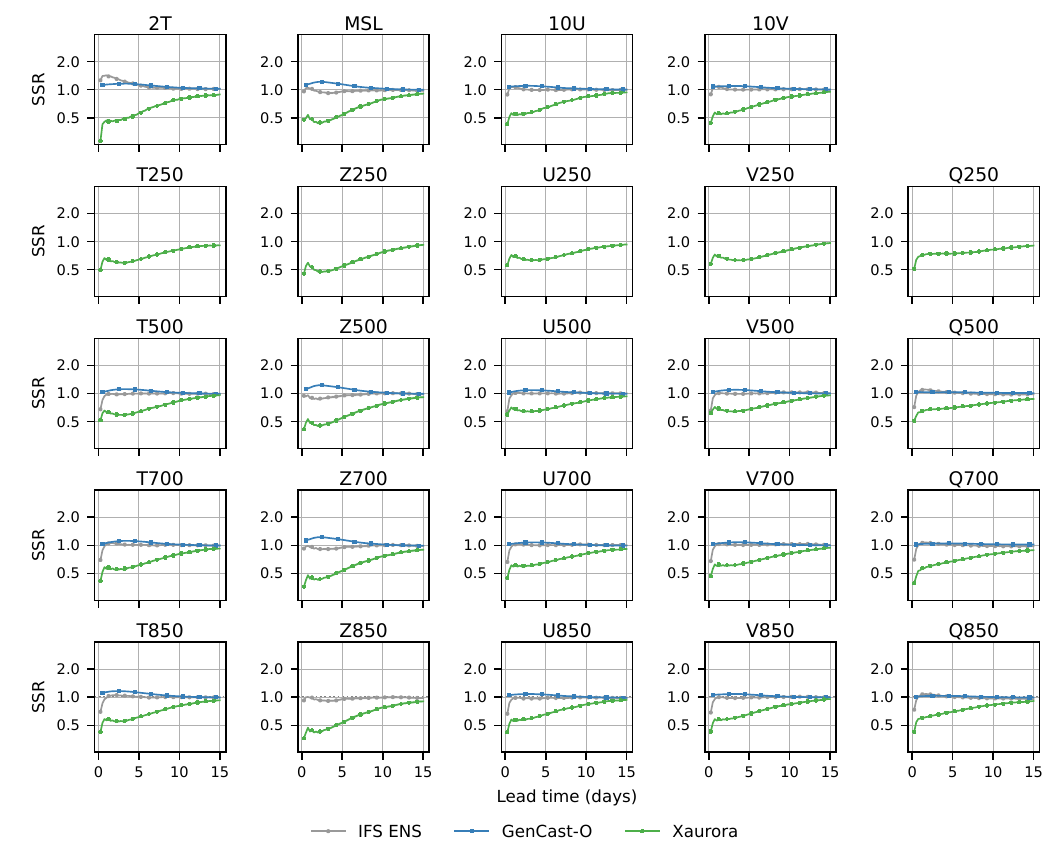}
    \caption{\rebuttal{\textbf{Global medium-range weather forecasting SSR for all recorded variables in 2022 at $0.25^\circ$.}}}
    \label{fig:all_ssr}
\end{figure}

\begin{figure}
    \centering
    \includegraphics[width=1\linewidth]{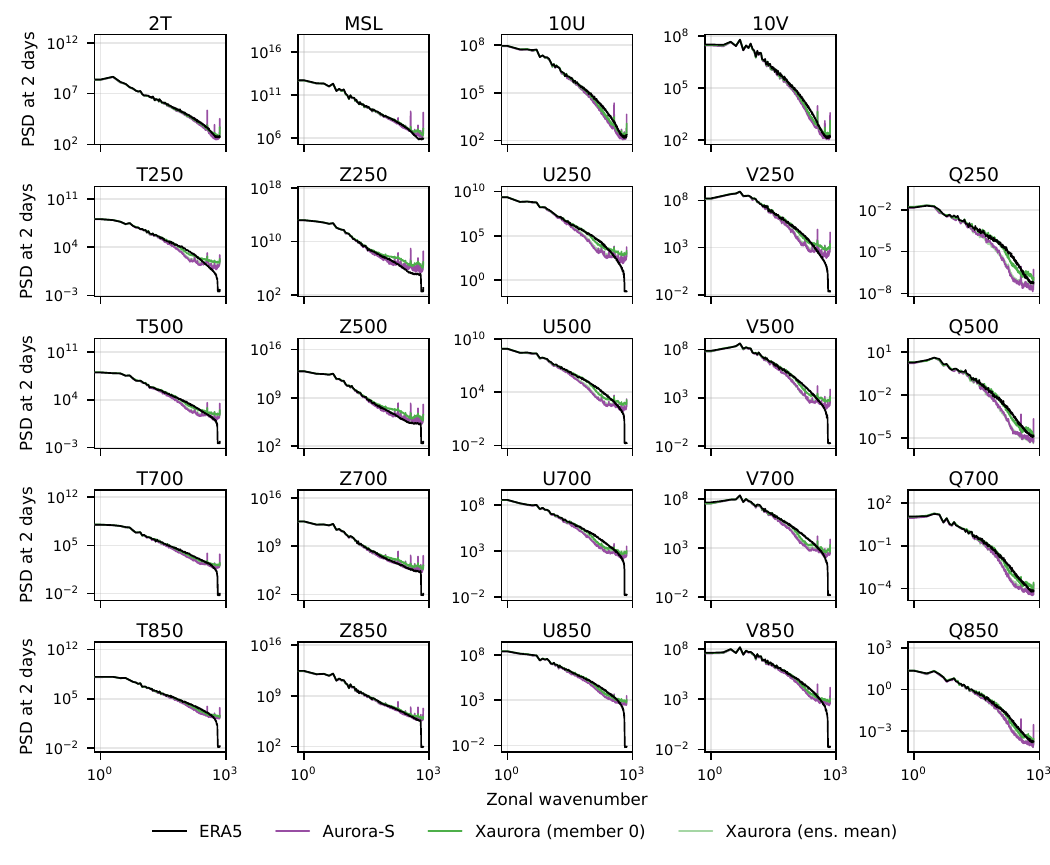}
    \caption{\chgd{\textbf{2-day PSD for all recorded variables in 2022 at $0.25^\circ$.}}}
    \label{fig:all_psd_2d}
\end{figure}

\begin{figure}
    \centering
    \includegraphics[width=1\linewidth]{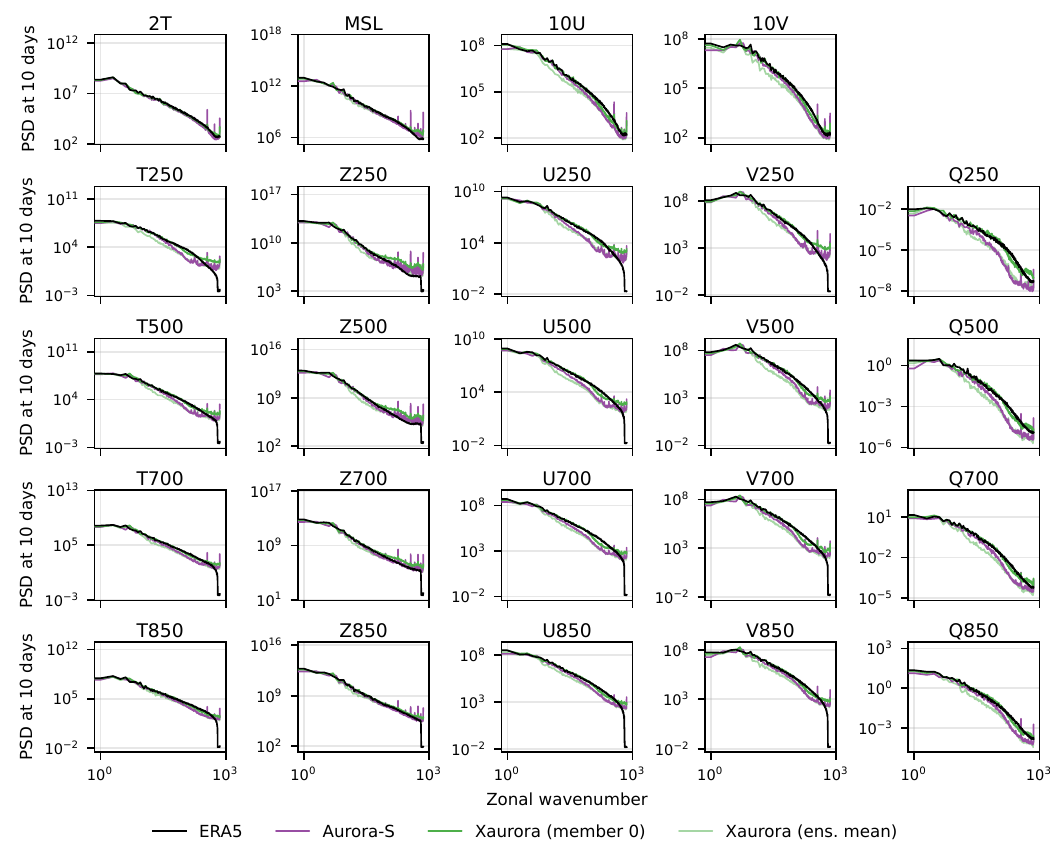}
    \caption{\chgd{\textbf{10-day PSD for all recorded variables in 2022 at $0.25^\circ$.}}}
    \label{fig:all_psd_10d}
\end{figure}

\begin{figure}
    \centering
    \includegraphics[width=1\linewidth]{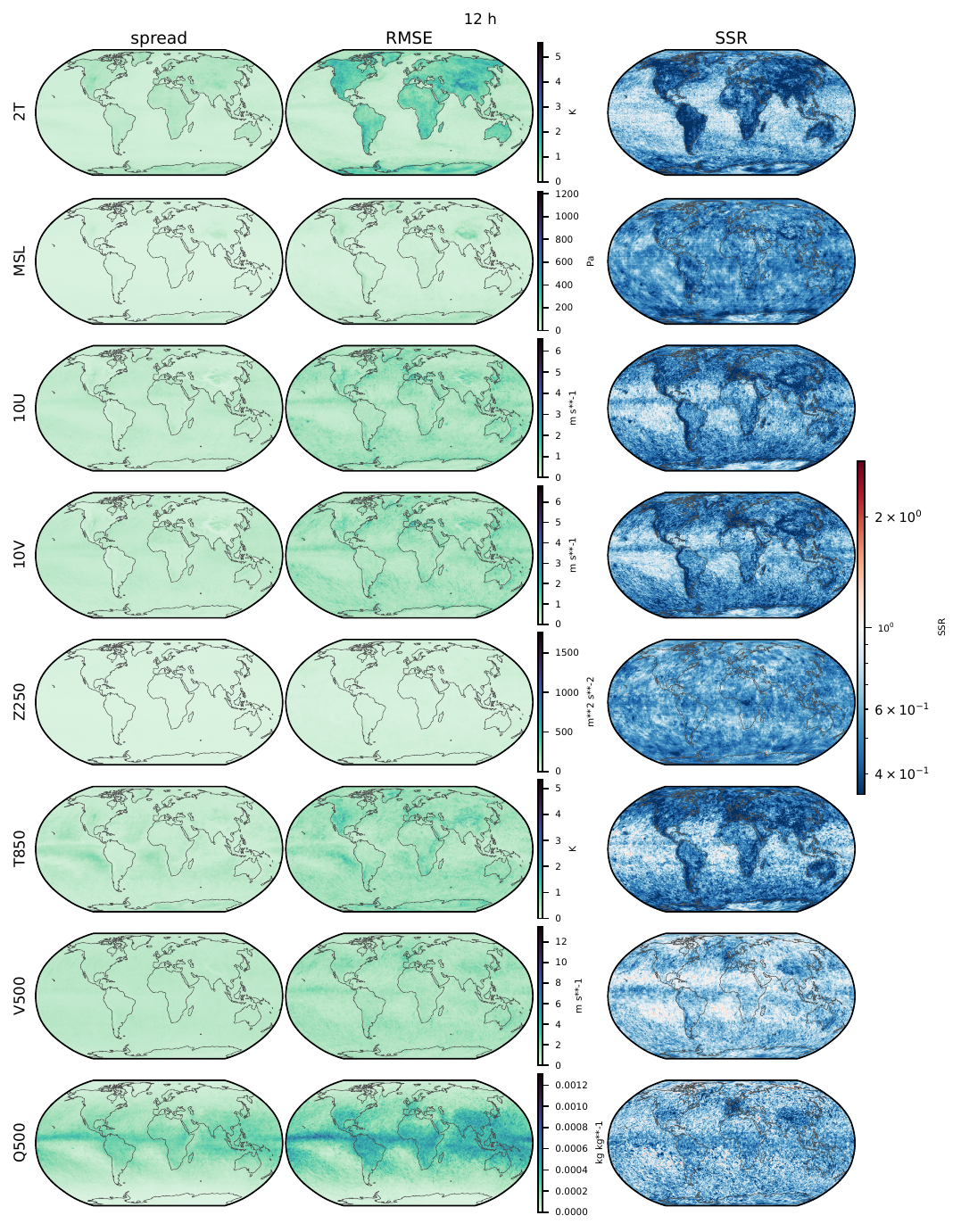}
    \caption{\chgd{\textbf{Calibration maps at \qty{12}{\hour} for a selection of variables.}}}
    \label{fig:many_calibration_maps_12h}
\end{figure}

\begin{figure}
    \centering
    \includegraphics[width=1\linewidth]{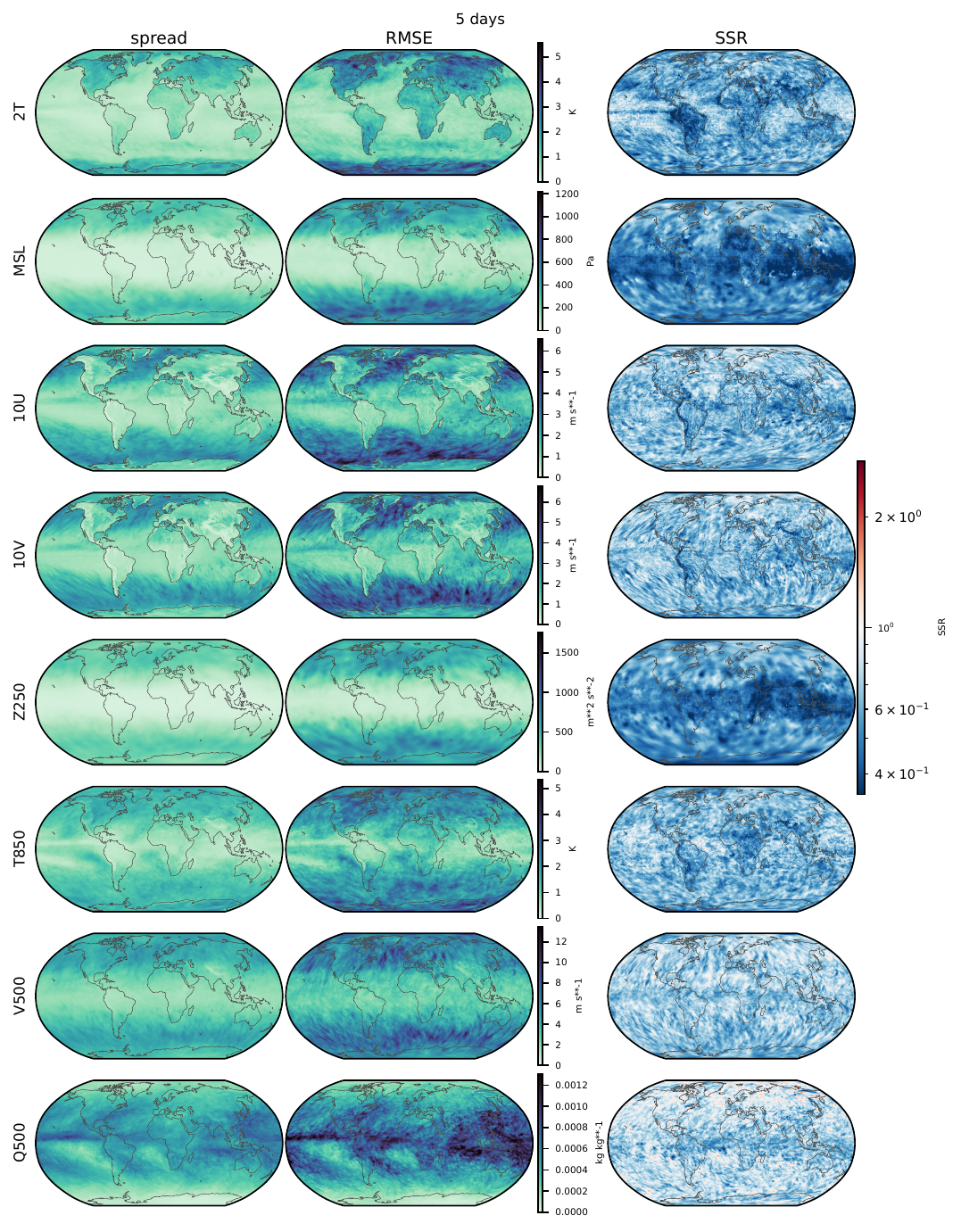}
    \caption{\chgd{\textbf{Calibration maps at 5 days for a selection of variables.}}}
    \label{fig:many_calibration_maps_120h}
\end{figure}

\begin{figure}
    \centering
    \includegraphics[width=1\linewidth]{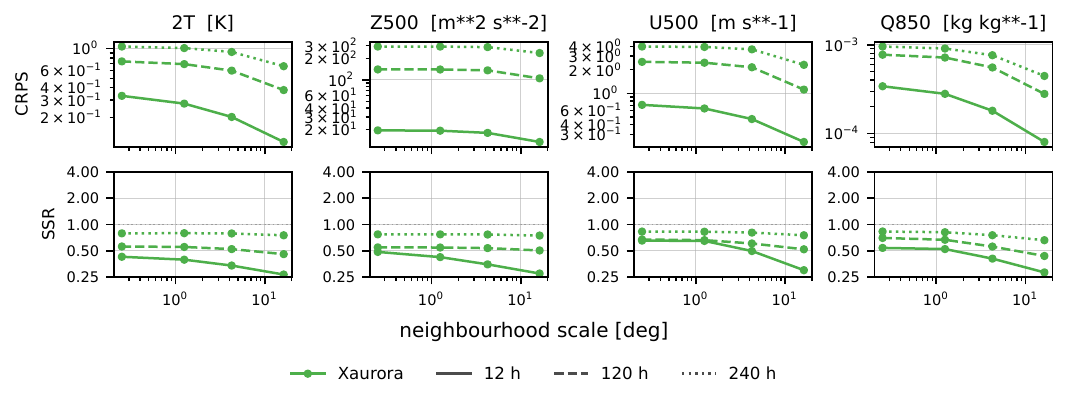}
    \caption{\chgd{\textbf{Spatially pooled calibration metrics at different lead times.} Across lead times (shown in different line styles), pooling generally improves the CRPS score (top row) and tends to decrease the SSR (bottom row).}}
    \label{fig:pooled_metrics}
\end{figure}

\begin{figure}
    \centering
    \includegraphics[width=1\linewidth]{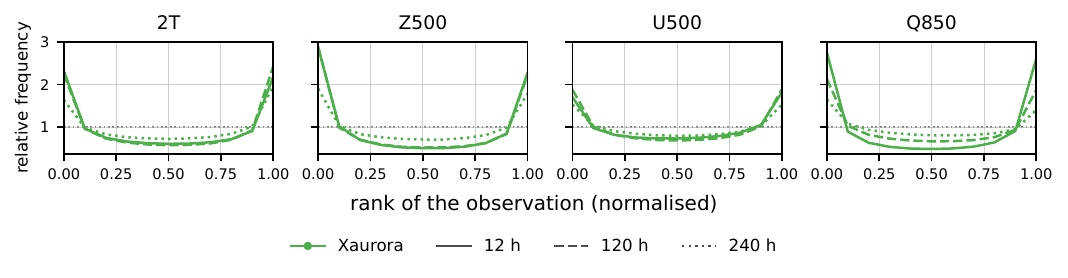}
    \caption{\chgd{\textbf{Rank histograms at different lead times.} All variables are underdispersive ($\cup$-shaped), most strongly on short lead times.}}
    \label{fig:rank_histograms}
\end{figure}

\cref{fig:all_rmse,fig:all_crps,fig:all_ssr,fig:all_psd_2d,fig:all_psd_10d} display the RMSE (uRMSE for ensemble models), CRPS, SSR, and PSD (2-day and 10-day) for all the recorded variables. \rebuttal{The metrics plots have confidence intervals bootstrapped over the initialisations, except for GenCast-O (see \cref{app:method/eval/baselines})}. \cref{fig:many_calibration_maps_12h,fig:many_calibration_maps_120h} show calibration maps for a selection of variables at \qty{12}{\hour} and 5 days. Finally, \cref{fig:pooled_metrics} and \cref{fig:rank_histograms} show spatially pooled metrics and rank histograms, respectively.

\rebuttal{\cref{tab:wb_full_compare} compares Xaurora with all models available on WeatherBench at the time of writing. The conclusions of the main paper are unaffected.}

\begin{table}[t]
\centering
\footnotesize
\setlength{\tabcolsep}{2pt}

\caption{CRPS skill [\%] against IFS ENS, 100 (1 - CRPS / CRPS$_{\mathrm{IFS}}$), with IFS ENS on the same grid and test year; positive values: better than IFS ENS. Rows are grouped by scoring grid and test year, and only models within a group are directly comparable. Bold: best per column within each group. IFS ENS and models marked $^\dagger$ are scored against the HRES analysis, the others against ERA5. PP stands for post-processed. --: variable not available.}
\label{tab:wb_full_compare}
\begin{tabular}{l*{12}{d}}
\toprule
 & \multicolumn{3}{c}{2T} & \multicolumn{3}{c}{Z500} & \multicolumn{3}{c}{U500} & \multicolumn{3}{c}{Q850} \\
\cmidrule(lr){2-4} \cmidrule(lr){5-7} \cmidrule(lr){8-10} \cmidrule(lr){11-13}
Model & \hd{3 d} & \hd{5 d} & \hd{10 d} & \hd{3 d} & \hd{5 d} & \hd{10 d} & \hd{3 d} & \hd{5 d} & \hd{10 d} & \hd{3 d} & \hd{5 d} & \hd{10 d} \\
\midrule
\multicolumn{13}{l}{\textit{0.25$^\circ$, 2022}} \\
\quad Xaurora & -11.6 & -16.1 & -12.4 & -27.3 & -21.4 & -11.5 & -15.4 & -14.0 & -9.6 & -16.4 & -15.8 & -11.9 \\
\quad Xaurora (PP) & -0.5 & -7.0 & -9.6 & -12.4 & -10.1 & -7.9 & -8.4 & -8.0 & -7.5 & -9.9 & -11.2 & -10.5 \\
\quad \mc{FGN-O$^\dagger$}{\citep{alet_skillful_2025}} & \bd{23.0} & \bd{18.3} & \bd{8.3} & \bd{21.9} & \bd{17.6} & \bd{7.1} & \bd{13.6} & \bd{12.7} & \bd{5.7} & \bd{13.0} & \bd{11.8} & \bd{6.4} \\
\quad \mc{GenCast-O$^\dagger$}{\citep{price_gencast_2023}} & 17.4 & 12.2 & 5.6 & 6.0 & 5.0 & 2.6 & 3.6 & 4.3 & 2.3 & 7.4 & 6.8 & 4.1 \\
\addlinespace[2pt]
\multicolumn{13}{l}{\textit{0.25$^\circ$, 2020}} \\
\quad \mc{GenCast}{\citep{price_gencast_2023}} & 18.6 & 12.1 & \bd{4.1} & 6.8 & 5.7 & \bd{3.3} & \bd{5.1} & \bd{5.4} & 3.3 & \bd{9.7} & 7.0 & 2.6 \\
\quad \mc{Astra}{\citep{choi_efficient_2026}} & \bd{19.3} & \bd{12.3} & 4.0 & \bd{8.4} & \bd{6.0} & \bd{3.3} & \bd{5.1} & \bd{5.4} & \bd{3.5} & 9.3 & \bd{7.2} & \bd{2.9} \\
\addlinespace[2pt]
\multicolumn{13}{l}{\textit{1.5$^\circ$, 2020}} \\
\quad \mc{NeuralGCM ENS}{\citep{kochkov_neural_2024}} & \hd{--} & \hd{--} & \hd{--} & \bd{5.7} & \bd{5.7} & \bd{3.4} & \bd{3.7} & \bd{5.3} & 3.3 & 1.4 & 1.7 & -0.8 \\
\quad \mc{ArchesWeatherGen}{\citep{couairon_archesweather_2024}} & \bd{8.4} & \bd{3.9} & -0.6 & 4.2 & 3.6 & 3.2 & 2.1 & 4.1 & 3.3 & \bd{5.9} & 4.4 & 0.5 \\
\quad \mc{Sonny ENS}{\citep{cheon_sonny-ens_2026}} & 5.3 & 3.0 & \bd{-0.4} & 5.0 & 4.4 & 3.0 & 3.3 & 5.0 & \bd{3.5} & 5.4 & \bd{4.5} & \bd{1.0} \\
\bottomrule
\end{tabular}
\end{table}

\subsection{Aurora scorecards}\label{app:results/aurora_scorecards}
\begin{figure}
    \centering
    \includegraphics[width=1\linewidth]{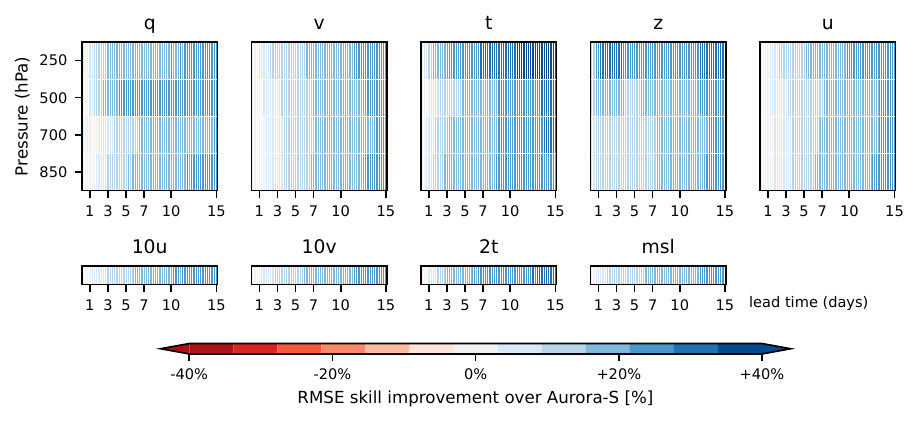}
    \caption{\chgd{\textbf{Xaurora vs Aurora-S RMSE scorecard.} Each panel shows the RMSE improvement percentage from Xaurora over Aurora-S for one variable across lead times and pressure levels (for atmospheric variables). We use uRMSE for Xaurora (unbiased) and the plain RMSE for Aurora-S. Red indicates degraded performance (negative improvement), blue shows improvement over the reference.}}
    \label{fig:xaurora_vs_aurora_s_scorecard}
\end{figure}

\begin{figure}
    \centering
    \includegraphics[width=1\linewidth]{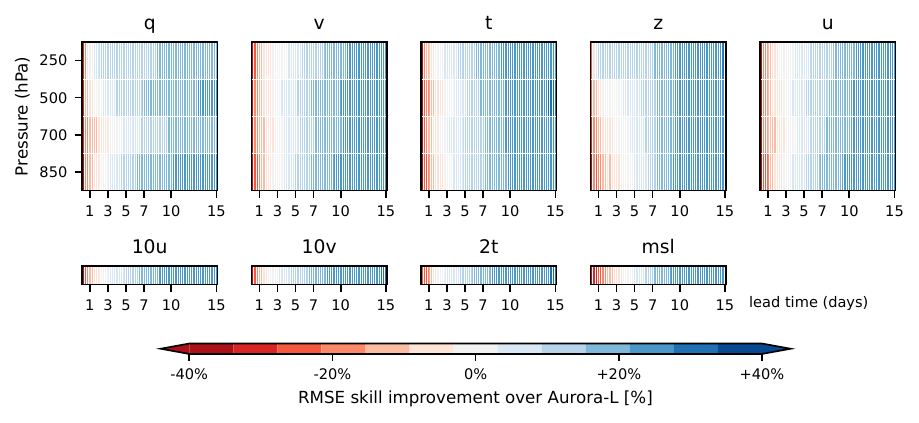}
    \caption{\chgd{\textbf{Xaurora vs Aurora-L RMSE scorecard.} Reads as \cref{fig:xaurora_vs_aurora_s_scorecard}.}}
    \label{fig:xaurora_vs_aurora_l_scorecard}
\end{figure}

Scorecards comparing Xaurora with Aurora-S (\cref{fig:xaurora_vs_aurora_s_scorecard}) and Aurora-L (\cref{fig:xaurora_vs_aurora_l_scorecard}) show the superiority of our model for medium-range weather forecasting. The benefits of Xaurora over Aurora-S are mostly apparent after 1 day, where most variables and lead times are outperformed. Aurora-L is generally better up to about 3 days, after which it falls short against Xaurora. We deliberately use the uRMSE (approximating the infinite-member ensemble mean error) since we want to characterise the maximal gain from Xaurora over Aurora models. The influence of the ensemble size is evaluated in \cref{app:results/ablations/ens_size}.

\subsection{Forecast visualisations}\label{app:results/forecast_viz}
\begin{figure}
    \centering
    \includegraphics[width=1\linewidth]{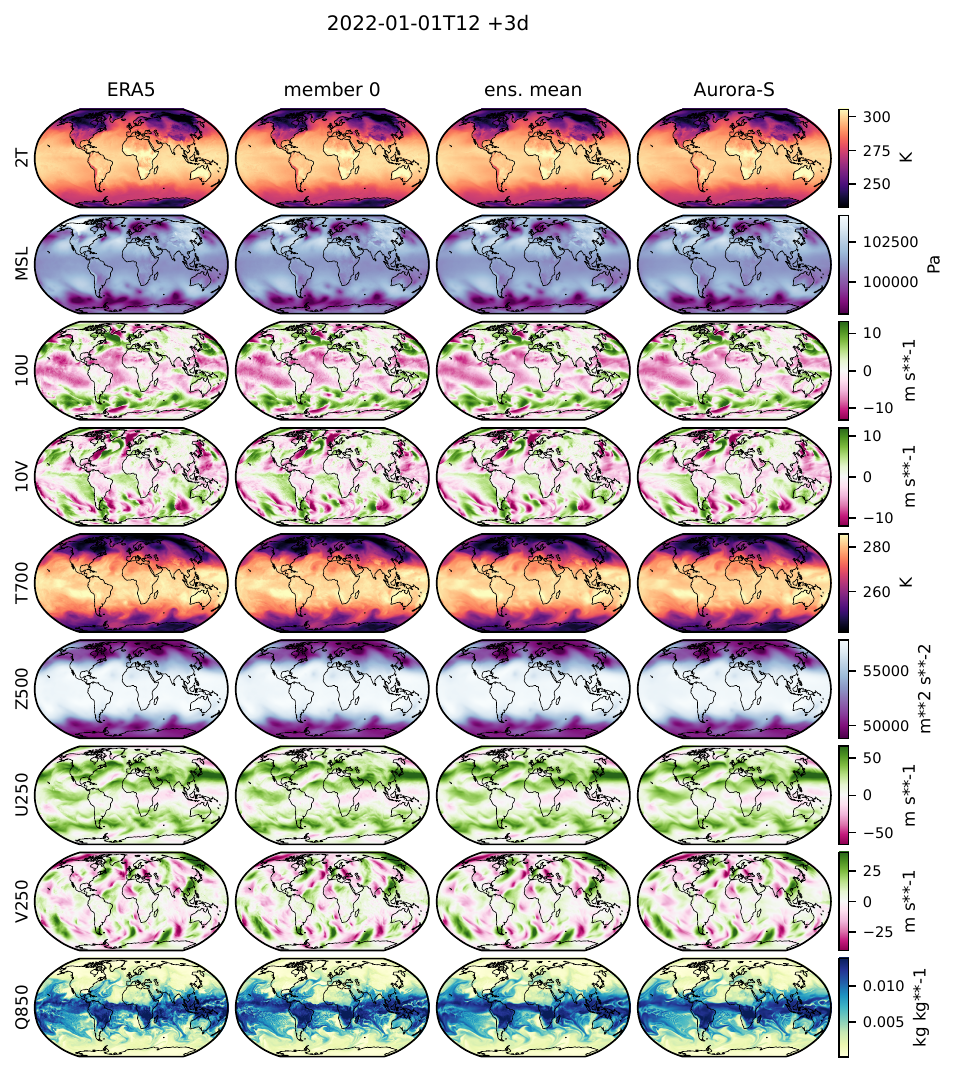}
    \caption{\chgd{\textbf{3-day rollout initialised on January 1st, 2022 at 12:00~UTC.} Columns correspond to ERA5, Xaurora first member (member 0), Xaurora ensemble mean (10 members), and Aurora-S. Variables (unnormalised) are shown on each row.}}
    \label{fig:mosaic_3d}
\end{figure}

\begin{figure}
    \centering
    \includegraphics[width=1\linewidth]{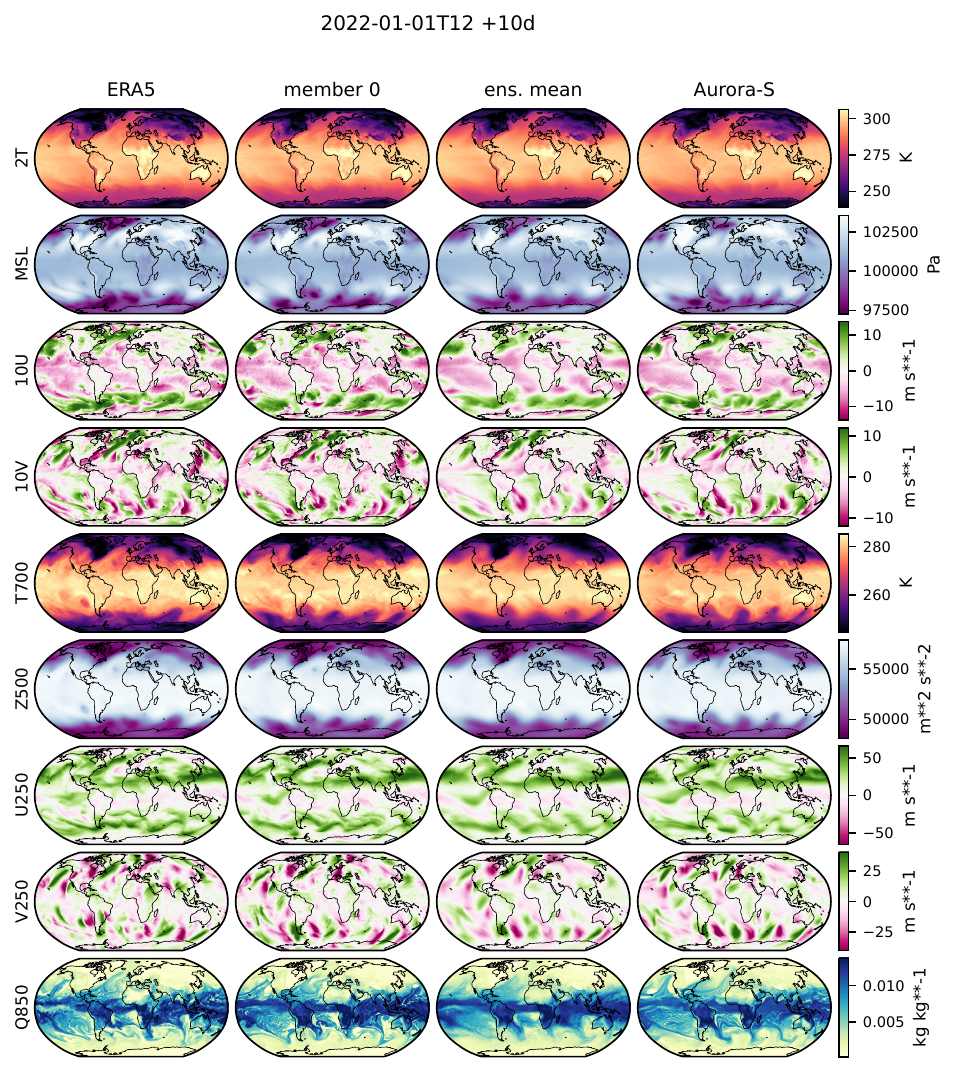}
    \caption{\chgd{\textbf{10-day rollout initialised on January 1st, 2022 at 12:00~UTC.} Columns correspond to ERA5, Xaurora first member (member 0), Xaurora ensemble mean (10 members), and Aurora-S. Variables (unnormalised) are shown on each row.}}
    \label{fig:mosaic_10d}
\end{figure}

\begin{figure}
    \centering
    \includegraphics[width=1\linewidth]{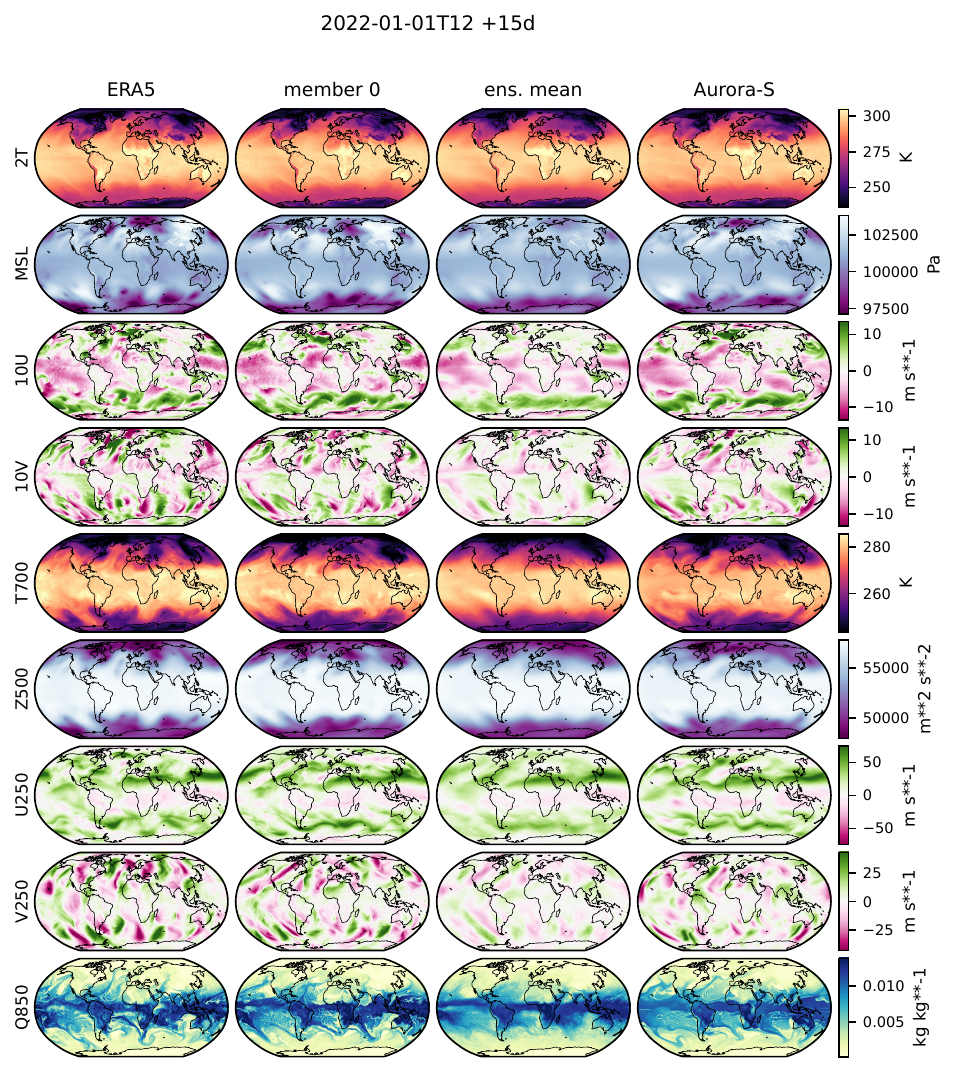}
    \caption{\chgd{\textbf{15-day rollout initialised on January 1st, 2022 at 12:00~UTC.} Columns correspond to ERA5, Xaurora first member (member 0), Xaurora ensemble mean (10 members), and Aurora-S. Variables (unnormalised) are shown on each row.}}
    \label{fig:mosaic_15d}
\end{figure}

\cref{fig:mosaic_3d,fig:mosaic_10d,fig:mosaic_15d} show a wide selection of variables and pressure levels for a forecast initialised on January 1st, 2022 at 12:00~UTC.

\subsection{Importance of subtracting the residual white noise}\label{app:results/importance_subtract_noise}
\begin{figure}
    \centering
    \includegraphics[width=1\linewidth]{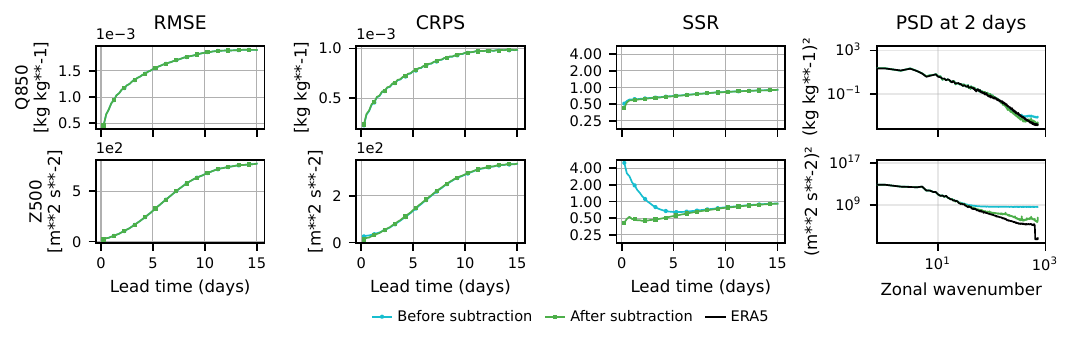}
    \caption{\chgd{\textbf{Effect of residual white noise subtraction.} RMSE (unbiased), CRPS, SSR and PSD (first member at two days) for Q850 and Z500. Subtracting the residual noise has a limited impact on RMSE and CRPS. However, the noisy predictions (cyan) are highly overdispersive and over-powered in the high frequencies, whereas the filtered predictions (green) are consistently underdispersed and resolve all frequencies.}}
    \label{fig:importance_subtract_noise}
\end{figure}

\begin{figure}
    \centering
    \includegraphics[width=\linewidth]{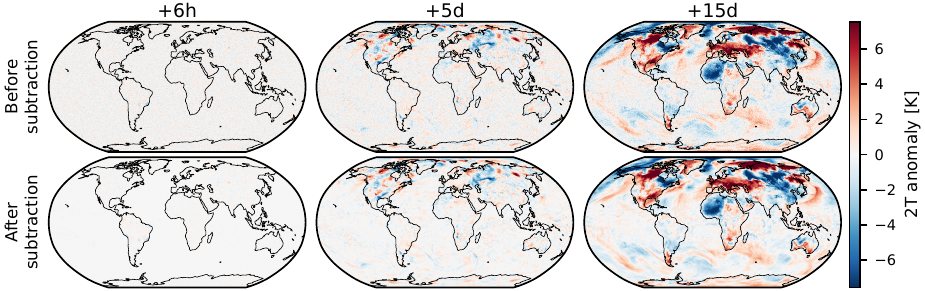}
    \caption{\chgd{\textbf{2T anomaly of the first ensemble member.} Anomaly relative to the ensemble mean at \qty{6}{\hour} (left), \qty{5}{\day} (middle) and \qty{15}{\day} (right) before (top) and after (bottom) residual noise subtraction. Large-scale features are maintained and the background noise visible in the top row panels is removed after subtraction.}}
    \label{fig:noie_subtraction_anomaly_maps}
\end{figure}

\chgd{As mentioned in \cref{app:method/eval/forecast_settings}, an Euler--Maruyama solver with $S$ integration steps retains residual white noise in the terminal state with variance $\gamma^2_{1-1/S}\cdot{1/S}$. We remove this noise in the saved forecasts by subtracting the exact last Wiener sample used by the solver. \cref{fig:importance_subtract_noise} shows a selection of metrics and spectra before and after subtracting the noise, highlighting the importance of this step. Anomaly maps in \cref{fig:noie_subtraction_anomaly_maps} show the effect of noise subtraction on the first ensemble member: low-amplitude high-frequency residual white noise is filtered out, while genuine spread is maintained.}

\subsection{Spectral post-processing}\label{app:results/post_processing}

\subsubsection{Factorised calibration procedure}\label{app:results/post_processing/procedure}
\chgd{With $s$ the lead time and $\ell$ the spherical harmonics degree, let $S(\ell,s)$, $\bar{E}(\ell,s)$ and $E_u(\ell,s) = \bar{E}(\ell,s) - S(\ell,s)/M$ be the spread, the ensemble mean error and the unbiased error. After calibration using \cref{eq:post_processing_core}, the spread and unbiased error become}
\begin{equation}
    S^\text{cal}(\ell,s) = \mathcal{G}^2(\ell)\mathcal{H}^2(s)S(\ell,s),\quad E_u^\text{cal}(\ell,s) = \bar{E}(\ell,s)-\mathcal{G}^2(\ell)\mathcal{H}^2(s)S(\ell,s)/M.
\end{equation}
\chgd{To avoid negative errors, we set $E_u^\text{cal} = \frac{M}{M+1}\bar E$ and mask $\ell$ such that $E_u^\text{cal}(\ell)< 0$. We also group low degrees to reduce the variance estimation error. We fit $\mathcal{G}(\ell)$ at $s_\text{fit}=\qty{48}{\hour}$. Setting $\text{SSR}^\text{cal}(s_\text{fit}) = 1 \iff S^\text{cal}(\ell,s_\text{fit}) = E_u^\text{cal}(\ell,s_\text{fit}) \iff \mathcal{G}^2(\ell)\underbrace{\mathcal{H}^2(s_\text{fit})}_{=1}S(\ell,s_\text{fit}) = \frac{M}{M+1}\bar{E}(\ell,s_\text{fit})$ yields:}
\begin{equation}
    \mathcal{G}(\ell) = \sqrt{\frac{M}{M+1}\cdot\frac{\bar{E}(\ell,s_\text{fit})}{S(\ell,s_\text{fit})}}.
\end{equation}
\chgd{Then, we fix $\mathcal{G}$ and set $\text{SSR}^\text{cal}(s) = \sqrt{\frac{\sum_\ell (2\ell+1)S^\text{cal}(\ell,s)}{\sum_\ell (2\ell+1)E_u^\text{cal}(\ell,s)}}=1$ for all $s$. Solving for $\mathcal{H}$ gives}
\begin{equation}
    \mathcal{H}(s)=\sqrt{\frac{M}{M+1}\cdot\frac{\sum_\ell(2\ell+1)\bar{E}(\ell,s)}{\sum_\ell(2\ell+1)\mathcal{G}^2(\ell)S(\ell,s)}}.
\end{equation}
\chgd{In practice, we compute the spread and error on the validation set (2021) and apply to the test set (2022) to evaluate the generalisability of the calibration. For both splits, we run a full 56-initialisations forecast with 10 ensemble members according to \cref{app:method/eval/forecast_settings}.}
\begin{figure}
    \centering
    \includegraphics[width=1\linewidth]{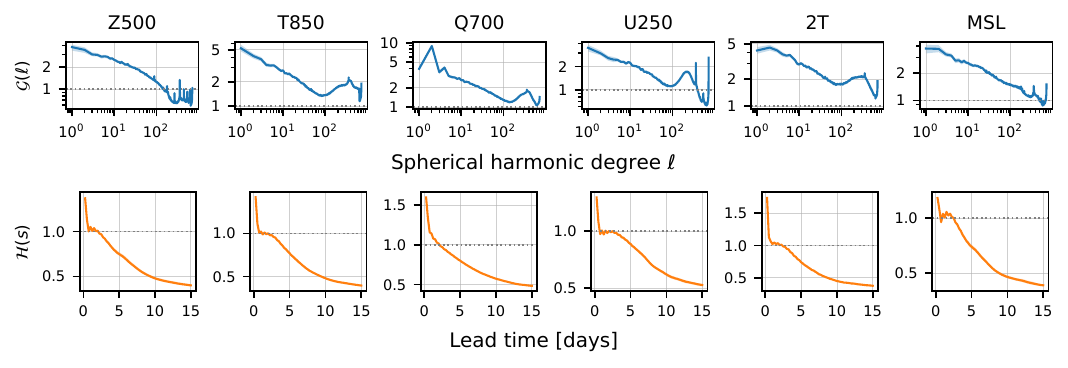}
    \caption{\chgd{\textbf{Calibration functions fitted on 2021.} Shape function $\mathcal{G}(\ell)$ (top) and per-lead correction factor $\mathcal{H}(s)$ (bottom) for a selection of surface and atmospheric variables.}}
    \label{fig:G_H_funcs}
\end{figure}

\begin{figure}
    \centering
    \includegraphics[width=1\linewidth]{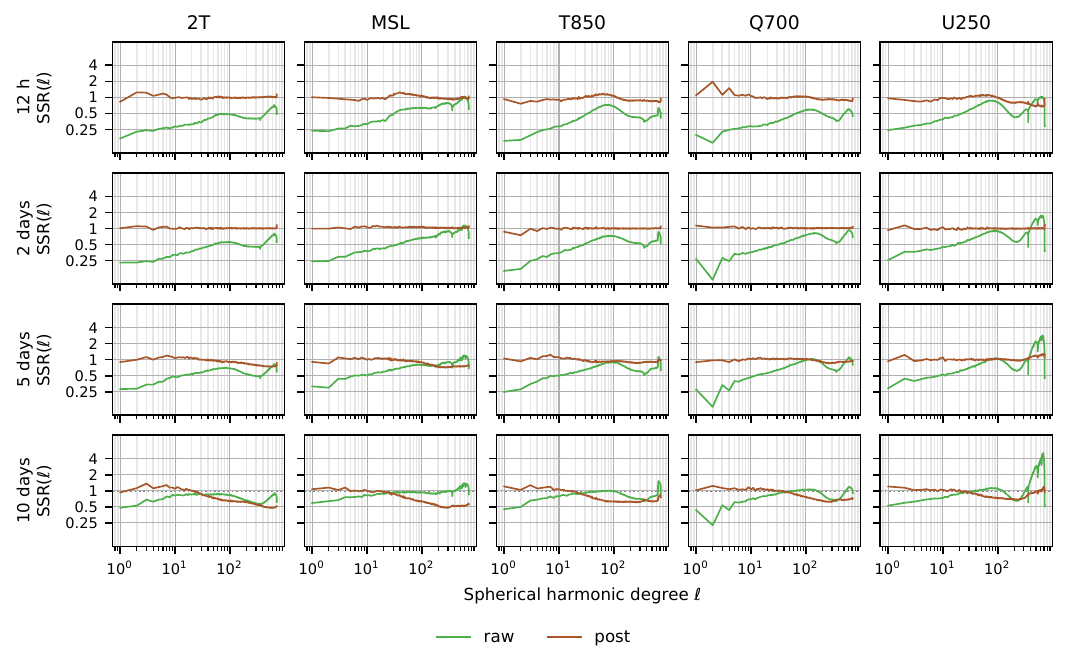}
    \caption{\chgd{\textbf{Spherical harmonics decomposition of the SSR before and after post-processing.} For each variable, the raw SSR (green) displays a similar spectral shape with level depending on the lead time. The post-processed ensemble (brown) has a flatter SSR across modes and variables. The flattening effect is best at $s_\text{fit}=\qty{48}{\hour}$. At 10 days, small scales tend to become slightly underdispersive. Results are shown on the test set (2022).}}
    \label{fig:ssr_l_more_vars}
\end{figure}

\subsubsection{Effect on ensemble members}\label{app:results/postprocessing/effect_on_members}
\begin{figure}
    \centering
    \includegraphics[width=1\linewidth]{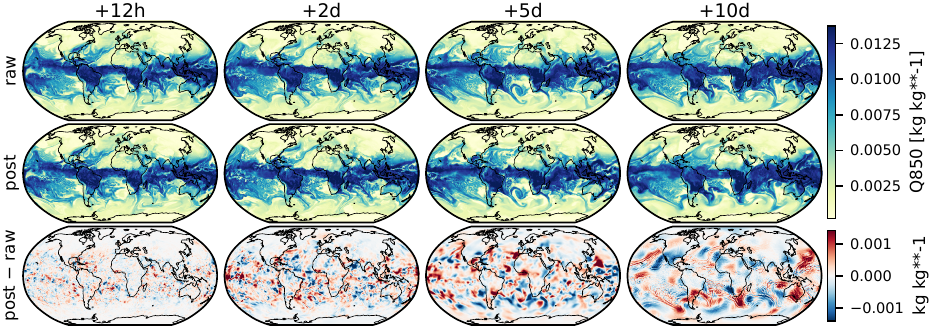}
    \caption{\chgd{\textbf{Spatial effect of post-processing on a Q850 single member.} The first member of the forecast initialised on January 1st, 2022 at 12:00~UTC for the raw predictions (top), the post-processed predictions (middle) and their difference (bottom).}}
    \label{fig:postproc_members_q850}
\end{figure}

\begin{figure}
    \centering
    \includegraphics[width=1\linewidth]{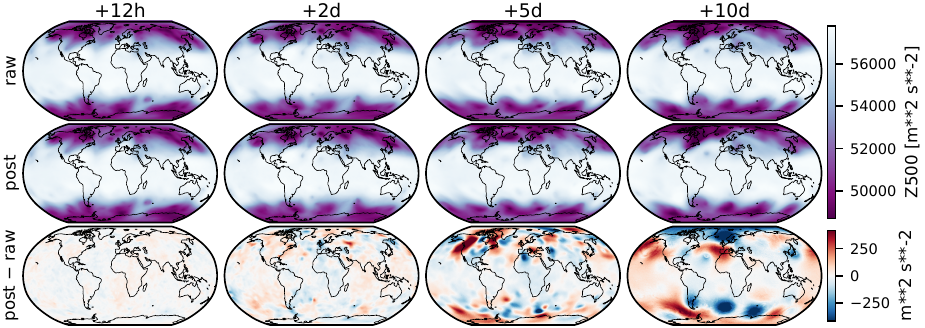}
    \caption{\chgd{\textbf{Spatial effect of post-processing on a Z500 single member.} The first member of the forecast initialised on January 1st, 2022 at 12:00~UTC for the raw predictions (top), the post-processed predictions (middle) and their difference (bottom).}}
    \label{fig:postproc_members_z500}
\end{figure}

\chgd{The maps in \cref{fig:postproc_members_q850,fig:postproc_members_z500} show the effect of the calibration. Since this is a pure post-processing step dependent on the variable, scale and lead time, dynamic structures are maintained and rescaled to increase ensemble spread. The raw and post-processed forecasts are almost visually indistinguishable. Yet, the changes are sufficient to calibrate the ensemble across lead times.}

\subsubsection{Significant skill improvement}\label{app:results/post_processing/skill_improvement}
\begin{figure}
    \centering
    \includegraphics[width=1\linewidth]{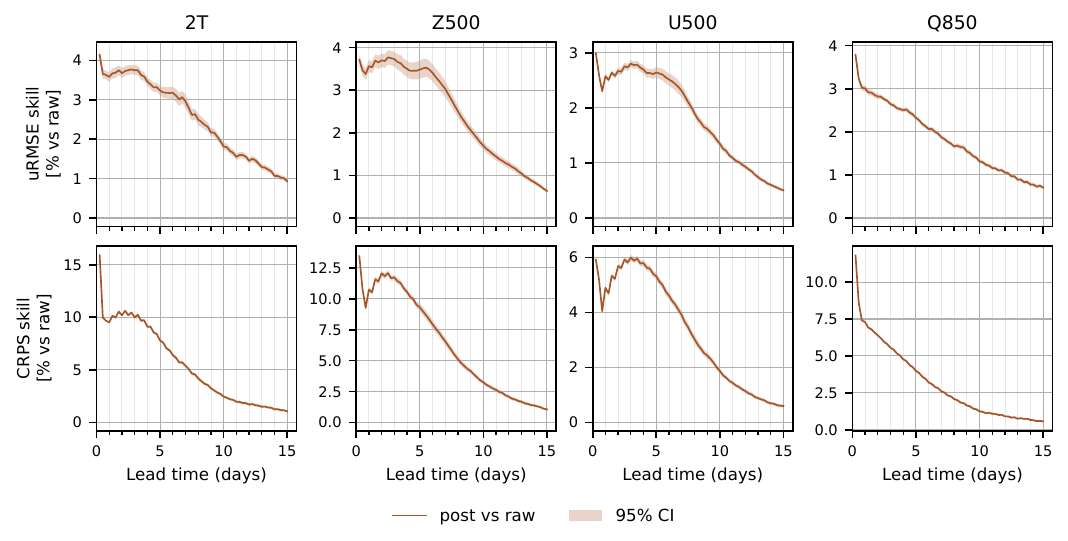}
    \caption{\chgd{\textbf{Skill of the post-processed forecasts relative to raw ensemble.} uRMSE (top) and CRPS (bottom) skill scores, calculated as $100\cdot(1 - \mathcal{M}_\text{post} / \mathcal{M}_{\text{raw}})$. Positive values indicate better performance of the post-processed predictions. Shading corresponds to the 95\% CI (see \cref{app:method/eval/significance}). The post-processing improves the CRPS significantly at all recorded lead times. The improvements are higher on short lead times where the raw ensemble is the most underdispersed. The uRMSE improvements are driven by the increase in spread since the ensemble mean is unchanged.}}
    \label{fig:postprocessing_skill_improvement}
\end{figure}

\chgd{\cref{fig:postprocessing_skill_improvement} shows significant CRPS improvements across variables and lead times. The largest improvements happen on short lead times (up to 15\% better CRPS for 2T) where the underdispersion is maximal for the raw forecasts.}

\subsection{Ablations}\label{app:results/ablations}
\subsubsection{Main paper ablations details}\label{app:results/ablations/main_paper_details}
All ablation experiments in \cref{subsec:experiments/ablations} run with the same hyperparameters as the best performing Xaurora model (see \cref{tab:hyperparam}), \chgd{including the residual noise subtraction}. The only differences are the changes detailed below.

\textbf{Optimisation methods.} The optimisation of Xaurora is based on an SDE-rollout replay buffer and a CRPS loss. We compare this setup with different configurations: 
\begin{itemize}[leftmargin=*]
    \item \textbf{Single-step + MAE:} No rollout training. We apply \cref{alg:ar_training} with the Aurora MAE loss (variable weighting, no latitude-weighting) \citep{bodnar_foundation_2025}. This model quickly overfits, achieving its best checkpoint after only 200 backward steps (12,800 samples).
    \item \textbf{Curriculum + MAE:} Curriculum rollout training approach similar to that of \citet{lang_aifscrps_2024}. Starting from the best checkpoint of the "Single-step + MAE" experiment, we perform a two-stage finetuning: (1) 2-step (\qty{12}{\hour} forecast) rollout training for 125,000 samples, and (2) linearly increasing rollout steps from 2 to 12 (\qty{12}{\hour} to \qty{72}{\hour}) over 125,000 samples. To make $N$-step rollout training computationally feasible, we performed $N-1$ gradient-less steps by estimating samples using the denoiser $\bD_\theta(\SIstate^t_0,\bh^t,\tau{=}0)\approx \E[\sistate^{t+\dt}\mid\bh^t]$, instead of solving the SDE. We then supervise the last step according to \cref{alg:ar_training}. This mechanism was efficient, but resulted in poor probabilistic performance as the conditional expectation approximation did not help the model generalise to the true inference distribution. This motivated the introduction of the replay buffer.
    \item \textbf{Replay buf. + MAE}: the replay buffer combined with an MAE loss. In this case, we store a single trajectory in the replay buffer entries that we supervise with the Aurora MAE loss. The metrics, especially the CRPS, degraded compared to "Curriculum + MAE", highlighting the importance of the probabilistic loss. 
\end{itemize}

\textbf{Parameterisations.} Xaurora leverages the Denoising Stochastic Interpolants introduced in this paper. As discussed in \cref{subsec:dsi/dsi_from_pretrained}, other parameterisations are unlikely to perform well \rebuttal{without output transformation} in the context of finetuning from a pretrained deterministic prior. We also expect the replay buffer mechanism to worsen the performance of the other parameterisations as early predictions do not fully benefit from the Aurora warm-start. Nevertheless, we train and evaluate the following Xaurora variants with the same optimisation (Replay buffer and CRPS):
\begin{itemize}[leftmargin=*]
    \item $\sistate_1$\textbf{-pred. + DDPM}: The diffusion equivalent of DSI. This parameterisation also targets the terminal state distribution and can therefore be properly initialised from Aurora, hence it shares most of the advantages of DSI in the context of finetuning deterministic priors. Yet, it falls short of Xaurora. We hypothesise that the new encoder heads used to embed the SDE state could be more difficult to train with diffusion, which feeds full noise states, than SI, which could lead to a worse or less stable local minimum. Additionally, the replay buffer algorithm depends on the number of integration steps, which is set to $S=10$, relatively low for diffusion models, potentially making the training of the DDPM network even more challenging.
    \item $\bW$\textbf{-pred.}: The parameterisation detailed in \cref{app:method/drift_param} with $\bWnet$ initialised to Aurora-S. The model is optimised with \cref{eq:xloss_in_W_space}. The training is extremely unstable, reaching infinite gradients after roughly 70 optimisation steps. Therefore, we do not run the evaluation, hence the empty row in \cref{tab:ablations}. Note that the instabilities could potentially be fixed by adding stabilisation layer norms in the encoder\footnote{As suggested in the official Aurora finetuning documentation: \url{https://microsoft.github.io/aurora/finetuning.html\#exploding-gradients}.} or by training the decoder from scratch, but this would prevent warm-starts ($\bD_{\theta^\text{init}}\not=\prior$).
    \item $\drift$\textbf{-pred.}: The vanilla SI parameterisation where the neural network predicts the drift term of the SDE, with $\drift_\theta$ initialised to Aurora-S. The network is trained on \cref{eq:xloss_in_drift_space}. This parameterisation is stable, but does not perform well, especially on probabilistic metrics.
\end{itemize}

\textbf{Aurora-L prior.} A central result of Xaurora is that it is able to outperform its prior. We show that this property scales to an Aurora-L initialisation. We initialise a Xaurora model from the 1.3B parameter Aurora version and train it with "Single-step + MAE". The resulting model outperforms its prior, Aurora-L, as shown in \cref{tab:ablations}. It is worth noting that the number of training steps is kept constant and this large version of Xaurora has not converged. Based on these results and the optimisation ablation, we expect a probabilistic replay buffer version of the large model to perform even better.

\subsubsection{Non-SDE baseline}\label{app:results/ablations/non-sde}
\rebuttal{We compare Xaurora with a non-SDE baseline. This model is trained with the same hyperparameters, loss function, replay buffer, but without the DSI. It is the CRPS-trained equivalent of Xaurora.}

\rebuttal{\textbf{Implementation details.} The non-SDE model replaces the SDE sampling with one stochastic forward pass per \qty{6}{\hour} step, $\gen{\sistate}^{t+\dt} = F_\theta(\bh^t, \bm{z}^t)$ with the noise $\bm{z}\sim\mathcal{N}(\bm{0}, \Id)$ resampled at every rollout step. Following the design of Aurora 1.5 \citep{weyn_aurora_2026}, we sample one random vector per token that enters the backbone adaptive layer-norm context as $\bm{c}_0 = \bm{c}_{\mathrm{lead time}} + \mathrm{MLP}_{\mathrm{noise}}(\bm{z})$. The last layer of the 2-layer MLP is zero-initialised to preserve the warm-start.}

\begin{figure}
    \centering
    \includegraphics[width=\linewidth]{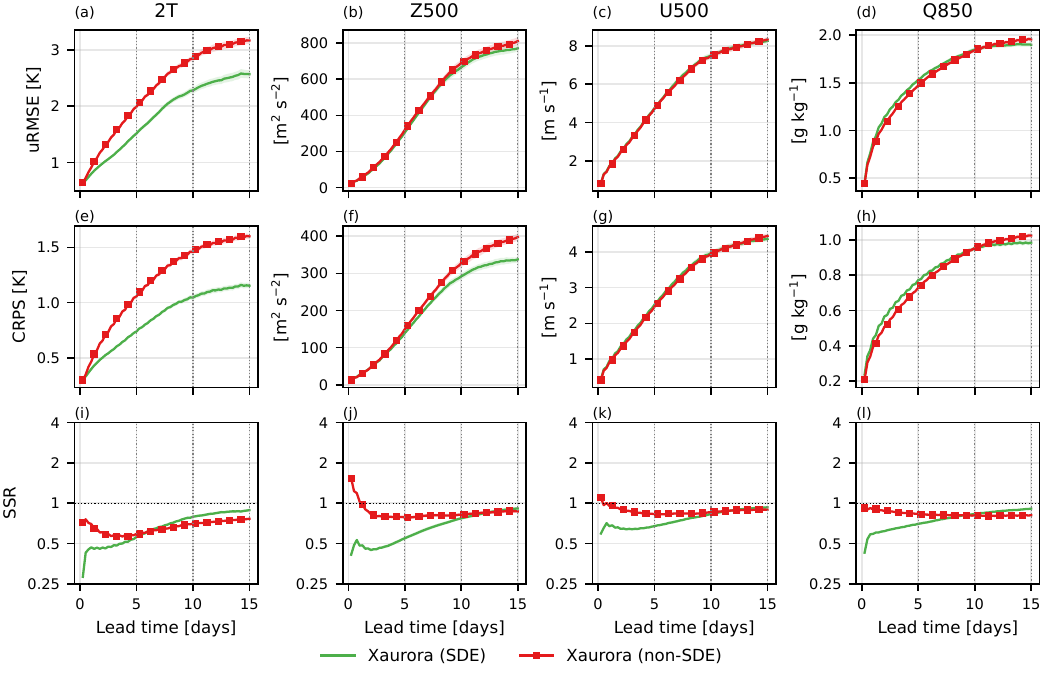}
    \caption{\rebuttal{\textbf{Medium-range comparison with non-SDE baseline.} 15-day uRMSE (a-d), CRPS (e-h), and SSR (i-l) for 2T, Z500, U500, Q850 with 56 initialisations in 2022 and 10 ensemble members. Xaurora is shown in green, the non-SDE baseline in red. The shading is the 95\% confidence interval bootstrapped over initialisations.}}
    \label{fig:non_sde_scores}
\end{figure}

\rebuttal{\textbf{Medium-range weather forecasting skill (\cref{fig:non_sde_scores}).} Xaurora is substantially better on 2T for both uRMSE and CRPS across lead times and has a smaller, but significant advantage on Z500. The models are comparable on U500 and Q850. Interestingly, the non-SDE version is overdispersive on short lead times for Z500 and, to a lesser extent, U500. Overall, both models are underdispersive. Xaurora surpasses the non-SDE version in terms of calibration at approximately 6 days for 2T, 15 days for Z500 and U500, and 10 days for Q850. At 15 days, the generative version of Xaurora is significantly better across all metrics and variables.}

\begin{figure}
    \centering
    \includegraphics[width=\linewidth]{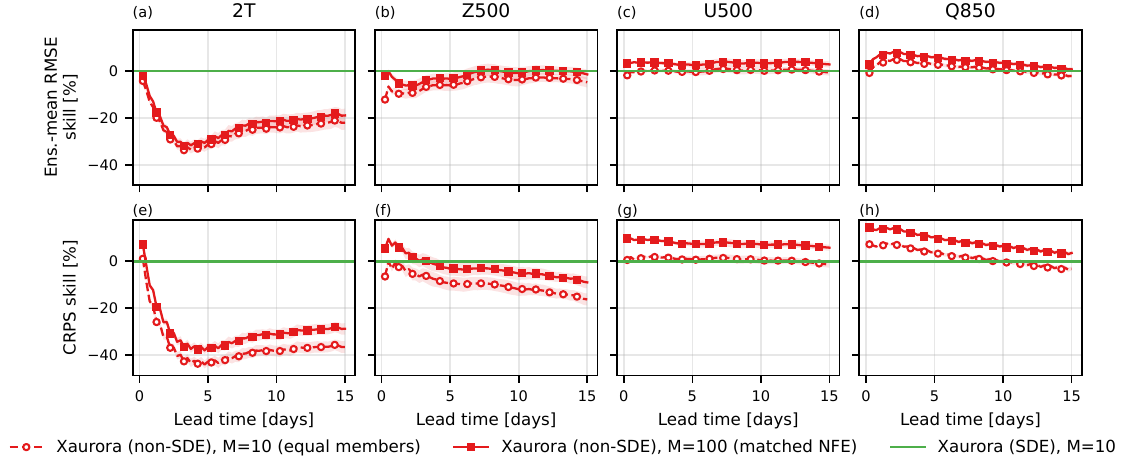}
    \caption{\rebuttal{\textbf{Comparison with non-SDE baseline at matched compute.} Finite-ensemble RMSE (a-d) and CRPS (e-h) skill of the non-SDE baseline (red) relative to 10-member Xaurora (green) for 2T, Z500, U500 and Q850. The round markers indicate the 10-member ensemble performance. Square markers estimate the 100-member metrics, matching the number of network evaluations of 10 SDE members. The shading is the 95\% confidence interval bootstrapped over initialisations.}}
    \label{fig:non_sde_matched_compute}
\end{figure}

\rebuttal{\textbf{Matched compute comparison (\cref{fig:non_sde_matched_compute}).} The non-SDE baseline requires a single network evaluation per \qty{6}{\hour} step. Hence, a 100-member ensemble is theoretically as expensive as a 10-member Xaurora ensemble. The expected scores of a 100-member ensemble follow exactly from the statistics of the 10-member evaluation, assuming members are exchangeable. The results are consistent with the previous paragraph. Xaurora is substantially better at forecasting 2T. For the other variables, RMSE values are comparable. For CRPS, Xaurora is better on Z500 and the non-SDE baseline on Q850.}

\begin{figure}
    \centering
    \includegraphics[width=\linewidth]{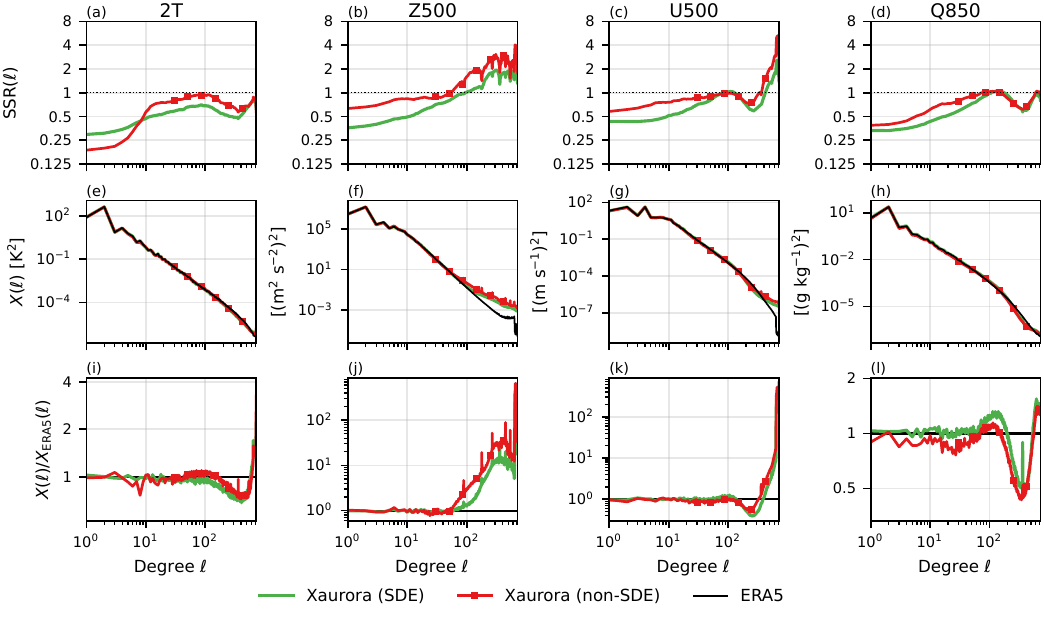}
    \caption{\rebuttal{\textbf{Spectral comparison with non-SDE baseline.} Spectral SSR (a-d), absolute (e-h) and relative (i-l) average power spectrum of the members at 5 days for 2T, Z500, U500, Q850. Xaurora is in green, the non-SDE baseline in red.}}
    \label{fig:non-sde_spectral}
\end{figure}

\rebuttal{\textbf{Spectral comparison (\cref{fig:non-sde_spectral}).} Both models suffer from large-scale underdispersion. This is most pronounced with the non-SDE model on 2T, which could be related to the poor skill observed on this variable. Small-scale overdispersion of Z500 and U500 is observed with both approaches, but most importantly for the non-SDE baseline. The non-SDE baseline also over-powers Z500 small scales and under-powers Q850 large scales. The power spectrum of the SDE version of Xaurora generally matches ERA5 more closely, except for 2T, which has slightly smoother high-frequencies.}

\subsubsection{Ensemble size}\label{app:results/ablations/ens_size}
\begin{figure}
    \centering
    \includegraphics[width=\linewidth]{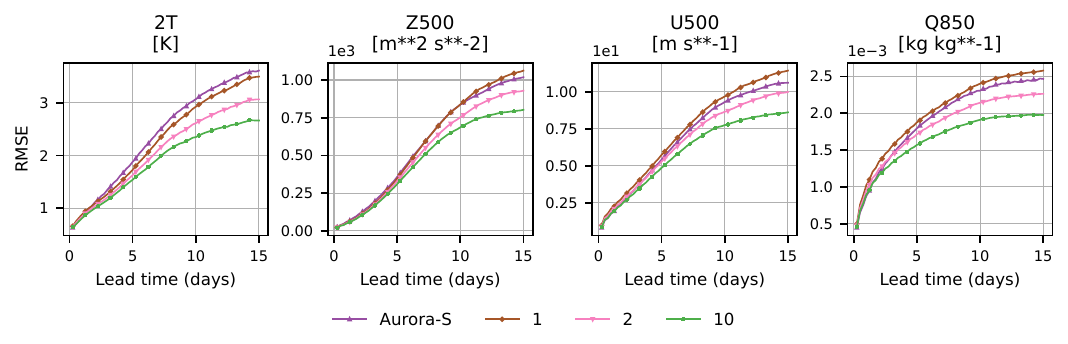}
    \caption{\chgd{\textbf{Effect of the ensemble size on the biased ensemble mean RMSE.} 15-day rollout evaluation of the RMSE (no ensemble size correction) for 2T, Z500, U500, and Q850 with ensemble sizes of 1, 2, and 10. Xaurora outperforms Aurora-S for $M\geq2$, except for 2T where a single member is sufficient.}}
    \label{fig:ens_size_ablation}
\end{figure}
Throughout this paper, we compare Xaurora ensemble mean to deterministic baselines, like Aurora-S or Aurora-L. This is technically not a fair comparison, as an ensemble has a statistical advantage over a deterministic prediction, which is one of the reasons why ensemble models are favoured over deterministic models. For completeness, we show the effect of the ensemble size on the error metrics. As expected, \cref{fig:ens_size_ablation} shows that the RMSE (without ensemble size correction) of Q850 and Z500 improves when the ensemble size grows. Nevertheless, with 2 members, Xaurora is already better than Aurora-S across variables.

\subsubsection{Finetuning vs training from scratch}\label{app:results/ablations/ft_vs_scratch}
\begin{figure}
    \centering
    \includegraphics[width=\linewidth]{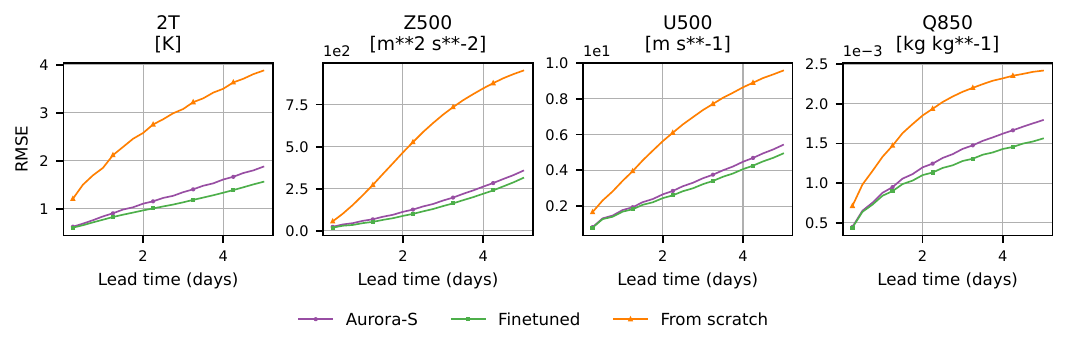}
    \caption{\chgd{\textbf{Effect of training from scratch vs finetuning on the ensemble mean uRMSE (unbiased).} 5-day rollout evaluation of the uRMSE for 2T, Z500, U500, and Q850 after training from scratch or finetuning from Aurora-S weights. The finetuned model performs much better. Both models were trained with "Single-step + MAE" (see \cref{app:results/ablations/main_paper_details}) because rollout training requires meaningful predictions at initialisation.}}
\label{fig:scratch_ablation}
\end{figure}
One benefit of Xaurora is the deterministic prior provided by Aurora-S. \cref{fig:scratch_ablation} shows a clear advantage for the finetuned model across lead times. These results should be treated carefully as a fairer comparison would add the number of samples seen during the pretraining of Aurora-S to the training from scratch, but we cannot afford this for an ablation.

\subsubsection{Naive probabilistic Aurora baseline: MC-DropPath}\label{app:results/ablations/mc_drop_path}
\begin{figure}
    \centering
    \includegraphics[width=1\linewidth]{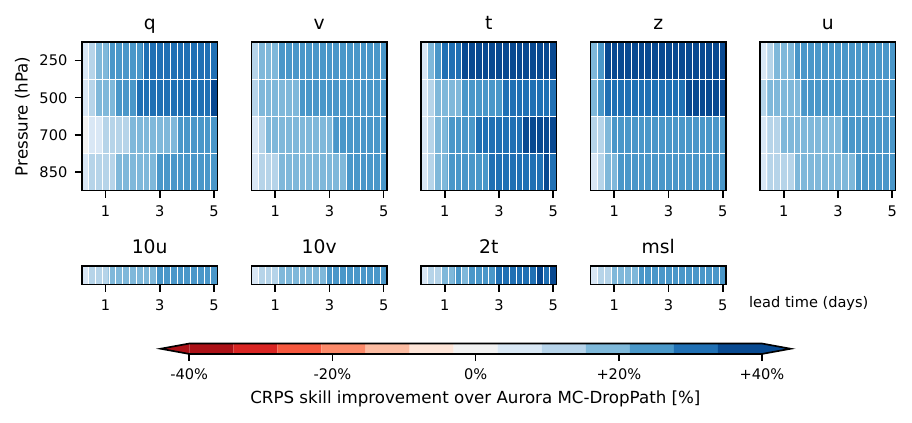}
    \caption{\chgd{\textbf{Xaurora vs Aurora MC-DropPath CRPS scorecard.} Each panel shows the CRPS improvement percentage from Xaurora over a naive probabilistic Aurora-S for one variable across lead times and pressure levels (for atmospheric variables). Both models have 10 ensemble members. Red indicates degraded performance (negative improvement), blue shows improvement over the reference.}}
    \label{fig:mc_drop_crps_scorecard}
\end{figure}

As Aurora is trained with stochastic depth (DropPath), we build a naive probabilistic baseline by keeping the mechanism on during inference, i.e., \textit{MC-DropPath} inspired by \citet{gal_dropout_2016}. \cref{fig:mc_drop_crps_scorecard} validates that Xaurora outperforms this simple baseline.

\subsection{Extreme temperature}\label{app:results/extremes}
\rebuttal{We resample the IFS ENS and FGN\footnote{GenCast-O 2022 forecasts are not available on WeatherBench. Hence, we use FGN to represent operational DLWP models. The forecasts are available at \url{gs://weatherbench2/datasets/fgn/2022-1440x721.zarr}.} \citep{alet_skillful_2025} forecasts to match Xaurora's initialisations and ensemble size. We computed the statistics of the daily maximum ($T^\text{max}$) and minimum  ($T^\text{min}$) temperature of ERA5 over the climatological period 1991-2020. We use these statistics for all models, including operational systems, which therefore have a disadvantage on shorter lead times due to mismatched analyses. Results in that range should be treated with caution.} 

\subsubsection{Definitions}\label{app:results/extremes/def}
\rebuttal{Let $\gen{T}^{(m)}_{t_0+s\dt,i,j}$ be the 2m temperature of forecast member $m$, initialised at $t_0$, at rollout step $s$ and cell $(i,j)$, and $T_{t,i,j}$ the ERA5 value at $t=t_0+s\dt$. For each lead day $d$, we gather the four \qty{6}{\hour} steps in $\mathcal{W}_d=\{4(d-1)+1,\ldots,4d\}$ and take daily extremes:}
\begin{align}
    \gen{T}^{\max,(m)}_{d,i,j} &= \max_{s\in\mathcal{W}_d} \gen{T}^{(m)}_{t_0+s\dt,i,j}, &
    \gen{T}^{\min,(m)}_{d,i,j} &= \min_{s\in\mathcal{W}_d} \gen{T}^{(m)}_{t_0+s\dt,i,j},
\end{align}
\rebuttal{and likewise the observed extremes $T^{\max}_{d,i,j}$ and $T^{\min}_{d,i,j}$ using ERA5. We define hot and cold extremes by calculating per-cell percentiles over all observed daily extremes in the climatological period. $T^{\mathrm{hot}}_{i,j}$ is the 95th percentile of $T^{\max}$, and $T^{\mathrm{cold}}_{i,j}$ the 5th percentile of $T^{\min}$. A hot extreme is $T^{\max}_{d,i,j}>T^{\mathrm{hot}}_{i,j}$, a cold extreme $T^{\min}_{d,i,j}<T^{\mathrm{cold}}_{i,j}$.}

\rebuttal{The definitions below are given for hot extremes. Cold extremes are defined analogously using minimum temperatures and the cold threshold. Scores are averaged over land cells and initialisations, with the latitude weights $w(i)$ (\cref{app:method/eval/metrics}). Ratios are taken after averaging, and deterministic models count as one-member ensembles.}

\rebuttal{\textbf{Threshold-weighted CRPS (twCRPS).} This is the fair CRPS (\cref{app:method/eval/metrics}) of the forecast and observed daily extremes after both pass through $v(x)=\max(x,T^{\mathrm{hot}}_{i,j})$. It scores only the tail beyond the threshold and remains proper.}

\rebuttal{\textbf{Brier score.} This is the weighted mean of $(p_{d,i,j}-o_{d,i,j})^2$, where $p_{d,i,j}=\frac{1}{M}\sum_m \mathbf{1}[\gen{T}^{\max,(m)}_{d,i,j}>T^{\mathrm{hot}}_{i,j}]$ is the forecast probability and $o_{d,i,j}=\mathbf{1}[T^{\max}_{d,i,j}>T^{\mathrm{hot}}_{i,j}]$ the outcome. The Brier skill score is $\mathrm{BSS}=1-\mathrm{BS}/\mathrm{BS}_c$, where $\mathrm{BS}_c$ is the Brier score of the climatological probability $p_c=0.05$ (percentile). With $M=10$ members, $p$ takes the values $p_k=k/10$. Let $W_k$ be the weighted fraction of cases with $p=p_k$, $\bar o_k$ their observed event frequency and $\bar o$ the climatological one. The Brier score then decomposes exactly as $\mathrm{BS}=\mathrm{REL}-\mathrm{RES}+\mathrm{UNC}$, with}
\begin{equation}
    \mathrm{REL}=\textstyle\sum_k W_k(p_k-\bar o_k)^2,\qquad
    \mathrm{RES}=\textstyle\sum_k W_k(\bar o_k-\bar o)^2,\qquad
    \mathrm{UNC}=\bar o(1-\bar o).
\end{equation}
\rebuttal{The reliability, REL, measures the calibration error, \emph{i.e.}, how far the reliability diagram lies from the diagonal. The resolution, RES, measures how well the forecast discriminates situations with different observed event frequencies, which prevents, for example, always predicting the climatological frequency from scoring well. We report $\mathrm{REL}/\mathrm{UNC}$ (lower is better) and $\mathrm{RES}/\mathrm{UNC}$ (higher is better), which are comparable across events and lead times.}

\subsubsection{Skill, reliability and resolution}\label{app:results/extremes/more_metrics}
\begin{figure}
    \centering
    \includegraphics[width=\linewidth]{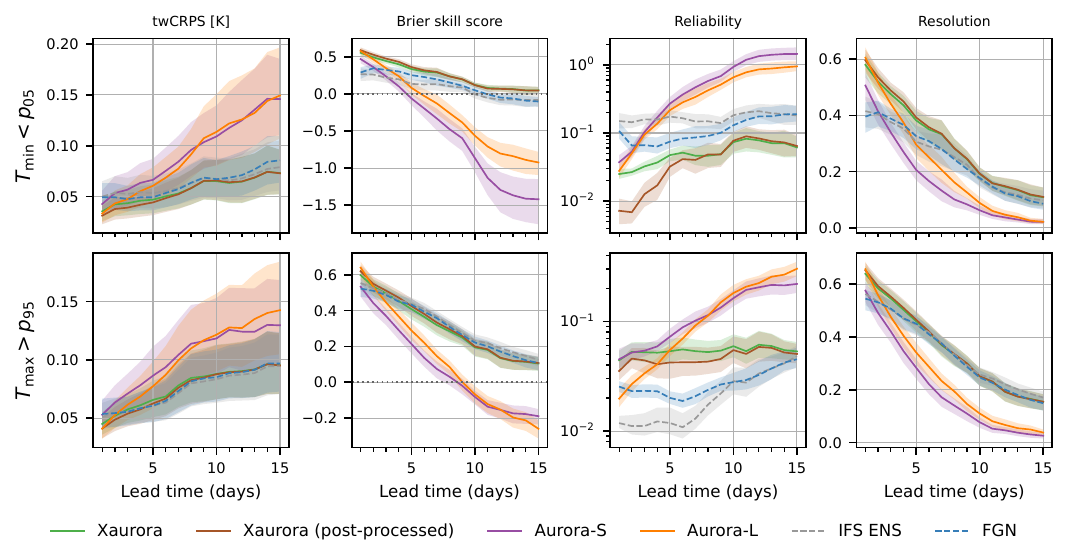}
    \caption{\rebuttal{\textbf{Skill, reliability and resolution of extreme temperature forecasts.} Threshold-weighted CRPS, Brier skill score, reliability, and resolution of Xaurora and Aurora models as well as operational systems IFS ENS and FGN. The shading indicates the 95\% bootstrap intervals over initialisations. Dashed lines highlight operational models evaluated on ERA5, which should be treated with caution on short lead times.}}
    \label{fig:extremes_lead}
\end{figure}

\rebuttal{\cref{fig:extremes_lead} shows the twCRPS and the decomposition of the Brier score in terms of reliability and resolution. On medium-range timescales, Xaurora's resolution is comparable to the operational models. Xaurora is the most reliable model on cold extremes against ERA5. On short lead times, operational models evaluated on ERA5 should be treated with caution. Hence, the improvements in reliability of cold extremes and the deterioration of hot extremes cannot be confirmed. Nevertheless, Xaurora is more reliable than Aurora-S on both extremes and post-processing improves short-term reliability.}

\begin{figure}
    \centering
    \includegraphics[width=\linewidth]{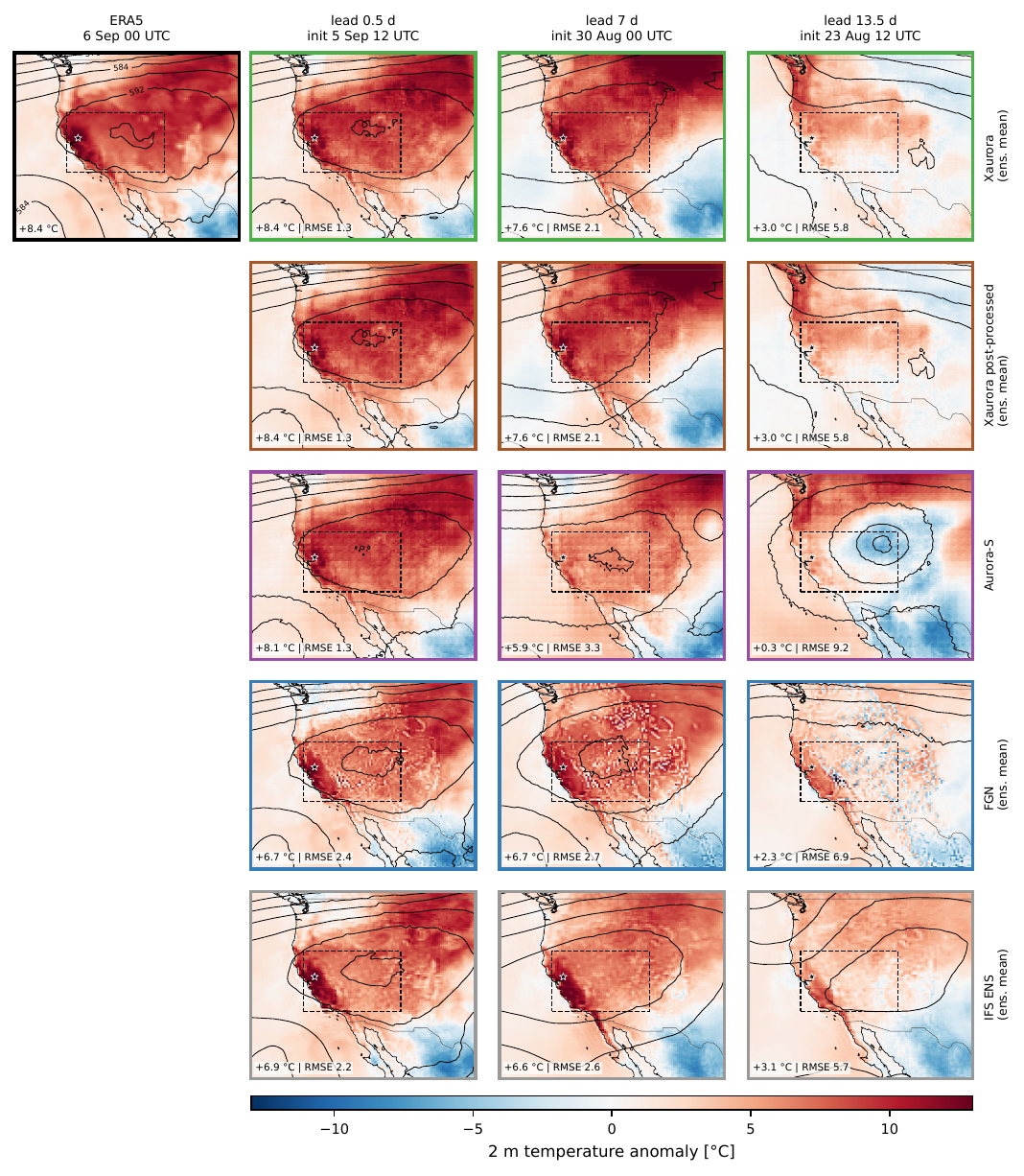}
    \caption{\rebuttal{\textbf{September 2022 western US heatwave.} 2 m temperature anomaly (colour) relative to the ERA5 1991-2020 climatology and 500 hPa geopotential height (contours) on September 6th, 2022 at 00:00 UTC, the day of the Sacramento (star) all-time record ($46.7^\circ C$). \textbf{Top left:} ERA5. \textbf{Other panels:} forecasts valid at the same time, at lead times \qty{0.5}{\day}, \qty{7}{\day}, \qty{13.5}{\day}. The ensemble mean is shown for Xaurora, Xaurora post-processed, FGN, IFS ENS. Aurora-S is deterministic. Panel text metrics are computed within the dashed box, chosen to enclose the core of the ERA5 anomaly. Post-processing rescales each member's deviation from the ensemble mean, so its members differ from those of Xaurora.}}
    \label{fig:uswest_heat_mosaic}
\end{figure}
\begin{figure}
    \centering
    \includegraphics[width=\linewidth]{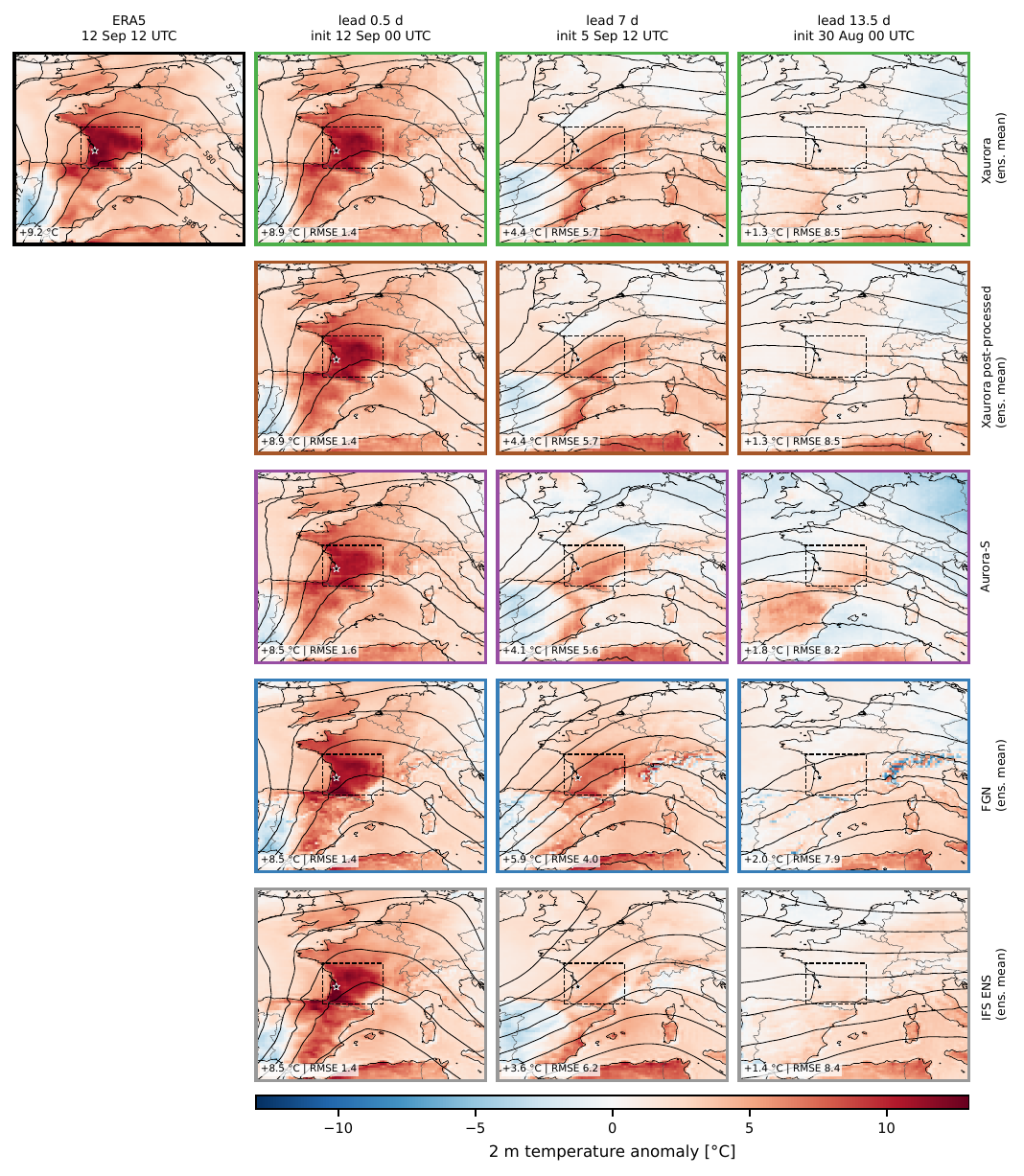}
    \caption{\rebuttal{\textbf{September 2022 southern France heatwave.} 2 m temperature anomaly (colour) relative to the ERA5 1991-2020 climatology and 500 hPa geopotential height (contours) on September 12th, 2022 at 12:00 UTC. The star shows the city of Bordeaux, at the centre of the event. \textbf{Top left:} ERA5. \textbf{Other panels:} forecasts valid at the same time, at lead times \qty{0.5}{\day}, \qty{7}{\day}, \qty{13.5}{\day}. The ensemble mean is shown for Xaurora, Xaurora post-processed, FGN, IFS ENS. Aurora-S is deterministic. Panel text metrics are computed within the dashed box, chosen to enclose the core of the ERA5 anomaly. Post-processing rescales each member's deviation from the ensemble mean, so its members differ from those of Xaurora.}}
    \label{fig:swfrance_heat_mosaic}
\end{figure}

\subsection{Native calibration with spectral interventions}\label{app:results/spectral_analysis}
\chgd{The final SDE samples and the calibrated ensemble presented in \cref{subsec:experiments/postprocessing} are separated by two steps: residual noise subtraction (\cref{app:results/importance_subtract_noise}) and spectral post-processing. Both interventions are applied after sampling. This section explores a native alternative where the random fields injected in the SDE are altered to improve the spectral properties and calibration of the raw forecasts. Since the residual noise inherits the spectrum of the injected noise, random fields with reduced high-frequency power could alleviate the need for the subtraction and affect the spectral calibration. This is similar to our post-processing approach with $\mathcal{G}(\ell)$ setting the spectral shape and $\mathcal{H}(s)$ tuning the level. However, the injected noise goes through the denoiser, which could produce flow-dependent spread. Yet, since the denoiser is highly non-linear, predicting the impact on the calibration is not trivial.}

\begin{figure}
    \centering
    \includegraphics[width=\linewidth]{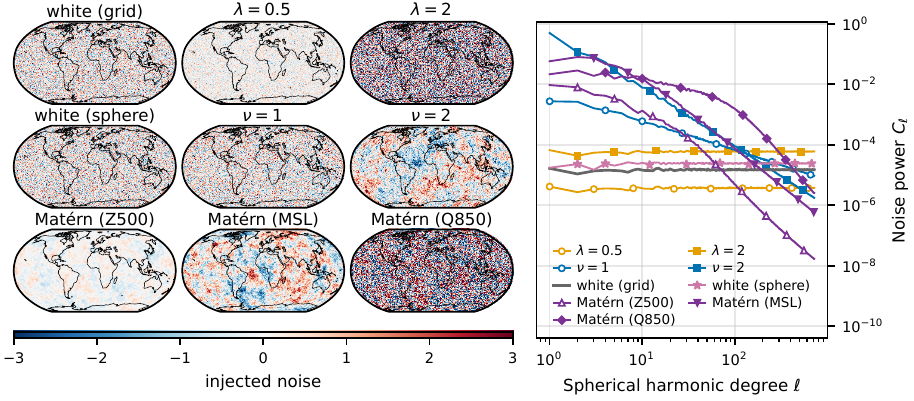}
    \caption{\chgd{\textbf{Random fields injected into the SDE.} \textbf{Maps:} one draw of each noise model on a \ang{1} grid. The $\lambda$ maps rescale the same white-noise draw ($\bz \leftarrow \lambda\bz$). The coloured fields have a power-law PSD, $\mathrm{PSD}(k)\propto k^{-\nu}$, and are normalised to unit pixel variance. White (sphere) is isotropic white noise sampled in the spherical harmonics \citep{price_gencast_2023}. The Mat\'{e}rn fields use the fitted spectrum $C_\ell$ and amplitude of each variable; Z500, MSL and Q850 are shown. The colour scale is clipped at $\pm 3$. \textbf{Right:} angular power spectrum $C_\ell$ of each noise model, averaged over 16 draws. Variance scaling shifts the white spectrum by $\lambda^2$. Colouring and the Mat\'{e}rn fits move power to large scales (low $\ell$).}}
    \label{fig:b7_noise_models}
\end{figure}

\begin{figure}
    \centering
    \includegraphics[width=1\linewidth]{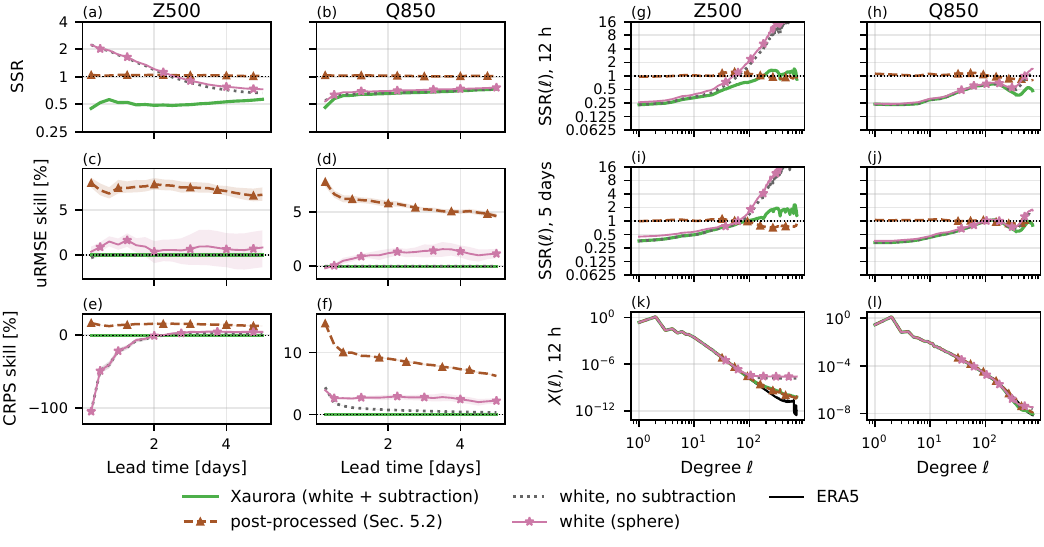}
    \caption{\chgd{\textbf{Spherical isotropic white noise: spectral calibration and skill.} Five-member ensembles from 16 initialisations in 2022, rolled out for \qty{5}{\day} with 10 Euler--Maruyama steps. All runs share initial conditions and noise seeds. The intervention is shown in pink, the Xaurora baseline in green (\cref{subsec:experiments/mwf}), the white noise without residual noise subtraction in dotted grey (\cref{fig:importance_subtract_noise}), and the post-processed forecasts in brown (\cref{subsec:experiments/postprocessing}). In each pair, the first panel is Z500 and the second Q850. \textbf{(a-b)} Global SSR. \textbf{(c-d)} uRMSE skill and \textbf{(e-f)} CRPS skill relative to Xaurora (higher is better, shaded bands are 95\% confidence intervals from a bootstrap over initialisations). \textbf{(g-h)} SSR$(\ell)$ at \qty{12}{\hour} and \textbf{(i-j)} at \qty{5}{\day} (smoothed with a 7-degree rolling geometric mean for readability; the dotted line marks perfect calibration). \textbf{(k-l)} Average power spectrum of the members, $X(\ell)$, at \qty{12}{\hour}, with ERA5 in black.}}
    \label{fig:b7_sphere}
\end{figure}

\begin{figure}
    \centering
    \includegraphics[width=1\linewidth]{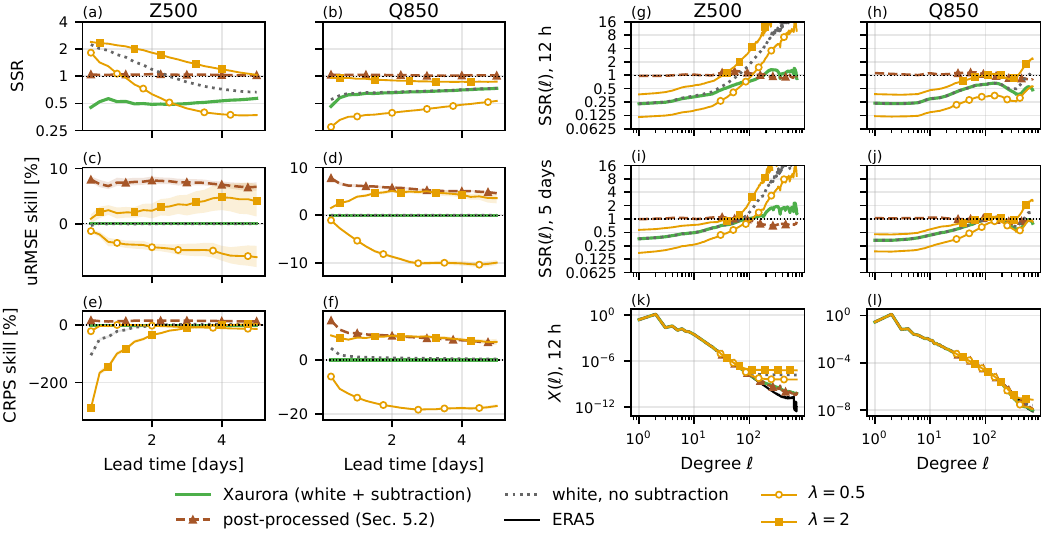}
    \caption{\chgd{\textbf{Variance scaling: spectral calibration and skill.} Five-member ensembles from 16 initialisations in 2022, rolled out for \qty{5}{\day} with 10 Euler--Maruyama steps. All runs share initial conditions and noise seeds. The interventions are shown in yellow, the Xaurora baseline in green (\cref{subsec:experiments/mwf}), the white noise without residual noise subtraction in dotted grey (\cref{fig:importance_subtract_noise}), and the post-processed forecasts in brown (\cref{subsec:experiments/postprocessing}). In each pair, the first panel is Z500 and the second Q850. \textbf{(a-b)} Global SSR. \textbf{(c-d)} uRMSE skill and \textbf{(e-f)} CRPS skill relative to Xaurora (higher is better, shaded bands are 95\% confidence intervals from a bootstrap over initialisations). \textbf{(g-h)} SSR$(\ell)$ at \qty{12}{\hour} and \textbf{(i-j)} at \qty{5}{\day} (smoothed with a 7-degree rolling geometric mean for readability; the dotted line marks perfect calibration). \textbf{(k-l)} Average power spectrum of the members, $X(\ell)$, at \qty{12}{\hour}, with ERA5 in black.}}
    \label{fig:b7_lambda}
\end{figure}

\begin{figure}
    \centering
    \includegraphics[width=1\linewidth]{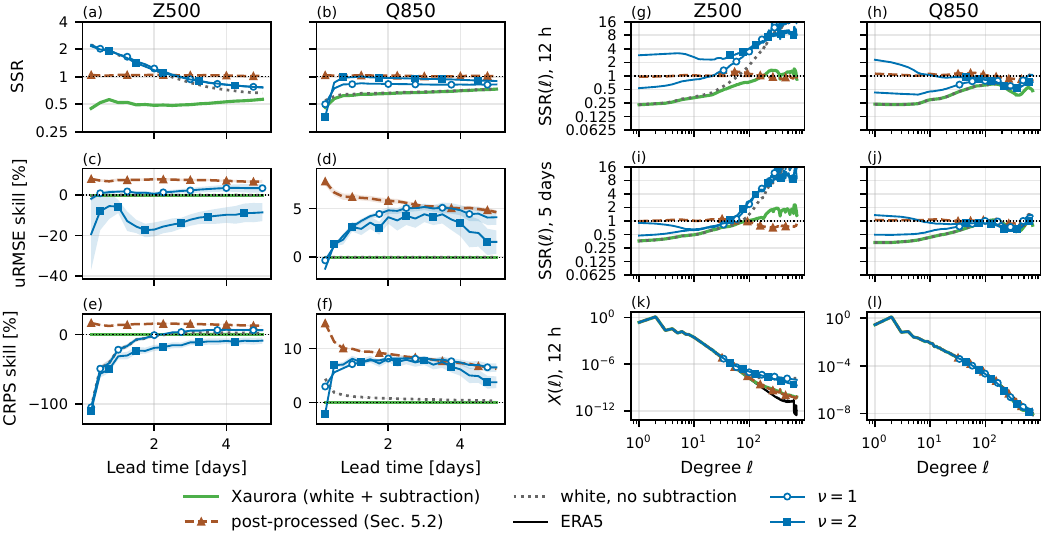}
    \caption{\chgd{\textbf{Noise colouring: spectral calibration and skill.} Five-member ensembles from 16 initialisations in 2022, rolled out for \qty{5}{\day} with 10 Euler--Maruyama steps. All runs share initial conditions and noise seeds. The interventions are shown in blue, the Xaurora baseline in green (\cref{subsec:experiments/mwf}), the white noise without residual noise subtraction in dotted grey (\cref{fig:importance_subtract_noise}), and the post-processed forecasts in brown (\cref{subsec:experiments/postprocessing}). In each pair, the first panel is Z500 and the second Q850. \textbf{(a-b)} Global SSR. \textbf{(c-d)} uRMSE skill and \textbf{(e-f)} CRPS skill relative to Xaurora (higher is better, shaded bands are 95\% confidence intervals from a bootstrap over initialisations). \textbf{(g-h)} SSR$(\ell)$ at \qty{12}{\hour} and \textbf{(i-j)} at \qty{5}{\day} (smoothed with a 7-degree rolling geometric mean for readability; the dotted line marks perfect calibration). \textbf{(k-l)} Average power spectrum of the members, $X(\ell)$, at \qty{12}{\hour}, with ERA5 in black.}}
    \label{fig:b7_nu}
\end{figure}

\begin{figure}
    \centering
    \includegraphics[width=1\linewidth]{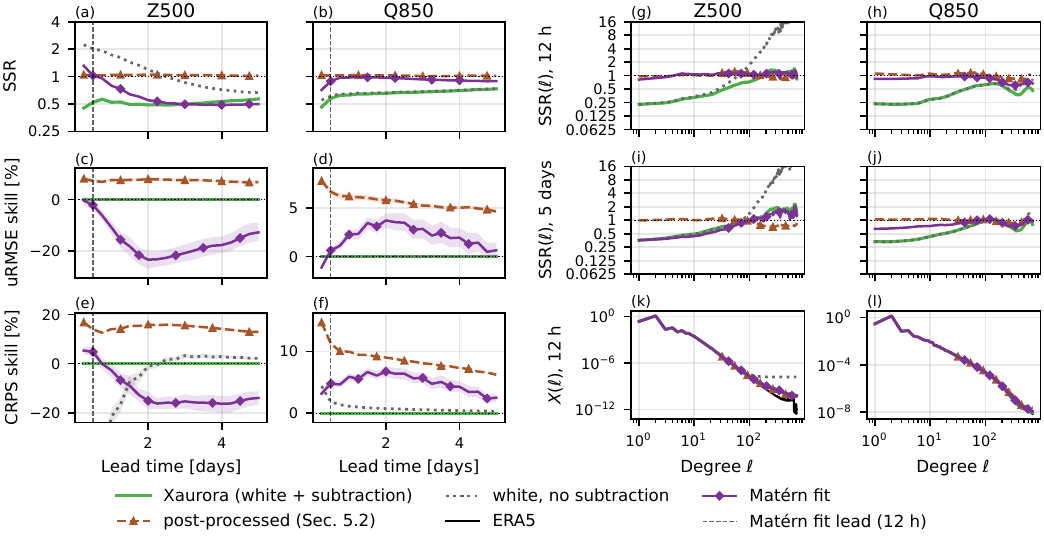}
    \caption{\chgd{\textbf{Fitted noise model: spectral calibration and skill.} Five-member ensembles from 16 initialisations in 2022, rolled out for \qty{5}{\day} with 10 Euler--Maruyama steps. All runs share initial conditions and noise seeds. The interventions are shown in purple, the Xaurora baseline in green (\cref{subsec:experiments/mwf}), the white noise without residual noise subtraction in dotted grey (\cref{fig:importance_subtract_noise}), and the post-processed forecasts in brown (\cref{subsec:experiments/postprocessing}). In each pair, the first panel is Z500 and the second Q850. \textbf{(a-b)} Global SSR. \textbf{(c-d)} uRMSE skill and \textbf{(e-f)} CRPS skill relative to Xaurora (higher is better, shaded bands are 95\% confidence intervals from a bootstrap over initialisations). \textbf{(g-h)} SSR$(\ell)$ at \qty{12}{\hour} and \textbf{(i-j)} at \qty{5}{\day} (smoothed with a 7-degree rolling geometric mean for readability; the dotted line marks perfect calibration). \textbf{(k-l)} Average power spectrum of the members, $X(\ell)$, at \qty{12}{\hour}, with ERA5 in black.}}
    \label{fig:b7_matern}
\end{figure}

\chgd{We run four experiments in which we manipulate the random fields used in the diffusion term of the SDE (\emph{i.e.} the $\bz$ in the Wiener process $\gamma_\tau \sqrt{\tau}\bz$, by default it is an \emph{i.i.d.} Gaussian) and observe the impact on calibration and skill in the spectral domain. All experiments occur at inference and use the Xaurora checkpoint and sampler of \cref{sec:experiments} without residual noise subtraction. Each method produces a 5-member ensemble from 16 initial conditions evenly spaced over 2022 (test set), rolled out for 20 steps (\qty{5}{\day}). All methods share the initial conditions and random seeds, so the comparison is paired. Sample fields are shown in \cref{fig:b7_noise_models}. The results can be found in \cref{fig:b7_sphere,fig:b7_lambda,fig:b7_nu,fig:b7_matern} and are discussed below.}

\chgd{\paragraph{Baselines: white noise, Xaurora, post-processing.} The baselines are shown in \cref{fig:b7_sphere,fig:b7_lambda,fig:b7_nu,fig:b7_matern}. Pure white noise results in overdispersion for Z500 on short lead times while Q850 is consistently underdispersed. This is explained by the spectral figures: the power of the residual noise injected by the last SDE step is high relative to the minimal power of Z500. Hence, the predictions have too much high-frequency power and small scales are strongly overdispersive, which drives the global SSR beyond 1. Subtracting the residual noise solves this problem but leaves the ensemble underdispersed. The spectral post-processing fixes the calibration across lead times, spatial scales and variables.}

\chgd{\paragraph{Spherical isotropic white noise (\cref{fig:b7_sphere}).} Following \citet{price_gencast_2023}, we sample isotropic white noise over the spherical harmonics to respect the geometry of the Earth. Gaussian spherical-harmonic coefficients (flat angular power spectrum) are transformed to the grid by an inverse spherical harmonic transform and normalised to unit variance. Sampling white noise on the sphere rather than on a grid has a negligible effect on the spectral properties and calibration. Q850 skill metrics show a slight improvement, whereas Z500 is significantly worse on short lead time CRPS. The effect on the SSR is limited.}

\chgd{\paragraph{Variance scaling (\cref{fig:b7_lambda}).} We rescale the variance of the Wiener process, \emph{i.e.} $\bz \leftarrow \lambda\bz$, with $\lambda\in\{0.5, 2\}$ ($\lambda=1$ is equivalent to white noise). We leave the drift unchanged and do not apply score correction. The residual noise power increases proportionally to the variance ($\lambda^2$) and the spectral SSR curves correspond to a near-perfect translation of the white noise baseline. Despite the residual noise that remains in the final SDE state, $\lambda=2$ works well for Q850, which improves in SSR, uRMSE, and CRPS. However, a single scaling cannot calibrate Z500 since it varies from overdispersive at 6 hours to underdispersive at 5 days.}

\chgd{\paragraph{Noise colouring (\cref{fig:b7_nu}).} To evaluate the impact of reduced high-frequency power, we sample random fields whose PSD follows a power law with exponent $\nu\in\{1,2\}$\footnote{Random fields ($\bm{\zeta}_{\nu}$) with $\mathrm{PSD}(k)\propto k^{-\nu}$ are constructed from a standard normal field ($\bm{\zeta}$) on the latitude-longitude grid with a radial amplitude filter, $\tilde{\bm{\zeta}}_\nu = \mathcal{F}^{-1}\big[A\odot\mathcal{F}[\bm{\zeta}]\big]$, with $A(\bm{k}\not=\bm{0})=\norm{\bm{k}}^{-\nu/2}$, $A(\bm{0})=0$ and $\mathcal{F}$ the 2D Fourier transform of the grid (treated as a plane, not the sphere), followed by unit-variance standardisation, $\bm{\zeta}_{\nu}=\tilde{\bm{\zeta}}_{\nu}/\mathrm{std}(\tilde{\bm{\zeta}}_{\nu})$.}, the white noise baseline corresponds to $\nu=0$. Noise colouring has a limited effect on the small scales, but seems to improve the large-scale calibration. For Q850, both values improve the SSR across lead times and are associated with higher skill. The short lead time overdispersion of Z500 cannot be addressed by the noise colour since it mostly affects the large scales. However, a small improvement can be observed with both values of the colour parameter once the variable enters the underdispersion regime. These gains are smaller than what we observed for Q850 and are not associated with any significant skill improvement.}

\chgd{\paragraph{Fitted noise model (\cref{fig:b7_matern}).} Finally, we fit a noise model to the error spectrum of Xaurora\footnote{We fit a two-component Mat\'{e}rn spectrum, $C_\ell = \sum_{i=0}^{1} F_{0,i}(1 + (L_i/R_e)^2\ell(\ell+1)/2)^{-p_i}$ (correlation lengths $L_i$, Earth radius $R_e$, smoothness $p_i$), to the unbiased ensemble-mean error spectrum of Xaurora for each channel, at \qty{12}{\hour} on 8 initial conditions and 10 members from 2021 (validation set). Fields are sampled on the sphere from spherical-harmonic coefficients scaled by $\sqrt{C_\ell}$, normalised to unit variance and mapped onto the 13 pressure levels by nearest neighbour. A per-channel amplitude is calibrated on the same data by two fixed-point iterations $\mathrm{amp}\leftarrow\mathrm{amp}/\mathrm{SSR}$.}. If the spectral properties of the noise were inherited by the forecast uncertainty, we would obtain a spread that matches the error, which is the definition of calibration. This approach effectively filters the residual high-frequency noise in the power spectra and reaches ensemble calibration at the lead time of the fit (\qty{12}{\hour}) that is comparable to the spectral post-processing. For Q850, the ensemble remains relatively well calibrated up to 5 days and shows improvements in skill up to 4 days. However, Z500 degrades beyond the lead time of the fit, showing that a single spectral correction cannot fix both over- and underdispersion. It might be possible to fit the residuals at every lead time to calibrate the ensemble over longer forecasts. However, this approach is not scalable since, unlike the post-processing, changing the structure of the injected noise at rollout step $s$ influences the residuals of consecutive steps. Therefore, we believe that a more promising approach would consist of training the model with structured noise such that it can learn the lead-dependent correction together with the spectral shape.}

\chgd{\paragraph{Summary.} The injected noise controls the high-frequency residuals and the spectral shape of the spread across scales. Coloured noise affects the large scales the most, whereas variance scaling can reduce small-scale noise. Geometry plays a minor role in the calibration and injecting noise that matches the residuals can calibrate the ensemble almost as well as the combination of residual noise subtraction and spectral post-processing at a specific lead time. Since the injected noise interacts with the sampler, generalisation across lead times is not as trivial as in the post-processing. Training with a different noise structure could potentially alleviate this limitation.}

\newpage
\begin{algorithm}
    \caption{Replay-buffer SDE-rollout training with 2-member fair CRPS}
    \label{alg:rb_crps}
    \begin{algorithmic}[1]
    \STATE {\bfseries Input:} $(\alpha_\tau,\beta_\tau,\gamma_\tau)$, Training data $\mathcal{D}_\text{train}$, pretrained denoiser $\bD_\theta$, training samples $N_\text{samples}$, optimiser, buffer capacity $N_\text{B}$, injection probability $p_\text{inject}$, SDE steps $S$, depth cap $K_{\max}$, step jitter $j$, SDE Solver, Replay buffer $\mathcal{B}$.
    \STATE {\bfseries Simplifying notations:} entry $e=(\SIstate^{(a)},\SIstate^{(b)},\bh^{(a)},\bh^{(b)},\tau,k,t)$ holds the SDE state and histories of two members $(a,b)$ at pseudo-time $\tau$, rollout depth $k$, and physical time $t$. $\mathrm{sg}(\cdot)$ is the stop-gradient operator, $\bar m$ is the member other than $m$. When $\tau>0$, SDE states and history are model-generated.
    \STATE
    \STATE $\mathcal{B}\gets\varnothing$
    \FOR{$n=1$ {\bfseries to} $N_\text{samples}$}
        \STATE \textit{// Sample entry: from the dataset or the buffer}
        \IF{$|\mathcal{B}|<N_\text{B}$ {\bfseries or} $U[0,1]<p_\text{inject}$}
            \STATE $(\sistate^{t-\dt},\sistate^{t},\sistate^{t+\dt})\sim\mathcal{D}_\text{train}$ 
            \IF{$|\mathcal{B}|=N_\text{B}$}
                \STATE evict oldest entry from $\mathcal{B}$ \quad 
            \ENDIF
            \STATE $e\gets(\SIstate^{(a)}{=}\SIstate^{(b)}{=}\SIstate^{t},\;\bh^{(a)}{=}\bh^{(b)}{=}\bh^t,\;\tau{=}0,\;k{=}0,\;t)$
        \ELSE
            \STATE $e\gets$ pop a uniformly random entry from $\mathcal{B}$
            \STATE Fetch target $\sistate^{t+\dt}$ from dataset $\mathcal{D}$
        \ENDIF
        \STATE
        \STATE \textit{// Supervision: the only gradient-tracked forward}
        \STATE $m\sim U\{a,b\}$ \quad(select supervised member)
        \STATE $\bD^{(m)}\gets\bD_\theta(\SIstate^{(m)},\bh^{(m)},\tau)$ \quad(gradient)
        \STATE $\bD^{(\bar m)}\gets\mathrm{sg}(\bD_\theta(\SIstate^{(\bar m)},\bh^{(\bar m)},\tau))$ \quad(no gradient; if $\tau{=}0$: $\bD^{(\bar m)}\gets\mathrm{sg}(\bD^{(m)})$)
        \STATE $\ell(\theta)\gets\ell_\text{fCRPS}(\sistate^{t+\dt},\bD^{(m)}, \bD^{(\bar m)};\theta)$ \quad (\cref{eq:fair_crps_loss})
        \STATE $\theta\gets\text{optimiser}(\ell(\theta),\theta)$ \quad(gradient reaches $\bD^{(m)}$ only)
        \STATE
        \STATE \textit{// Advance both members with the SDE Solver}
        \STATE $\dtau\gets\min\big((1/S)\,U[1{-}j,\,1{+}j],\;1-\tau\big)$
        \FOR{$m\in\{a,b\}$}
            \STATE $\SIstate_+^{(m)}\gets \text{SDE Solver}\big(\SIstate^{(m)},\; \drift_\theta(\cdot,\bh^{(m)},\cdot),\; \tau,\; \dtau\big)$ \quad (drift function from \cref{eq:dsi/drift_from_denoiser}, value at $\tau$ should be cached from $\bD^{(m)}$ and $\bD^{(\bar m)}$ to limit compute costs.)
        \ENDFOR
        \STATE \textit{// Push to replay buffer and handle end of SDE trajectories at $\tau{=}1$}
        \IF{$\tau+\dtau\ge 1$}
            \IF{$k+1<K_{\max}$ {\bfseries and} $\sistate^{t+2\dt} \in \mathcal{D}$}
                \FOR{$m\in\{a,b\}$}
                    \STATE $\bh^{(m)}\gets\text{slide}(\bh^{(m)},\SIstate_+^{(m)})$ \quad(generated state becomes next input)
                \ENDFOR
                \STATE push $(\SIstate_+^{(a)},\SIstate_+^{(b)},\bh^{(a)},\bh^{(b)},\;\tau{=}0,\;k{+}1,\;t{+}\dt)$ to $\mathcal{B}$ 
            \ENDIF
            \STATE \quad(else the lineage terminates; $\mathcal{B}$ shrinks $\Rightarrow$ next step injects a fresh sample)
        \ELSE
            \STATE push $(\SIstate_+^{(a)},\SIstate_+^{(b)},\bh^{(a)},\bh^{(b)},\;\tau+\dtau,\;k,\;t)$ to $\mathcal{B}$
        \ENDIF
    \ENDFOR
    \STATE {\bfseries Output:} Finetuned denoiser $\bD_\theta$
    \end{algorithmic}
\end{algorithm}

\end{document}